%% file: main.tex
\documentclass[acmtog]{acmart}

\usepackage{algorithm}  
\usepackage{algorithmic}

\input{ourcommands}

\AtBeginDocument{%
  }

\setcopyright{acmlicensed}
\copyrightyear{2025}
\acmYear{2025}
\acmDOI{XXXXXXX.XXXXXXX}

\begin{document}

\title{RemixFusion: Residual-based Mixed Representation for Large-scale Online RGB-D Reconstruction}

\author{Yuqing Lan}
\orcid{0000-0003-0546-355X}
\authornote{Both authors contributed equally to this research.}
\email{lanyuqingkd@nudt.edu.cn}
\affiliation{%
  \department{National Key Laboratory of Parallel and Distributed Computing} 
  \institution{National University of Defense Technology}
  \country{China}}

\author{Chenyang Zhu}
\orcid{0000-0003-2838-8601}
\authornotemark[1]
\affiliation{%
  \institution{National University of Defense Technology}
  \country{China}}
\email{zhuchenyang07@nudt.edu.cn}

\author{Shuaifeng Zhi}
\orcid{0000-0002-5927-5426}
\affiliation{%
  \institution{National University of Defense Technology}
  \country{China}}
\email{zhishuaifeng@outlook.com}

\author{Jiazhao Zhang}
\orcid{0000-0001-9459-293X}
\affiliation{%
  \institution{Peking University}
  \country{China}}
\email{jiazhao.zhang@stu.pku.edu.cn}

\author{Zhoufeng Wang}
\orcid{0009-0005-8786-8500}
\affiliation{%
  \institution{National University of Defense Technology}
  \country{China}}
\email{wangzhoufeng7346@gmail.com}

\author{Renjiao Yi}
\orcid{0000-0002-6057-1089}
\affiliation{%
  \institution{National University of Defense Technology}
  \country{China}}
\email{yirenjiao@nudt.edu.cn}

\author{Yijie Wang}
\orcid{0000-0002-2913-4016}
\authornote{Corresponding authors.}
\affiliation{%
  \department{National Key Laboratory of Parallel and Distributed Computing}
  \institution{National University of Defense Technology}
  \country{China}}
\email{wangyijie@nudt.edu.cn}

\author{Kai Xu}
\orcid{0000-0002-9054-0216}
\authornotemark[2]
\affiliation{%
  \institution{National University of Defense Technology and Xiangjiang Laboratory}
  \country{China}}
\email{kevin.kai.xu@gmail.com}

\renewcommand{\shortauthors}{Lan, Y. et al.}

\begin{teaserfigure}
    \includegraphics[width=\textwidth]    {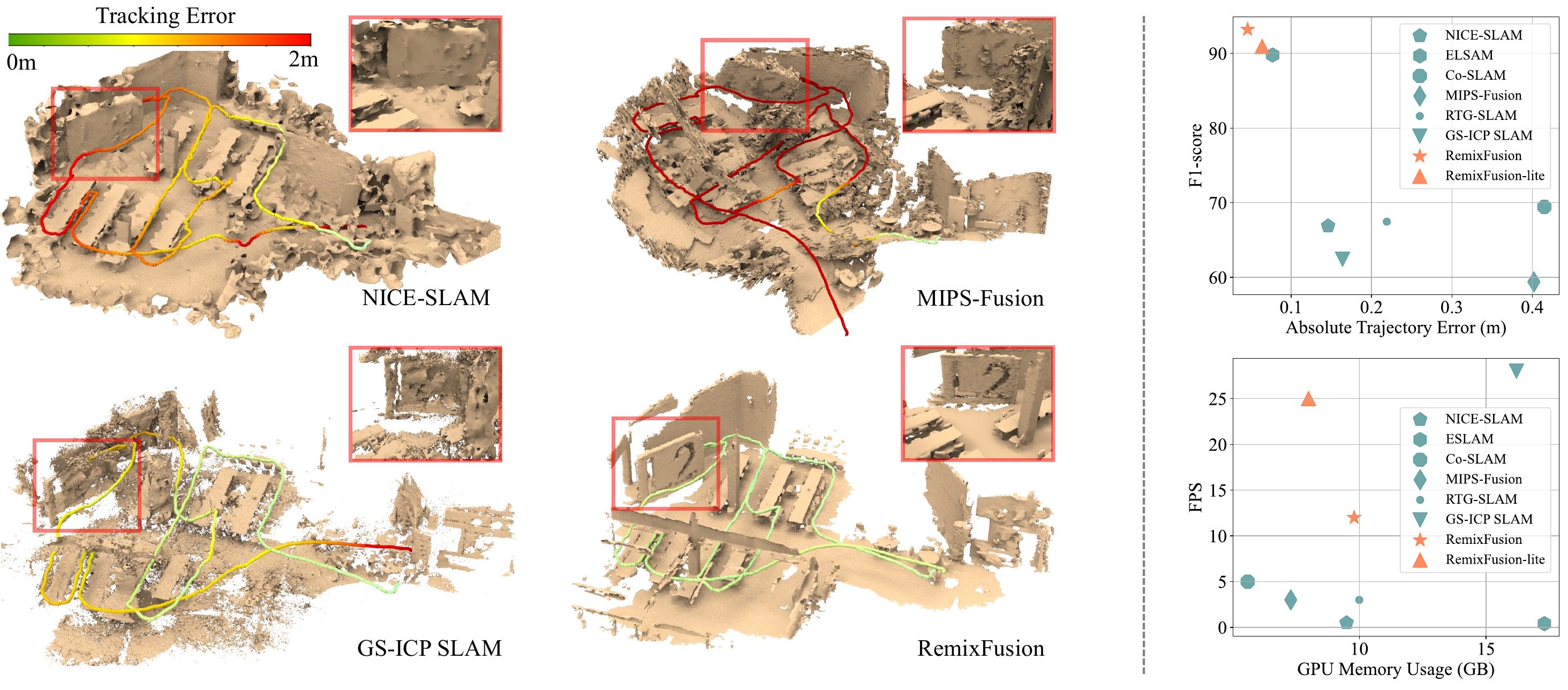}
    \caption{We present RemixFusion, a residual-based RGB-D framework by virtue of both explicit and implicit representations for large-scale online dense reconstruction. RemixFusion can support real-time fine-grained reconstruction in a memory-efficient way. It only costs 9.8GB GPU memory with 12 FPS for the about $400m^2$ reconstruction above, while other methods~\cite{zhu2022nice,tang2023mips,johari2023eslam} struggle in both tracking and reconstruction 
    in real time.
    Traditional explicit methods fail for this scene. GS-ICP SLAM~\cite{ha2024rgbd} is the SOTA 3DGS-based SLAM. The average results of reconstruction and tracking on the \texttt{BS3D} dataset, as well as the system FPS and GPU memory usage on the above scene, are shown on the right, which illustrate that RemixFusion obtains better performance and efficiency. RemixFusion-lite denotes the lightweight version and achieves decent performance with about 25 FPS. }
    \Description{Visualization of the tracking and mapping results of different methods on a cafeteria, and RemixFusion is the best method in terms of performance and efficiency.}
    \label{fig:teaser}
\end{teaserfigure}

\begin{abstract}
    \input{01-abstract}

\end{abstract}



\received{20 February 2007}
\received[revised]{12 March 2009}
\received[accepted]{5 June 2009}

\maketitle

\input{02-introduction}

\input{03-relatedwork}
\input{04-method}

\input{06-experiments}

\input{07-conclusion}

\begin{acks}
This work is supported in part by the NSFC (62522219), the Major Program of Xiangjiang Laboratory (23XJ01009), NSFC (62325211, 62132021, 62372457, 62572477), Young Elite Scientists Sponsorship Program by CAST (2023QNRC001), the Natural Science Foundation of Hunan Province of China (2021RC3071, 2022RC1104), the National Science and Technology Major Project (2022ZD0115302), NSFC (61379052), the Science Foundation of Ministry of Education of China (2018A02002), the Natural Science Foundation for Distinguished Young Scholars of Hunan Province (14JJ1026).
\end{acks}

\bibliographystyle{ACM-Reference-Format}
\bibliography{ref}

\appendix
\input{appendix}











\end{document}

%% file: ourcommands.tex
\usepackage{multirow}

\usepackage{multicol}

\usepackage{color}
\definecolor{green}{rgb}{0, 0.4, 0}
\definecolor{red}{rgb}{1.0, 0.0, 0.0}
\definecolor{teal}{rgb}{0.0, 0.4, 0.4}
\definecolor{purple}{rgb}{0.65,0,0.65}
\definecolor{saffron}{rgb}{0.95,0.75,0.2}
\definecolor{turquoise}{rgb}{0.0,0.5,0.5}
\definecolor{brown}{rgb}{0.5, 0.16, 0.16}

\usepackage{overpic}

\newcommand{\fix}[1]{{\color{black}#1}}
\newcommand{\ffix}[1]{{\color{black}#1}}

\definecolor{lightgray}{rgb}{0.6, 0.6, 0.6}

\definecolor{DeltaColor}{rgb}{0.039,0.73,0.71}
\definecolor{SetaColor}{rgb}{0.867, 0.0235, 0.376}
\definecolor{SigmaColor}{rgb}{0.98,0.45,0.0}
\definecolor{RedColor}{rgb}{0.8,0,0}
\definecolor{AlphaColor}{rgb}{1.0, 0.4, 0.8}
\definecolor{BetaColor}{rgb}{0.8,0,0.8}
\definecolor{GammaColor}{rgb}{0.0,0,0.7}
\definecolor{EpsilonColor}{rgb}{0.353,0.725,0.906}
\definecolor{TauColor}{rgb}{0.423,0.235,0.192}

\newcommand{\un}[1]{\underline{#1}}
\newcommand{\oout}[1]{--}



%% file: 01-abstract.tex
The introduction of the neural implicit representation has notably propelled the advancement of online dense reconstruction techniques. Compared to traditional explicit representations, such as TSDF, it substantially improves the mapping completeness and memory efficiency. However, the lack of reconstruction details and the time-consuming learning of neural representations hinder the widespread application of neural-based methods to large-scale online reconstruction.
We introduce RemixFusion, a novel residual-based mixed representation for scene reconstruction and camera pose estimation dedicated to high-quality and large-scale online RGB-D reconstruction.
In particular, we propose a residual-based map representation comprised of an explicit coarse TSDF grid and an implicit neural module that produces residuals representing fine-grained details to be added to the coarse grid.
Such mixed representation allows for detail-rich reconstruction with bounded time and memory budget, contrasting with the overly-smoothed results by the purely implicit representations, thus paving the way for high-quality camera tracking. Furthermore, we extend the residual-based representation to handle multi-frame joint pose optimization via bundle adjustment (BA). 
In contrast to the existing methods, which optimize poses directly, we opt to optimize pose changes.
Combined with a novel technique for adaptive gradient amplification, our method attains better optimization convergence and global optimality.
Furthermore, we adopt a local moving volume to factorize the whole mixed scene representation with a divide-and-conquer design to facilitate efficient online learning in our residual-based framework.
Extensive experiments demonstrate that our method surpasses all state-of-the-art ones, including those based either on explicit or implicit representations, in terms of the accuracy of both mapping and tracking on large-scale scenes.

%% file: 02-introduction.tex
\section{INTRODUCTION}

Online dense reconstruction based on RGB-D cameras has made significant advances in recent years. Lately, the progress has been propelled by the fast development of robust tracking based on randomized optimization~\cite{zhang2021rosefusion,zhang2022asro} and scalable mapping based on neural map representation~\cite{wang2021neus,chen2022tensorf,li2023neuralangelo}.
The latter has been incurring an in-depth revolution in scene representation methods, from explicit volumetric fields or point clouds to implicit neural fields of occupancy or radiance. 

Implicit scene representation leads to improved completeness of dense reconstructions, yet it also introduces new challenges in scene mapping and pose tracking. These challenges are particularly evident in large-scale scene reconstruction. As the size of the scenes scales up, it becomes increasingly difficult to balance the high-precision reconstruction of details with the significant computation and memory overheads required by online systems. In the traditional volumetric fusion approach~\cite{Izadi2011}, depth maps are directly fused into a TSDF field. The reconstructed details can be aligned with the depth maps as long as sufficient TSDF resolution is adopted, albeit at the expense of higher storage cost. When working with implicit scene representations~\cite{sucar2021imap,zhu2022nice,wang2023co,johari2023eslam,zhang2023go}, however, this issue is not straightforward to address as reconstruction details are encoded with neural networks. Even with larger networks and extended training time, encoding high-frequency geometric information remains difficult, and the problem is exacerbated as the scene scales up. Consequently, to make an online dense reconstruction system real-time capable with a limited memory footprint, neither explicit nor implicit scene representation can balance between effective modeling of fine-grained details and efficient reconstruction of scenes at scale.

Lack of details in real-time reconstruction also adversely affects camera tracking in an online reconstruction system, which consequently decreases the overall reconstruction quality. Current methods mostly rely on optimizing a rendering loss in a frame-to-model approach for camera pose estimation~\cite{zhu2022nice,wang2023co}. 
In large-scale reconstructions, excessively flat optimization gradients lead to poor convergence. This issue is particularly prominent in multi-frame pose optimization based on bundle adjustment, causing severe optimization oscillations and failure to achieve the desired joint optimization effect, as we will show in the experiments (see Table~\ref{tab:ablation-ba} and Figure~\ref{fig:exp-pmvis}).

Residual learning has been proven in various works~\cite{he2016deep,xiangli2022bungeenerf,kim2016multimodal,qin2022geometric} to 
improve the convergence of neural network training, enhance generality, and enable the network to encode rich details. These advantages are especially useful when learning implicit representations for large-scale scene reconstruction. Motivated by that, we propose RemixFusion, adopting the concept of residuals with a mixed representation of implicit and explicit, for both detail-rich mapping and accurate pose tracking in online dense reconstruction. To address the inefficiency of implicit representations in capturing high-frequency details in large scenes, we first combine explicit and implicit scene representations in a residual form. The base of our mixed representation is a coarse-grained explicit 3D grid volume that stores low-frequency scene structure. 
High-frequency geometry details, which serve as residuals, are captured in the parameters of the neural module and coupled with the coarse-grained base.
This approach reduces memory overhead by lowering the resolution of the explicit representation while preserving scene details in a memory-efficient form through neural representation. Furthermore, the training complexity of the residual neural module is significantly reduced since it focuses on encoding only high-frequency features. Thus, this mixed representation not only preserves fine-grained details in a memory-efficient way but also improves online learning efficiency. 

For pose estimation, based on the mixed representation, we propose a residual-based multi-frame pose optimization method. 
Specifically, we estimate residual pose corrections to refine the initial poses from front-end tracking, thereby enhancing geometric consistency during bundle adjustment (BA).
By encoding only residual poses in 
the network,
our approach allows the
BA
based on implicit representations to focus on optimizing pose changes rather than absolute poses, 
thus improving the learning efficiency and multi-view consistency. 
In practice, the real-time constraints make it impractical to perform bundle adjustment on the whole reconstruction. The sparse sampling of the scene results in
discontinuous perception of surface details during joint optimization, which in turn reduces the optimization's ability to escape local minima. 
To address this challenge, we introduce 
an adaptive gradient amplification based on the reconstructed surface,
allowing the BA to obtain better optimization gradients even when the detailed perception is discontinuous in real-world large-scale scenarios.

As shown in Figure~\ref{fig:teaser}, our online dense reconstruction method, employing an effective fusion of explicit and implicit representations, achieves fine-grained reconstruction of large-scale scenes with a relatively low GPU memory cost. Our method exhibits significantly improved online learning efficiency thanks to the explicit map, resulting in a frame rate that is 
2.4
times higher than that of other methods based solely on implicit representation. Furthermore, the residual-based camera pose bundle adjustment benefits pose tracking, which improves tracking accuracy by 
$28.1\%$  
over the state-of-the-art dense SLAM 
~\cite{johari2023eslam} 
on BS3D~\cite{mustaniemi2023bs3d}. 
In summary, our contributions include:

\begin{itemize}
    \item We propose a mixed residual-based representation for dense RGB-D reconstruction of large-scale scenes, which preserves fine-grained details with relatively low memory and computational cost.
    \item We propose a residual-based bundle adjustment technique that employs 
    a tiny MLP for
    residual-based pose refinement.
    Compared to traditional BA, our method improves pose estimation in terms of both efficiency and robustness.
    \item We have implemented an efficient system of online RGB-D dense reconstruction, which realizes robust and fine-grained real-time reconstruction for large scenes over 1000$m^2$ with an affordable GPU memory footprint.
\end{itemize}

%% file: 03-relatedwork.tex
\section{RELATED WORK}

The field of online reconstruction methods constitutes a substantial area of research. In this context, we review the most relevant literature on large-scale indoor scene reconstruction, including representations based on both explicit and implicit neural methods.

\paragraph{Explicit methods.}
The explicit representation has been widely studied in the last decades, including voxel-based and point-based methods. 
In the context of voxel-based methods, KinectFusion~\cite{Izadi2011} proposes the Truncated Signed Distance Function (TSDF) for encoding the scenes with a consumer depth sensor. Considering the inefficiency of encoding the empty space of the scene, follow-up works~\cite{Whelan2012, Roth2012} extend the KinectFusion by leveraging a moving TSDF volume to encode larger scenes. \cite{niessner2013real, dai2017bundlefusion} encode the scenes with hash blocks to further improve the scalability of scene reconstruction. Global bundle adjustment is utilized to reduce the drift in pose estimation and enhance model consistency in terms of large-scale indoor environments~\cite{Dai2017}. 
For points-based methods, ~\cite{keller2013real,Whelan2015,Whelan2016ElasticFusionRD} leverages points and surfels to achieve scalability and flexibility for encoding scenes. With explicit scene representation, these methods leverage 
classic second-order optimization methods, such as  Gaussian-Newton or Levenberg-Marquardt (LM).
Recently, ~\cite{zhang2021rosefusion, zhang2022asro} propose random optimization to improve the robustness of camera tracking and demonstrate good performance on large-scale fast-moving motions. 
However, explicit methods suffer from memory consumption, especially in large-scale scenes.

\paragraph{Implicit methods.} 
Inspired by NeRF~\cite{mildenhall2021nerf}, the pioneering work of NeRF-based SLAM, iMAP~\cite{sucar2021imap} proposes to use keyframe-based joint optimization via differential volume rendering and encode the entire scene in a multilayer perceptron (MLP). To improve geometric details, NICE-SLAM~\cite{zhu2022nice} and Vox-Fusion~\cite{yang2022vox} incorporate multi-scale feature grids with shallow decoders to collaborate on memorizing the geometry and texture. Recent advanced works, Co-SLAM~\cite{wang2023co}, ESLAM~\cite{johari2023eslam} and Point-SLAM~\cite{Sandstrm2023PointSLAMDN} exploit more efficient parametric embedding like multi-resolution hash encoding~\cite{muller2022instant}, multiscale feature planes~\cite{chan2022efficient} and data-driven neural point clouds~\cite{Xu2022PointNeRFPN} for memory efficiency and faster convergence speed. These approaches seamlessly integrate pose estimation and reconstruction by a render-and-align pattern. ~\cite{Xu2022HRBFFusionA3} addresses robust pose estimation of implicit representations of Hermite Radial Basis Functions (HRBFs). Alternative approaches decouple tracking from reconstruction and seek more accurate and robust pose estimation, paying less attention to the reconstruction quality. Feature matching is utilized in traditional approaches~\cite{chung2022orbeez}, while neural approaches~\cite{koestler2022tandem,teed2021droid,zhang2023go} leverage the deep multi-view stereo or learnable optical flow estimator to achieve frame-to-frame registration. However, these methods struggle to preserve real-time and detailed reconstruction in large scenes.

\paragraph{Representations in large scenes.}
Reconstruction of large-scale scenes requires efficient representations. As for explicit voxel-based methods, hashing schemes~\cite{Niessner2013, Dai2017} and octrees~\cite{liu2020neural,mao2023ngel} are adopted to reduce the memory footprint. 
However, these methods only reduce the storage required for empty space
and lack the ability of completion. In terms of implicit methods, learnable hash grids~\cite{wang2023co,zhang2023go}, tri-planes~\cite{johari2023eslam}, and neural points~\cite{Sandstrm2023PointSLAMDN,hu2023cp} are utilized to achieve the trade-off of efficiency and reconstruction quality. However, these methods suffer from the high latency of the model update. Another line of work leverages submaps~\cite{tancik2022block,tang2023mips,mao2023ngel} to dynamically allocate the implicit maps with the moving camera. The multi-submap techniques improve the scalability while costing much more time in submap management. Recently, 3D Gaussian Splatting (3DGS)~\cite{kerbl20233d} shows promising results in novel view synthesis. A series of works
~\cite{keetha2024splatam,huang2024photo,matsuki2024gaussian,ha2024rgbd}
explore applying 3DGS to monocular or RGB-D SLAM.
Submaps and loop closure techniques~\cite{zhu2025_loopsplat} are incorporated to improve the tracking accuracy. Furthermore, a more compact representation~\cite{peng2024rtgslam} is proposed to enhance the efficiency of 3DGS.
These methods demonstrate high-fidelity rendering and generalization capabilities in different scenarios. 
However, these 3DGS-based methods are 
still
less memory-friendly than implicit methods
since they need numerous 3DGS for the explicit model, especially in large-scale scenes. 
We propose to take advantage of both explicit and implicit representations to remain both memory-efficient and time-efficient via residual mixing. 

%% file: 04-method.tex
\section{METHOD}

\subsection{Overview}

\begin{figure*}[!t]
  \centering
  \includegraphics[width=1.0\linewidth]{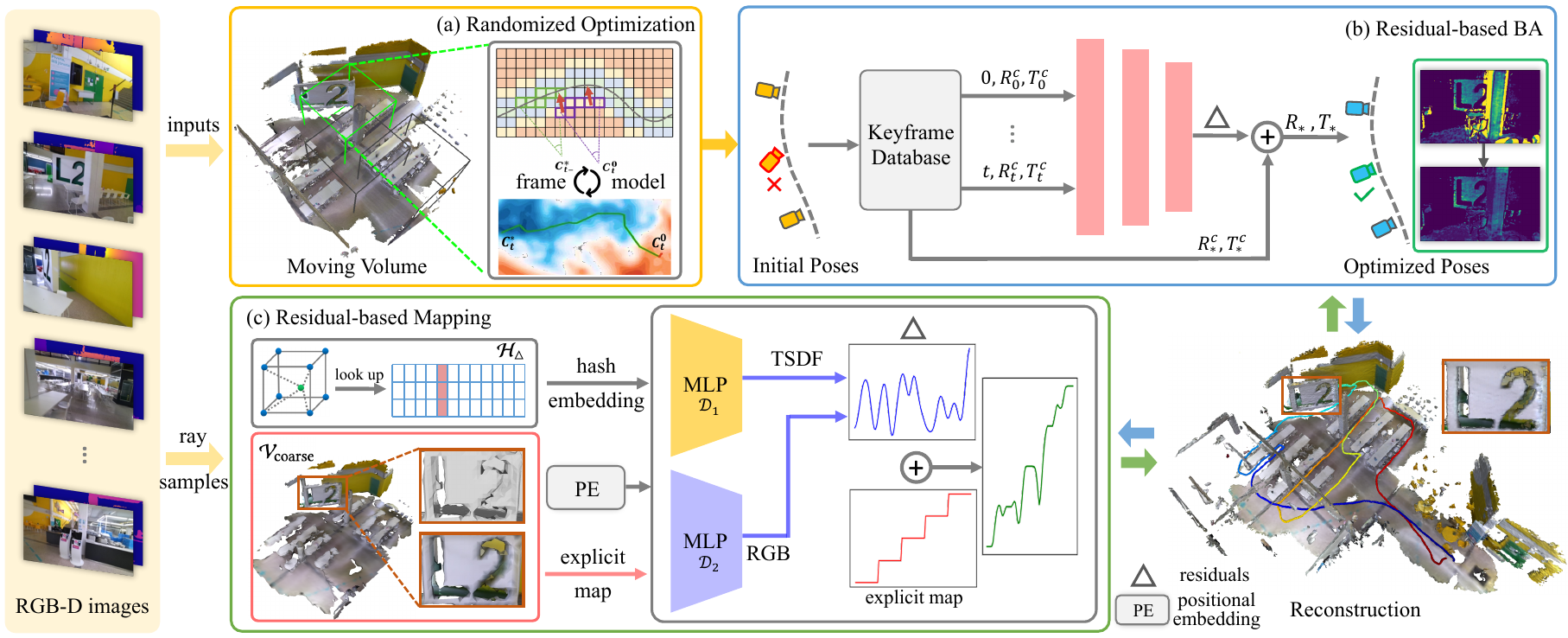}
  \caption{Method overview. (a) Given RGB-D inputs, the pose estimation is based on the frame-to-model randomized optimization on a scalable moving volume, providing the initial pose estimation. (b) Based on the initial poses, a global MLP is utilized to output the residuals for multi-view consistent pose refinement, using the rendering loss and geometric loss, which are backward propagated through the global reconstruction model. (c) For reconstruction, RemixFusion consists of a coarse TSDF grid $\mathcal{V}_{\text{coarse}}$, which records the low-frequency scene structure, and an implicit neural map $\Theta$ including the hash embedding \fix{$\mathcal{H}_{\triangle}$} and tiny decoders $\mathcal{D}$ ($\mathcal{D}_1$ and $\mathcal{D}_2$), which encode the high-frequency geometry details. TSDF and RGB residuals are decoded based on these embeddings, which are added to the coarse grid to recover the final reconstruction. 
  The residual-based BA and mapping are parallel to the front-end tracking.
  The residual designs in both pose estimation and reconstruction ensure efficiency and accuracy.}
  \Description{The proposed modules in RemixFusion are necessary and beneficial to each other.}
  \label{fig:method-pipeline}     
\end{figure*}   

In this section, we first introduce our residual-based framework, including the explicit-implicit mixed scene representation and residual-based bundle adjustment for camera poses. Our insight behind these designs is that the residual can be adopted to force networks to focus on the high-frequency details upon the existing coarse geometry, relieving the burden of neural learning and enhancing efficiency. In this way, the proposed online dense reconstruction framework is both storage-efficient and time-efficient, which is important for application in large-scale scenes with real-time requirements.

More specifically, as shown in Eq. \ref{eq:1}, we decouple the complete representation $\mathcal{F}$ into a coarse representation $\mathcal{F}_c$ and a residual one $\mathcal{F}_{\triangle}$, which are optimized in reconstruction and pose estimation.
In general, $\mathcal{F}_c$ is expected to be fast and robust, such as TSDF, providing the coarse 
geometry seamlessly with the input data. $\mathcal{F}_{\triangle}$ is supposed to be more expressive and capture the fine-grained details based on $\mathcal{F}_c$. The coarse estimation $\mathcal{F}_c$ is usually constructed based on explicit representations, and the neural networks are adopted for residual refinement $\mathcal{F}_{\triangle}$. The coarse estimation and residual refinement are performed alternately for reconstruction and pose estimation. 

\begin{equation}
  \label{eq:1}
  \mathcal{F}=\mathcal{F}_{c}\oplus\mathcal{F}_{\triangle},
  \end{equation}
where \fix{$\oplus$} denotes operation to aggregate $\mathcal{F}_{c}$ and 
$\mathcal{F}_{\triangle}$.

The overview of our method is demonstrated in Figure \ref{fig:method-pipeline}. Given the successive RGB-D inputs, we first leverage the scalable randomized optimization with limited and fixed footprints for the coarse pose estimation (Fig.~\ref{fig:method-pipeline}(a)). Keyframes are selected from these input frames. The corresponding initial camera poses are fed into a
tiny
MLP
for the joint residual pose refinement (Fig.~\ref{fig:method-pipeline}(b)).

Note that the proposed residual-based BA of the camera poses is optimized by the gradients given by the multi-view constraints based on the reconstructed scene. To improve pose estimation, an explicit-implicit scene representation mixture is introduced simultaneously. The online learning (Fig.~\ref{fig:method-pipeline}(c)) of this representation depends on the optimized camera poses. 
Specifically, we perform ray casting and sample many points along the ray directions accordingly. Embeddings of these points, including the hash and explicit embedding together with positional embedding, are sent to two MLPs for TSDF and RGB residual prediction. The residuals indicating the high-frequency details are 
added to the explicit coarse base, 
outputting the final reconstruction. 
The residual-based BA and residual-based mapping are running alternately with continuously updated global reconstruction and optimized keyframe poses.

In the following sections, we first introduce the residual-based mapping (Section~\ref{sec:mapping}), and then demonstrate the optimization of residual-based BA based on the global residual-based map (Section~\ref{sec:ba}). The strategies of the scalable randomized optimization and other implementation techniques are illustrated in Section~\ref{sec:system}.

\subsection{Residual-based Mapping}
\label{sec:mapping}

In large-scale scenes, various layouts and complex structures make it challenging for the scene representations to remain both expressive and efficient.
Shown in Figure~\ref{fig:method-pipeline}(c), we employ a mixed representation to address this challenge. This representation consists of an explicit coarse TSDF grid $\mathcal{V}_{\text{coarse}}$ 
and an implicit neural map $\Theta$, which correspond to $\mathcal{F}_c$ and 
$\mathcal{F}_{\triangle}$
in Eq. \ref{eq:1} respectively. The key insight of the residual-based representations is to obtain the coarse reconstruction in real time and provide the initial \fix{surface} for the implicit module to efficiently deform and refine. Consequently, given an arbitrary 3D position $p$, the attributes $\mathcal{O}(p)=\{\mathcal{O}^{\text{TSDF}}(p),\mathcal{O}^{\text{RGB}}(p)\}$ which are the geometry and appearance of $p$ can be formed in a residual-based aggregation as:
\fix{
\begin{equation}
\label{eq:Op}
    \mathcal{O}(p) =\text{TriLerp}(\mathcal{V}_{\text{coarse}}(p))+\mathcal{D}(\Theta(p)),
\end{equation}}
where $\text{TriLerp}(\cdot , \cdot)$ denotes the trilinear interpolation operation to query RGB and TSDF values for $p$ and $\mathcal{D}$ is the decoder for the implicit neural map $\Theta$.

\paragraph{Mixed Scene Representation}
Given the posed RGB-D frames, the explicit component $\mathcal{V}_{\text{coarse}}$ of the proposed scene representation is constructed via TSDF Fusion~\cite{curless1996volumetric} with a relatively low resolution to store the coarse TSDF and RGB attributes, serving as explicit
base
of our reconstruction. Similar to some previous implicit-based methods, the implicit component $\Theta$ includes two kinds of embedding functions, which are hash embedding 
$\mathcal{H}_{\triangle}$~\cite{muller2022instant} and positional embedding $\rho$~\cite{muller2019neural}, to encode the reconstruction residuals for an arbitrary position upon the $\mathcal{V}_{\text{coarse}}$. Specifically, adopting this joint embedding in the representation can balance the training efficiency and reconstruction quality~\cite{wang2023co}, which is critical for online reconstruction. Furthermore, the residual learning depends on the explicit base, and the final residual-based aggregation to calculate the attributes for a position $p$ can be formed based on Eq.~\ref{eq:Op} as:
\begin{equation}
\label{eq:Op_final}
    \mathcal{O}(p) = \beta (p)+\mathcal{D}(\beta (p), \rho(p), \mathcal{H}_{\triangle}(p)),
\end{equation}
\fix{
\begin{equation}
\label{eq:trilinear}
    \beta(p) = \text{TriLerp}(\mathcal{V}_{\text{coarse}}(p)),
\end{equation}}
where $\text{TriLerp}(\cdot , \cdot)$  denotes the trilinear interpolation operation for $p$ on the grid $\mathcal{V}_{\text{coarse}}$ to query the coarse RGB and TSDF values. $\rho (p)$ denotes the positional embedding.

\paragraph{Neural Learning of the implicit residual}
Different from the $\mathcal{V}_{\text{coarse}}$ can be reconstructed directly based on the fusion of the posed RGB-D frames, 
$\mathcal{H}_{\triangle}$
and $\mathcal{D}$ in our implicit neural map $\Theta$ need to be carefully optimized during the reconstruction ($\rho$ is a constant projection). Therefore, proper supervision for learning and formulating a good loss function
are
critical here. 

Following ~\cite{sucar2021imap,zhu2022nice,wang2023co}, we adopt the input RGB-D frames with the estimated poses as the approximation of ground-truth modeling to provide the supervision in a projection-and-measure fashion. Volume rendering is leveraged to output the re-projected RGB-D images based on the weights $w(p)$ of points on the casting ray from each image pixel. Similar to~\cite{azinovic2022neural}, the weights $w(p)$ are obtained by the Sigmoid functions $\sigma$ and $\mathcal{O}^{\text{TSDF}}$ as shown in Eq. \ref{eq:weight}. 

Given an image pixel $x$, the TSDF truncation threshold $tr$, and the corresponding $N_r$ sampled points $\mathcal{P}$ on its casting rays with the sampled depth values $d^{\prime}$, the rendered RGB values $\mathbf{c}(x)$ and depth values $\mathbf{d}(x)$ can be obtained as Eq. \ref{eq:volume-rendering} \fix{and Eq. \ref{eq:volume-rendering-depth}}.

\begin{equation}
\label{eq:weight}
  w(p)=\sigma\left(\frac{\mathcal{O}^{\text{TSDF}}(p)}{tr}\right)\cdot\sigma\left(-\frac{\mathcal{O}^{\text{TSDF}}(p)}{tr}\right).
\end{equation}

\begin{align}
\label{eq:volume-rendering}
  \mathbf{c}(x)&=\frac{1}{\sum_{p\in\mathcal{P}}w(p)}\sum_{p\in\mathcal{P}}w(p)\mathcal{O}^{\text{RGB}}(p), \\
  \label{eq:volume-rendering-depth} \mathbf{d}(x)&=\frac{1}{\sum_{p\in\mathcal{P}}w(p)}\sum_{p\in\mathcal{P}}w(p)d^{\prime}(p).
\end{align}

One more thing, due to the resolution of $\mathcal{V}_{\text{coarse}}$ being low, the truncation threshold $tr_c$ for it needs to be larger than the one $tr_i$ adopted in the residual component $\mathcal{D}$. To align these two modules, the explicit 
base
$\beta(p)$ adopted in TSDF component of Eq.~\ref{eq:Op_final} needs to be specified as:

\begin{equation}
  \label{eq:tsdf-align}
  \hat{\beta}(p)=\Upsilon (\frac{\beta(p)\cdot tr_{c}}{tr_i},\tau_c).
\end{equation}

The $\Upsilon$ is the \textit{clamp} function which ensures $\hat{\beta}(p)$ to be in the range of $[-1,1]$ with threshold $\tau_c$. The aligned explicit base $\hat{\beta}(p)$ is only adopted for the TSDF calculation, while $\beta(p)$ with the direct trilinear interpolation is proper for the RGB prediction.

\begin{figure}[!t]
  \centering
\includegraphics[width=8.5cm]{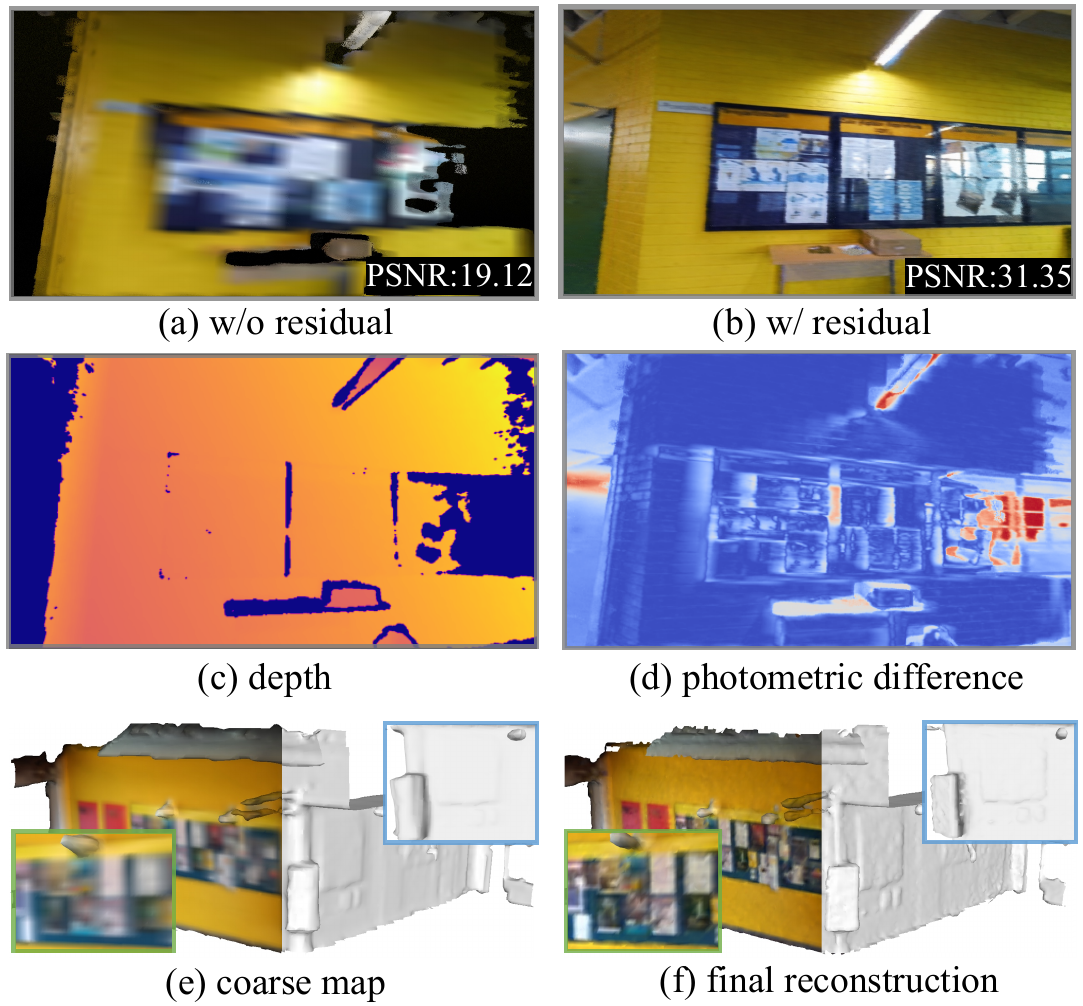}
\caption{Comparison of the 2D rendering results on \texttt{corridor} and 3D mesh on \texttt{waiting} of \texttt{BS3D} about whether to use the residual based on the coarse geometry. (a)-(b) The rendered RGB image of the global explicit coarse map and our residual-based mixed representation. (c) Ground-truth depth image. (d) Visualization of the photometric difference between (a) and (b) indicates that our residual module can not only fill the empty holes but also learn high-frequency information like the pictures and cracks on the wall. The lighter parts denote the larger residuals. (e) Coarse map using TSDF Fusion. (f) Reconstruction with residuals based on (e).}
\Description{Figure 3. Fully described in the text.}
\label{fig:methods-highfreq}       
\end{figure}

For the running efficiency, \fix{we} sample $N_s$ pixels $X$ in each iteration for neural learning. More details about sampling strategies can be found in Section~\ref{sec:system}. 
We use the observed RGB $\hat{c}$ and depth images $\hat{d}$ for supervision. The photometric and geometric losses are as follows:
\begin{equation}
  \label{eq:render_loss}
  \mathcal{L}_p=\frac{1}{N_s} \sum_{x \in X}({\mathbf{c}}(x)-\hat{c}(x))^2, \quad \mathcal{L}_g=\frac{1}{N_s} \sum_{x \in X}({\mathbf{d}}(x)-\hat{d}(x))^2 .
  \end{equation}

\begin{figure}[!t]
  \centering
  \includegraphics[width=8.5cm]{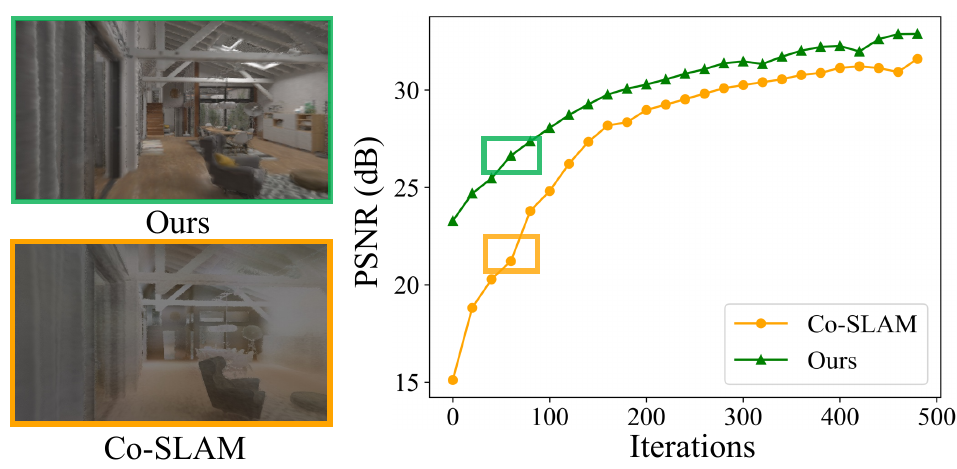}
  \caption{
  Comparison of the initial mapping of \texttt{apartment} on \texttt{uHumans2}~\cite{Rosinol20icra-Kimera} dataset.
  Based on the initial explicit map, our residual-based mixed representations can learn faster and present more high-fidelity rendering than Co-SLAM~\cite{wang2023co}.}
  \Description{RemixFusion can reconstruct the scene faster than Co-SLAM, demonstrating the efficiency of the proposed representations.}
  \label{fig:methods-speed-compare}   
\end{figure}
  
Following~\cite{wang2023co,johari2023eslam}, we leverage an approximate TSDF loss $\mathcal{L}_{tsdf}$ for points within the truncation ($|d^{\prime}(p)-\hat{d}(x)|<tr$) and a free-space loss $\mathcal{L}_{f s}$ for points $p$ in front of the surface ($\hat{d}(x) - d^{\prime}(p)>tr$), where $d^{\prime}(p)$ refers to the sampled depth values on the rays. These two designs are aimed at recovering the accurate and detailed geometry using depth observations as approximate ground-truth TSDF supervision. 
\begin{equation}
  \label{eq:8}
  \mathcal{L}_{t s d f}=\frac{1}{N_s} \sum_{x \in X} \frac{1}{N_{r}^\text{tr}} \sum_{p \in \mathcal{P}_\text{tr}}\left(s(x)-\mathcal{O}^{\text{TSDF}}(p)\right)^2 ,
  \end{equation}
\begin{equation}
  \label{eq:9}
  \mathcal{L}_{f s}=\frac{1}{N_s} \sum_{x \in X} \frac{1}{N_r^\text{fs}} \sum_{p \in \mathcal{P}_\text{fs}}\left(\mathcal{O}^{\text{TSDF}}(p)-1\right)^2,
  \end{equation}
where $s(x)$ is the approximate ground-truth TSDF value for the pixel $x$, $\mathcal{P}_\text{tr}$ and $\mathcal{P}_\text{fs}$ ($N_{r}^\text{tr}+N_r^\text{fs}=N_r$) denotes the sampled points allocated within and farther than a pre-defined distance threshold $tr$ to the observed approximate surface $\hat{d}$.

Besides these loss functions, a smoothness loss $\mathcal{L}_{smo}$ is applied to ease the hash collisions, avoiding the noisy points in empty space and make the predictions smooth.
\begin{equation}
  \mathcal{L}_{smo}=\frac{1}{|\mathcal{Q} |}\sum_{\mathbf{x}\in\mathcal{Q}}\Delta_x^2+\Delta_y^2+\Delta_z^2,
\end{equation}
where $\mathcal{Q}$ is the randomly sampled vertices on the hash grids and $\Delta_{x,y,z}^2$ means the difference of hash features between the adjacent vertices along these three axes.

In general, the mixed mapping $\mathcal{O}$ is carried out with $N_m$ iterations every $K$ frames, and the loss function under observation $\mathcal{X}$ and its corresponding poses $\mathcal{G}$ is defined as Eq. \ref{eq:total loss}. $(\lambda_p,\lambda_g,\lambda_t,\lambda_f,\lambda_s)$ are the corresponding weights for each loss component. 

\begin{equation}
  \label{eq:total loss}
\mathcal{L}(\mathcal{O}|\mathcal{X},\mathcal{G})=\lambda_p \mathcal{L}_p+\lambda_g\mathcal{L}_g+\lambda_t\mathcal{L}_{tsdf}+\lambda_f\mathcal{L}_{fs}+\lambda_s\mathcal{L}_{smo}.
\end{equation}

Figure \ref{fig:methods-highfreq} illustrates that the mixed representations focus on the high-frequency details and completion based on the coarse representations. Moreover, 
Figure~\ref{fig:methods-speed-compare} illustrates the better efficiency of RemixFusion in reconstruction. We, in essence, achieve a better trade-off by the proposed residual-based representation, which takes advantage of both explicit and implicit representations.

\subsection{Residual-based Bundle Adjustment}
\label{sec:ba}

As mentioned above, the residual-based mapping is based on the estimated camera poses, which are given by online camera tracking. For the best efficiency, the pose estimation can be obtained through an existing randomized optimization-based method~\cite{zhang2021rosefusion} for each RGB-D frame. Different from the gradient-based optimization used in \cite{bylow2013real}, the randomized optimization is more robust even if the optimization is of high nonlinearity. This method aims to minimize the distance between the transformed frame depth based on the estimated pose and the zero-crossing surface geometry of the reconstruction, which forms a frame-to-model optimization. Details of this method can be found in the supplementary materials.

However, there is still inevitable accumulated drift in front-end estimated camera poses, which addresses the necessity of multi-view bundle adjustment (BA), especially for large-scale scenes.

\paragraph{Bundle adjustment for residual pose} 
In contrast to the reconstruction, the bundle adjustment optimizes the camera poses with the neural implicit map fixed. Generally, the objective loss functions are the same as Eq. \ref{eq:total loss} based on $\mathcal{O}(p)$ obtained by Eq. \ref{eq:Op_final} for sampled points $p\in\mathcal{P}$. The difference is that we further optimize the camera poses $\mathcal{G}$ for the observed RGB-D frames while the mixed mapping $\mathcal{O}$ is fixed. The goal of bundle adjustment is formulated as:

\begin{equation}
  \label{eq:ba-formulation}
\arg\min_{\mathcal{G}} \mathcal{L}(\mathcal{G}|\mathcal{X}, \mathcal{O}).
  \end{equation}

Bundle adjustment (BA) in previous neural SLAM and traditional alternatives are trying to optimize $\mathcal{G}_i\in\mathcal{G}$ for each frame directly based on the back-propagation of the gradients given by Eq.~\ref{eq:ba-formulation}, which makes the optimization for each frame relatively independent. Lack of global awareness among independent optimization of $\mathcal{G}_i$ for each frame may lead to conflicts under the multi-view consistency constraints. These conflicts would limit the magnitude of optimization in BA to promise consistency. Therefore, we propose to adopt a single MLP $\mathcal{M}_p$ encoding individual poses $\mathcal{G}_i$ to be optimized for better global awareness in BA.

\begin{equation}
  \label{eq:M_p}
\mathcal{{G}}_i = \mathcal{M}_p(i).
\end{equation}

However, using the MLP $\mathcal{M}_p$ to encode the complete camera poses $\mathcal{G}_i$ only based on the frame index is inefficient and difficult for training (Eq. \ref{eq:M_p}). 
Inspired by the residual-based mixed representation (Section \ref{sec:mapping}) and \cite{bian2024porf}, we propose to change the output of $\mathcal{M}_p$ from the complete pose to only the residual pose, aiming to reduce network learning complexity and thereby enhance training efficiency.
Specifically, we encode the 
pose changes
$\mathcal{G}_\vartriangle = (r_x^\vartriangle ,r_y^\vartriangle,r_z^\vartriangle,t_x^\vartriangle,t_y^\vartriangle,t_z^\vartriangle)$ along with the frame index $i$ in $\mathcal{M}_p$.
Note that the frame index $i$ should be normalized to $[-1,1]$ according to the number of keyframes for better convergence. The residuals $\mathcal{G}_\vartriangle$ are added back to the initial poses $\mathcal{G}_i$ to gain the final results $\mathcal{\hat{G}}_i$. Compared to the independent optimization of each keyframe, the MLP $\mathcal{M}_p$ is more comprehensive, and the residual-based BA is more globally aware and efficient. The residual-based BA can be formulated as Eq. \ref{eq:res-ba}. The $\mathcal{G}$ and $\mathcal{G}_\vartriangle$ correspond to $\mathcal{F}_c$ and $\mathcal{F}_{\triangle}$ in Eq. \ref{eq:1} respectively. The residual-based BA is running with $N_b$ iterations every $K$ frames. $\mathcal{N}$ denotes the normalization of the frame index. 

\begin{equation}
  \label{eq:res-ba}
  \mathcal{\hat{G}}_i = \mathcal{G}_i + \mathcal{{G}}_\vartriangle,\quad \mathcal{{G}}_\vartriangle = \mathcal{M}_p(\mathcal{N}(i),\mathcal{G}_i).
\end{equation}

\paragraph{\fix{Adaptive gradient amplification of the BA module}}
Although the residual-based BA is global-aware and efficient, 
there is still room for improvement to BA in large-scale scenes.
Due to the real-time and memory requirements, the sampling has to be sparse.
Meanwhile, multi-view constraints of BA are applied only to a subset of all pixels, which makes the optimization of Eq.~\ref{eq:ba-formulation} easy to get stuck in the local optima, especially in large-scale scenes. 

\begin{figure}[!t]
  \centering
\includegraphics[width=8.5cm]{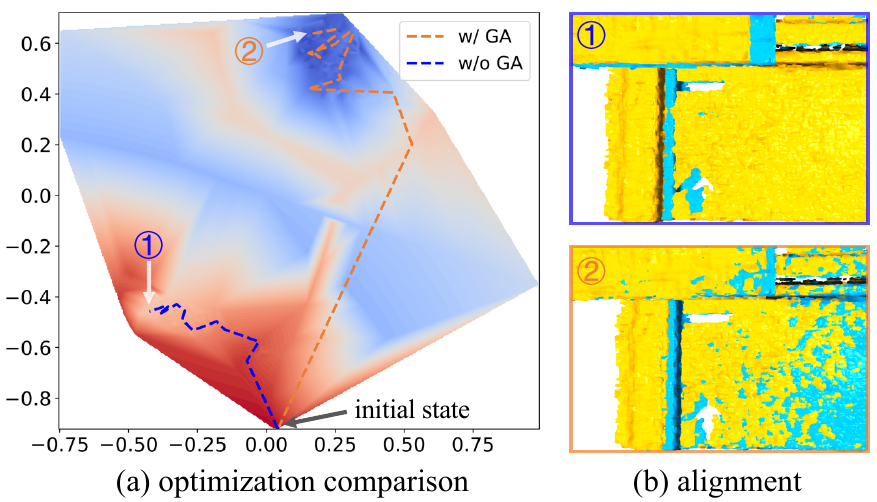}
\caption{Visualization of the 
adaptive gradient amplification (GA). (a) Comparison of whether to use GA for BA on \texttt{foobar} of \texttt{BS3D}. The states denote the 6D poses of all keyframes, colorized by the errors of pose estimation from large to small, which correspond to red and blue. The proposed GA helps BA converge to the global minima rather than being stuck in the local minima. (b) The optimized poses in (a) are visualized by the alignment of reconstructed surfaces with ground-truth surfaces. }
\Description{The proposed optimization momentum improves the performance of bundle adjustment, which is validated by the aligned reconstruction.}
\label{fig:vis-PM2}       
\end{figure}

Inspired by the Simulated Annealing algorithm (SA), gradient amplification (GA) seems like a good choice to overcome this problem.
The key idea is to amplify the optimization gradients near the reconstructed surface and encourage the BA optimization to explore more. The simplest way to amplify the gradients is to move the cameras in the direction they are facing or in a random direction by a certain distance.
This would lead to misalignment of the currently observed surfaces and reconstructed surfaces.  
The lack of consideration of the 3D geometric structures makes the BA module hard to achieve the expected results due to gradient amplification disorder.

We propose an adaptive gradient amplification technique
based on the reconstruction geometry, which amplifies the optimization gradients of the BA process near the zero-crossing surface, further improving the accuracy of pose estimation.
Specifically, we change Eq. \ref{eq:tsdf-align} to Eq. \ref{eq:tsdf-align-k-new} with $\tau_c$ multiplied by $k (k>1)$ only in BA, which results in the imbalance of the predicted and approximate ground-truth TSDF values, which should both be equal to 1 in the free space. In this way, the predicted TSDF absolute values $\mathcal{O}^{\text{TSDF}}(p)$ are clamped with the threshold $k\cdot \tau_c$ (>1). According to Eq.~\ref{eq:9}, the loss $\mathcal{L}_{f s}$ and $\mathcal{L}_{t s d f}$ are amplified to drive the observed surfaces to get close to the zero-crossing surface. 
The residual-based BA would then try to minimize the amplified losses and achieve a more multi-view consistent balance.

\begin{equation}
  \label{eq:tsdf-align-k-new}
  \hat{\beta}(p)=\Upsilon (\frac{\beta(p)\cdot tr_{c}}{tr_i}, k \cdot\tau_c).
\end{equation}

Figure~\ref{fig:vis-PM2}(a) visualizes the optimization process of the bundle adjustment on the scene \texttt{foobar} of \texttt{BS3D}. The states, colorized by the errors of pose estimation, are 6D poses of all keyframes, which are visualized in 2D. The optimization starts with the poses of front-end tracking. Red denotes higher errors, while blue denotes lower errors. BA with the proposed GA converges fast to the global minima, whereas BA without GA tends to be stuck in local minima and yields only marginal improvement. The visualization is shown by the alignment of point clouds generated with the optimized poses and ground-truth poses, which correspond to the yellow and blue ones in 
Figure~\ref{fig:vis-PM2}(b). Overall, the proposed GA helps the bundle adjustment in large-scale scenes explore more and converge to the global minima. Extensive experiments show that the residual-based BA is efficient and robust, which can steadily refine the initially estimated camera poses with more global comprehension.

\begin{algorithm}[!t]
  \begin{algorithmic}
      \STATE \textbf{Input:} Input RGB-D images $I_*^t$, threshold $\tau_v$ and anchor pose $\varphi_a$.
      \STATE \textbf{Output:} Updated anchor pose $\varphi_a$ and updated volume $\mathcal{V}_a$.
      \STATE \textbf{Initialize:} Current anchor volume $\mathcal{V}_a$, $\varphi_t=\varphi_{t-1}$.
      \STATE $\varphi_*\leftarrow$ \texttt{Randomized Optimization}$(\varphi_t, I_*^t)$.  
      \STATE $\mathcal{V}_a\leftarrow$\texttt{Integration}($I_*^t$, $\varphi_*$).
      \FORALL{$i\in \lbrace x,y,z\rbrace$}
      \STATE $d = |\varphi_*(i)-\varphi_a(i)|$;
      \IF{$d>\tau_v$} 
      \STATE create volume $\mathcal{V}_t$ with default values and swap the overlap between $V_t$ and $\mathcal{V}_a$.
      \STATE $\mathcal{V}_a\leftarrow \mathcal{V}_t$, $\varphi_a\leftarrow\varphi_*$.

      \ENDIF 
      \ENDFOR

      \STATE Return the updated anchor pose $\varphi_a$ and volume $\mathcal{V}_a$.
  \end{algorithmic}
  \caption{Scalable volume management
  }\label{algorithm1}
\end{algorithm}

\subsection{System Implementation}
\label{sec:system}
In this section, we illustrate the proposed scalable pose estimation for large-scale scenes using randomized optimization and system implementation, including sampling and keyframe selection.

\paragraph{Scalable Randomized Optimization}
First, we review the core insight of randomized optimization used in ~\cite{zhang2021rosefusion}, as well as the basic principles and cost functions of pose estimation. In brief, the goal is to provide accurate 6DoF camera poses $[\mathbf{R}\mid\mathbf{t}]\in SE(3)$ for $N$ live RGB-D frames $I_*^t = \{I_c^t,I_d^t\}_{t=0:N}$. We first back-project the $(i,j)$ pixel with depth $I_d^t(i,j)$ to its 3D position $x_{ij}$ in the camera space, its point-to-surface distance in the world space could then be queried using $\phi$ and the transformation $\mathbf{X}_{ij}=\mathbf{R}\mathbf{x}_{ij} + \mathbf{t}$.
\fix{$\phi$ defines the difference between the 
queried TSDF values from the moving volume
and the approximated ground-truth TSDF supervision, which is similar to Eq. \ref{eq:8} and Eq. \ref{eq:9}.}
The minimization of these point-to-surface distances described in Eq. \ref{eq:3} is tackled by the randomized optimization with $N_{ro}$ iterations using a pre-sampled particle swarm template (PST). We encourage readers to refer to supplementary materials for more details. 

\begin{equation}
  \label{eq:3}
  (\mathbf{R}^*,\mathbf{t}^*)=\arg\min_{R,t}\sum_{(i,j)\in I_d}\phi(\mathbf{R}\mathbf{x}_{ij}+t)^2,
  \end{equation}
  where $I_d$ denotes the depth image and $\phi:\mathbb{R}^3\to\mathbb{R}$ means the query of the reconstructed TSDF volume.

One main difference of our method against ROSEFusion~\cite{zhang2021rosefusion} is that we leverage a moving volume to perform pose estimation to make our tracking module scalable to large-scale scenes. This is inspired by the success of the moving TSDF-Fusion strategy~\cite{Whelan2012,Roth2012}. The process of the management of the moving explicit volume and pose estimation based on randomized optimization is illustrated in Algorithm~\ref{algorithm1}. Given the first camera position $\varphi_a$ and the input RGB-D frames $I_*^t$, we initialize the first moving volume $\mathcal{V}_a$. Only the parts observed within the volume $\mathcal{V}_a$ are integrated. The camera pose would be marked as an anchor $\varphi_a$ each time the volume is moved. We transform the moving volume $\mathcal{V}_a$ only when the Euclidean distance between the current camera $\varphi_t$ and the last anchor camera $\varphi_a$ exceeds a threshold $\tau_v$ in any direction of the 3 axes. We create a new moving volume $\mathcal{V}_{t}$, inheriting the overlapping regions from $\mathcal{V}_{a}$ to minimize the transferring costs, shown in Eq. \ref{eq:mov_volume}. 
\begin{equation}
\label{eq:mov_volume}
  \mathcal{V}_{t}(x, y, z)= \begin{cases}\mathcal{V}_a(x, y, z) & \text { if }(x, y, z) \in \mathcal{V}_t \\ \mathcal{V}_0(x, y, z) & \text { if }(x, y, z) \notin \mathcal{V}_t\end{cases} ,
  \end{equation}
where $(x, y, z)$ refers to the world coordinate and $\mathcal{V}_0$ is the default value of the TSDF volume. Here, we set its TSDF value to one and the other attributes, including color and weights, to zero.

After swapping, $\mathcal{V}_{t}$ is marked as the new anchor volume $\mathcal{V}_{a}$. In our implementation, we use a dense axis-aligned moving TSDF volume $\mathcal{V}_a$ with resolution 
higher than that of $\mathcal{V}_{\text{coarse}}$. Our system can accurately perceive local environments with a fixed and moderate memory cost to maintain efficiency.

\paragraph{Sampling and Keyframe Selection} Following Co-SLAM~\cite{wang2023co}, we downsample each keyframe and only save 5$\%$ of the pixels in the keyframe database that contains the RGB, depth, and pose data. The mapping process is continuously running every $K$ frame, and we sample $N_s$ pixels from the keyframe database each time. After sampling, we can back-project the pixels to 3D space, and sample $N_r$ points along the rays of different pixels by ray casting, including uniformly sampled points and points sampled near the observed surface. Then, we can get the points $\mathbf{x}_{ij}=\mathbf{o}+d_j \mathbf{r_{i}}, i\in \{1,\ldots N_s\}, j\in \{1,\ldots N_r\} $, which are $N_s\times N_r$ points in total to be queried and learned.

%% file: 06-experiments.tex
\section{EXPERIMENTS}

\subsection{Experimental Setup}
\label{Experimental dataset}
\paragraph{Dataset}
We evaluate methods on two large-scale indoor datasets \texttt{BS3D}~\cite{mustaniemi2023bs3d} and \texttt{uHumans2}~\cite{Rosinol20icra-Kimera}. We also provide the comparison for the room-level publicly available dataset: \texttt{Replica}~\cite{straub2019replica}, \texttt{TUM RGB-D}~\cite{sturm2012benchmark}, \texttt{ScanNet}~\cite{dai2017scannet}, and \texttt{FastCaMo}~\cite{zhang2021rosefusion}. This evaluation aims to benchmark our tracking and reconstruction performance against current state-of-the-art methods. \texttt{BS3D} is a challenging real-world large-scale RGB-D dataset, comprising over 10,000 RGB-D images annotated with ground-truth trajectories from a motion capture system. \fix{\texttt{uHumans2} is a large-scale publicly available synthetic dataset with multiple rooms and complicated layouts, presenting a significant challenge for SLAM.} 
We evaluate eight large-scale sequences of \texttt{BS3D}, two sequences of \texttt{uHumans2}, alongside commonly referenced sequences from \texttt{Replica}, \texttt{ScanNet}, \texttt{TUM RGB-D}, and \texttt{FastCaMo-Synth} (noise-free).
Furthermore, additional qualitative comparisons are conducted on \texttt{BS3D}, \fix{\texttt{uHumans2}}, \texttt{FastCaMo-Large} proposed in MIPS-Fusion \cite{tang2023mips}, and self-captured sequences to highlight the different performance on large-scale scenes.

\paragraph{Metrics}
In terms of 3D reconstruction quality, we follow~\cite{zhu2022nice} and adopt ~\emph{Accuracy(cm), Completion(cm)}, and \emph{Completion ratio(\%)} at the threshold of 5cm and 10cm, as well as the F1-score. Following~\cite{azinovic2022neural,zhu2022nice}, we filter the noisy points that are not visible in any observation with ground-truth depth images for a fair comparison, as many neural approaches tend to predict numerous noisy points in empty spaces. Then we use the \emph{Iterative Closest Point (ICP)} for alignment with ground-truth meshes. Metrics like PSNR, SSIM, LPIPS, and Depth-L1 (m) for 2D rendering comparison are adopted following~\cite{keetha2024splatam}. As for camera tracking, the ATE RMSE~\cite{sturm2012benchmark} is adopted. The system FPS (frames per second) and GPU memory usage are considered for efficiency comparison.

\paragraph{Baselines}
We compare our method to both the implicit and explicit state-of-the-art methods. The former includes NeRF-based methods like iMAP~\cite{sucar2021imap}, NICE-SLAM~\cite{zhu2022nice}, Co-SLAM~\cite{wang2023co}, ESLAM~\cite{johari2023eslam} and MIPS-Fusion~\cite{tang2023mips}. The latter indicates ElasticFusion~\cite{Whelan2015},  BundleFusion~\cite{Dai2017} and BAD-SLAM~\cite{schops2019bad}. \ffix{RTG-SLAM~\cite{peng2024rtgslam}, GS-ICP SLAM~\cite{ha2024rgbd}, SplaTAM~\cite{keetha2024splatam}, MonoGS~\cite{matsuki2024gaussian}, Photo-SLAM~\cite{huang2024photo} and LoopSplat~\cite{zhu2025_loopsplat} are encompassed as the representative of the SOTA 3DGS-based SLAM approaches.} For more comprehensive comparisons, the sparse SLAM ORB-SLAM3~\cite{campos2021orb} \ffix{and DROID-SLAM~\cite{teed2021droid} that specializes in robust camera tracking are included}.
The implementation of iMAP* is from NICE-SLAM. \fix{ESLAM, SplaTAM, and MonoGS are evaluated using images at half resolution as inputs on \texttt{BS3D} and \texttt{uHumans2} datasets due to out-of-memory issues. For RTG-SLAM, a modified version was employed due to its high GPU memory allocation in global optimization on these two datasets. More details can be found in the supplementary materials.}

\begin{table*}[!t]\centering
  \renewcommand{\arraystretch}{1.1}
\caption{
Comparing tracking accuracy (ATE RMSE in cm) on 8 large-scale RGB-D sequences of \texttt{BS3D}. The first 5 methods are based on learnable implicit parameters, and the rest are traditional explicit methods, 3DGS-based methods, and a sparse explicit method (ORB-SLAM3). {RemixFusion-lite} denotes the {lightweight} version of our method. `--' denotes that the tracking failed for corresponding methods (error $>$ 500cm), and `$\_$' denotes the {second-best} method. 
}
\scalebox{0.95}{
\setlength{\tabcolsep}{3.8mm}{
\begin{tabular}{l l|c c c c c c c c | c}
    \toprule 
        &Methods  & {cafeteria} & {corridor}  & {foobar} & {hub} & {juice} & {lounge} & {study} & {waiting} & Avg. \\ \hline
        \multirow{5}{*}{\rotatebox[origin=c]{90}{Implicit}}  & iMAP$^{*}$ & \oout{534.8}& 255.4&295.8&184.6&64.7&299.0&318.5&143.6&262.1 \\ 
        &NICE-SLAM  & 52.9&8.7&13.1&12.1&4.9&11.0&7.5&6.6&14.6 \\ 
        &Co-SLAM & 127.6 &11.5 &106.5 &6.2 &5.2 &48.8 &6.5 &19.8 & 41.5 \\ 
        &MIPS-Fusion  & 122.1&78.5&41.2&36.5&6.2&11.3&6.1&19.7&40.2 \\ 
        &\fix{ESLAM}  & {7.7}&{\textbf{5.7}}&{10.4}&{\textbf{4.3}}&{4.1}&{6.3}&{4.7}&{8.3}&{6.4} \\\midrule
        \multirow{10}{*}{\rotatebox[origin=c]{90}{Explicit}}&ElasticFusion & 193.1&230.6	&64.2&76.9&119.3&314.0&85.5&142.4&153.3 \\ 
        &BundleFusion & -- & --  & -- & -- & 10.0 & -- & -- & -- & -- \\  
        &BAD-SLAM & -- & 170.0  & 334.1 & -- & 22.4 & 8.3 & 4.7 & -- & -- \\ 
        &\fix{SplaTAM} & \oout{948.2}&\oout{1078.8}&{168.1}&{260.3}&{31.2}&\oout{621.7}&{372.5}&{63.5}&\oout{443.0}\\
        &\fix{MonoGS} & {448.4}&\oout{504.8}&{491.2}&{207.1}&{182.8}&{317.2}&{211.0}&{211.4}&\oout{321.7} \\  
        &\ffix{LoopSplat} & \oout{954.7}&\oout{1617.7}&31.5&\oout{525.8}&\oout{503.3}&149.1&\oout{584.0}&	\oout{1091.7}&\oout{800.9} \\
        &\ffix{Photo-SLAM} & 15.4&23.2&26.6&7.6&\un{4.0}&	24.6&6.0&6.5&14.3\\
        &\fix{RTG-SLAM} & {21.9}&{11.6}&{11.1}&{8.4}&{9.9}&{12.0}&{6.7}&{15.6}&{12.2}\\ 
        &GS-ICP SLAM & 35.1&23.9&19.9&9.2&6.3&24.0&5.8&7.3&16.4\\  
        &ORB-SLAM3 & \textbf{5.6}&{6.7}&7.5&6.1&6.0&15.8&5.4&\textbf{2.9}&7.0 \\ \hline 
        &RemixFusion &\un{6.8} & \un{6.3} & \textbf{5.8} & \un{4.5} & \textbf{3.1} & \textbf{4.2} & \textbf{3.2} & \un{3.0} & \textbf{4.6} \\ 
        &RemixFusion-lite &10.1 & 11.0 & \un{6.8} & {5.0} & {4.1} & \un{4.6} & \un{4.3} & 5.1 & \un{6.4} \\ 
        
    \bottomrule 
    
\end{tabular}
}}
\label{tab:BS3D-tracking}
\end{table*}

\paragraph{Implementation Details}
We evaluate RemixFusion on a desktop PC equipped with a 3.90GHz Intel Core i9-14900K CPU and an NVIDIA RTX 3090 Ti GPU. In terms of camera tracking, we use $N_{\text{ro}}=20$ iterations per frame for the randomized optimization. We implement the scalable randomized optimization based on the PyCUDA~\cite{kloeckner_pycuda_2012} libraries for acceleration. The tracking process is parallel to mapping. For Mapping, we sample $N_s=2048$ pixels and $N_r=59$ points along each ray, including 11 for uniform samples and 48 for samples near the surface. The resolution is 200 for $\mathcal{V}_{\text{coarse}}$ and $14m\times14m\times6m$ for $\mathcal{V}_a$ in BS3D. The threshold used in residual-based BA is $\tau_c = 1$ and $k=2$. The mapping and BA process are iteratively performed with $N_m=N_b=5$ iterations. More details can be found in the supplementary materials.

\subsection{Quantitative and Qualitative Comparison} 
In this section, we present quantitative and qualitative results, including the camera pose estimation, 3D mesh reconstruction, and 2D rendering on different datasets. \fix{The lightweight version of our method, requiring fewer optimization iterations, is denoted as RemixFusion-lite. Details can be found in the supplementary materials.}

\begin{table}[!t]
  \renewcommand{\arraystretch}{1.25}
  \centering
  \caption{Comparison of reconstruction accuracy (Acc.), completeness (Comp.), completeness Ratio (\%) (Comp. Ratio(\%)) with 5cm threshold and frames per second (FPS) of mapping using ground-truth camera poses on 8 scenes of the \texttt{BS3D} dataset. 
  {'$\_$' denotes the second-best method. 
  }}
  \scalebox{0.95}{
  \setlength{\tabcolsep}{2.0mm}{
  \begin{tabular}{@{}l ccc c}
  \toprule
  \multirow{2}{*}{Methods} & \multicolumn{3}{c}{Metrics}   & \multirow{2}{*}{FPS}  \\ \cline{2-4}
  
  & Acc.$\downarrow$ & Comp.$\downarrow$ & Comp. Ratio(\%)$\uparrow$&   \\ \hline  
  \fix{Co-SLAM} & {\un{4.57}} & {\un{3.91}}&  {85.30}& {20} \\
  MIPS-Fusion & 6.93 & 16.78 & 60.15 & 9\\
  \fix{ESLAM} & {5.92}&{8.40}&{\un{85.73}} & {5}\\
  \fix{RTG-SLAM} & {6.71} & {8.28} & {63.68} & {5}\\
  GS-ICP SLAM & 9.09 &  8.37 & 53.44 & 26\\
  RemixFusion &\textbf{4.34} &  \textbf{3.56} & \textbf{86.88} & \un{27}\\
  RemixFusion-lite &{4.70} & {4.06} & {83.45} & \textbf{94}\\
  \bottomrule 

\end{tabular}
}}
\label{tab:BS3D-gt}
\end{table}

\subsubsection{Real-time Reconstruction Regardless of Tracking.}

Table~\ref{tab:BS3D-gt} is the result of 3D reconstruction using ground-truth poses to eliminate the tracking effects in SLAM. 
Thanks to the residual-based mixed representation, our method is the most competitive one in accuracy (4.34cm) and completeness (3.56cm), surpassing the state-of-the-art by 5$\%$ in accuracy and 9$\%$ in completeness. The second-best method, after our own, is Co-SLAM, which achieves the accuracy of 4.57cm and the completeness of 3.91cm. While the completeness ratio of ESLAM is the second-best for implicit methods, its completeness and accuracy remain suboptimal, indicating the incomplete reconstruction of its tri-plane representation. 

Notably, with ground-truth poses, the best 3DGS-based method demonstrates inferior reconstruction performance (46.8$\%$ worse than the best implicit method in accuracy) with the online setting.
RTG-SLAM leverages an efficient and compact 3DGS representation to save memory, but the reconstruction performance is not desirable, indicating the trade-off between efficiency and accuracy. Although GS-ICP SLAM has a comparable mapping FPS to our method, its reconstruction accuracy is 4.75cm worse than ours. This indicates that achieving high-quality geometric reconstruction with 3DGS requires significantly more optimization iterations in real-world large-scale scenarios, which may lead to compromised 3D reconstruction under real-time SLAM constraints.
Note that the outputs of MIPS-Fusion are obtained by running the official code, and some objects are missing. This leads to worse performance of completeness and comparable performance of completeness ratio. The lightweight version of our method only performs mapping with a few optimization iterations in this experiment and is significantly faster, with only a slight decrease in reconstruction accuracy (-0.36cm) and completeness (-0.5cm), which further proves the effectiveness and robustness of our method. 

\begin{figure}[!t]
  \centering
\includegraphics[width=8.5cm]{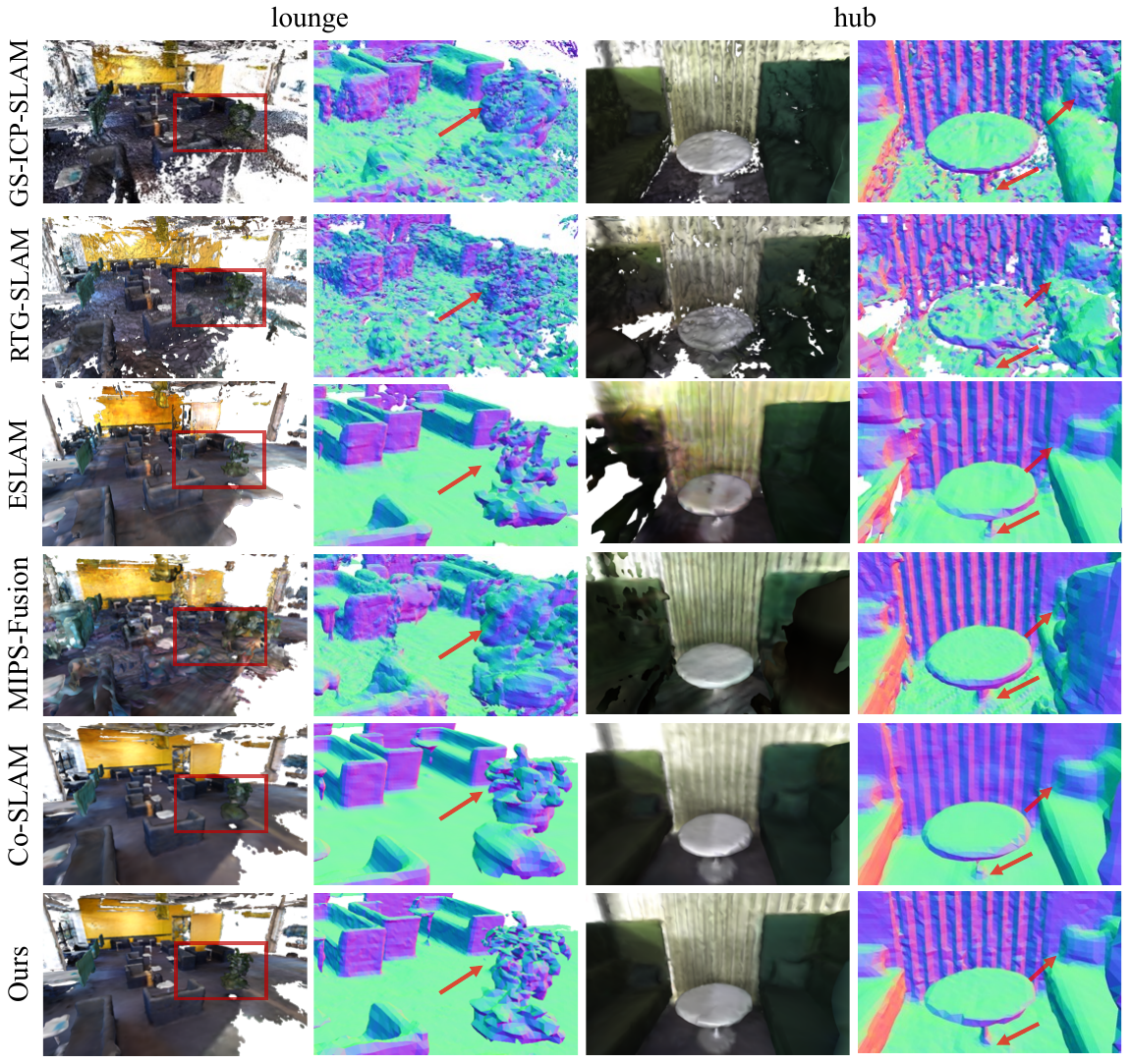}
\caption{Comparison of reconstruction using ground-truth camera poses. The first column of each scene is the overview, and the second column of each scene is the zoom-in comparison, colorized with normals. The detailed comparisons are pointed out with red arrows. Our method attains the best reconstruction details for the fine-grained objects in large-scale scenes.} 
\Description{The details, including the leaves of plants and pillows on the sofa, are more fine-grained for RemixFusion.}
\label{fig:gtmapping}       
\end{figure}

In Figure~\ref{fig:gtmapping}, we present the visualization of the reconstruction of different approaches. \fix{There are many artifacts for  RTG-SLAM and GS-ICP SLAM, primarily due to the discontinuity in rendered depths.} 
Our method excels in preserving detailed geometry, such as the leaves on the floor (the second column) and the pillows (the fourth column).
Additionally, the comparison of ours and Co-SLAM (the best approach except for ours in Table~\ref{tab:BS3D-gt}) with camera poses of RemixFusion can be found in the supplementary materials. The improvement compared to the implicit methods proves that the residual-based mixed representation enhances the reconstruction quality with limited time, which is crucial for SLAM.

\subsubsection{Pose Estimation.}

Table ~\ref{tab:BS3D-tracking} shows ATE RMSE (cm) on 8 sequences of \texttt{BS3D} for our method compared to 
five
state-of-the-art implicit methods and three explicit traditional methods.
Evaluations of the state-of-the-art 3DGS-based methods (SplaTAM, MonoGS, RTG-SLAM, GS-ICP SLAM, \ffix{LoopSplat and Photo-SLAM}) and the sparse SLAM (ORB-SLAM3) are also incorporated for a more comprehensive comparison for tracking accuracy.

Our method attains the best tracking performance (4.6cm) for 8 challenging sequences on average while running in real time. \fix{There is 28.1$\%$ improvement for our method in tracking on average compared to the SOTA methods.} Implicit methods struggle with pose estimation in these large-scale scenes. Specifically, implicit methods, except for NICE-SLAM and ESLAM, all struggle in tracking on the scene \texttt{cafeteria}, which is more than 400$m^2$. The layouts are complex, even including some textureless areas. These factors pose challenges for robust camera pose estimation and real-time reconstruction, resulting in unsatisfactory reconstruction quality, unstable pose estimation, and inadequate bundle adjustment for most methods. Although our method is inferior to the sparse SLAM ORB-SLAM3 on this challenging sequence, our method surpasses ESLAM by 11.7$\%$. For the scene \texttt{hub}, our method indicates a marginal performance gap of 0.2cm between our method and the best method.
\fix{Although NICE-SLAM and ESLAM demonstrate stable tracking performance on \texttt{BS3D}, they suffer from low FPS and high GPU memory consumption in large-scale scenes, as shown in Table~\ref{tab:runtime}.}
In contrast, our method maintains accuracy and robustness on these large-scale scenes.
RemixFusion-lite (over 25FPS), the lightweight version, is the second-best alternative on average, further proving the efficiency and robustness of our method.

\begin{table*}[h]
  \renewcommand{\arraystretch}{1.2}
  \centering
  \caption{\fix{Comparing tracking accuracy (ATE RMSE in cm) and rendering quality on 2 large-scale RGB-D sequences of \texttt{uHumans2}. `--' denotes that the tracking failed for the corresponding methods, and `$\_$' denotes the second-best method. The rendering is evaluated every 10 frames using the estimated camera poses.}
  }  
  \scalebox{0.88}{
  \setlength{\tabcolsep}{1.05mm}
  {
  \begin{tabular}{@{}l  ccccc |ccccc | ccccc}
  \toprule

  \multirow{2}{*}{Methods} & \multicolumn{5}{c|}{{office}} &\multicolumn{5}{c|}{{apartment}}   & \multicolumn{5}{c}{{Avg.}}  \\ \cmidrule{2-6}  \cmidrule{7-11} \cmidrule{11-16}    
  & {ATE RMSE$\downarrow$} & {PSNR$\uparrow$} & {SSIM$\uparrow$} & {LPIPS$\downarrow$}& {D-L1$\downarrow$} & {ATE RMSE$\downarrow$} & {PSNR$\uparrow$} & {SSIM$\uparrow$} & {LPIPS$\downarrow$} & {D-L1$\downarrow$}& {ATE RMSE$\downarrow$} & {PSNR$\uparrow$} & {SSIM$\uparrow$} & {LPIPS$\downarrow$} & {D-L1$\downarrow$} \\ \hline  

  \ffix{DROID-SLAM} & 6.84 & --	& --& --& -- & \un{3.98} & --&--&	--&-- & 5.41 &--&--&--&--\\
  NICE-SLAM  & 14.49 &21.80&0.929&{0.395} &0.134 & 6.79 &25.35&0.976&0.345&0.086 & 10.64&23.58&0.953&0.370&0.110\\ 
  Co-SLAM & 1268.23 & 17.79 & 0.838 & 0.752 & 1.885 & 10.95 &26.08&0.977&0.412&0.171& 639.59 & 21.94&0.908&0.582&1.028\\ 
  MIPS-Fusion  & 56.46 & 19.99&0.888&0.586&0.989 & 16.07 & 22.29&0.953&0.510&0.669 & 36.27 &21.14&0.921&0.548&0.829 \\ 
  ESLAM & \un{6.53} & \un{24.32}	& \un{0.948}& \textbf{0.198}& \un{0.129} & {4.13} & \un{27.91}&\textbf{0.984}&	\un{0.170}&\textbf{0.065} & \un{5.33} &\un{26.12}&\un{0.966}&\textbf{0.184}&\un{0.097} \\
  \midrule

  ElasticFusion & 1314.37 & -- &  --&  --& -- & 116.42 & -- & -- & -- & -- & 715.40& -- & -- & --&  -- \\ 
  BAD-SLAM & -- & -- &  --&  --& -- & 19.54 & -- & -- & -- & -- & -- & -- & -- & --&  -- \\ 
  SplaTAM & 1930.69 &10.29&0.264&0.719&1.564 & 87.32 & 22.35&0.766&0.384&0.221 & 1009.01 & 16.32&0.515&0.552&0.893\\
  MonoGS& 26.21&20.63&0.653&0.534&1.939&6.92&\textbf{32.40}&0.927&\textbf{0.092}&0.270&16.57 &\textbf{26.52}&0.790&\un{0.313}&1.105 \\  
  \ffix{Photo-SLAM} & --  & -- & -- & -- & --  &  11.12 &23.58&0.961&0.364&1.371& -- & -- & -- & -- & --\\ 
  \ffix{LoopSplat} & 160.47  & 21.99 &0.925 &0.458 & 1.177  &  56.09 &24.99&0.969&	0.384&0.867& 108.28 & 23.49&0.947&0.421&1.022\\ 
  RTG-SLAM & 330.49 &16.24&0.768&0.618&2.405 & 90.23 &20.01&0.922&0.542&1.299 & 210.36&18.13&0.845&0.580&1.852\\ 
  GS-ICP SLAM & 4799.29 & 18.84&0.612&0.585&3.132 & 185.17 &23.33&0.784&0.411&2.571 & 2492.23 & 21.09&0.698&0.498&2.852 \\  \hline 
  RemixFusion & \textbf{5.44} & \textbf{24.66} &\textbf{0.951}&\un{0.368}&\textbf{0.092}& \textbf{3.93}& {27.46}&\un{0.983}&0.326&\un{0.077}&\textbf{4.69}&26.06&\textbf{0.967}&0.347&\textbf{0.085}\\
  \bottomrule 

\end{tabular}
}}
\label{tab:uhumans2-benchmark}
\end{table*}

Explicit dense RGB-D methods struggle with robustness in these challenging scenes. 
Among them, only ElasticFusion, RTG-SLAM, GS-ICP SLAM, \ffix{Photo-SLAM}, and ORB-SLAM3 consistently succeed in tracking on all scenes. While BundleFusion and BAD-SLAM encounter failures in certain cases, primarily due to front-end odometry issues, which are also validated in \cite{mustaniemi2023bs3d}.
Although ElasticFusion can finish tracking on these sequences, its trajectory proves inaccurate. 

\ffix{3DGS-based methods, including RTG-SLAM, GS-ICP SLAM, and Photo-SLAM, are comparable to NICE-SLAM in tracking. These methods leverage either multi-level ICP or ORB features~\cite{mur2017orb} for tracking, which is decoupled from 3DGS optimization. In contrast, other 3DGS-based methods, including SplaTAM, MonoGS, and LoopSplat, which leverage the rendering loss as the key objective function for camera pose optimization, exhibit unstable tracking performance on this large-scale dataset. Although they can attain high-fidelity rendering given good camera poses, it is challenging to optimize the 3DGS and camera poses simultaneously in large-scale scenes, since there are drastically more 3DGS to optimize. }
\ffix{Notably, LoopSplat achieves successful tracking on the \texttt{foobar} and \texttt{lounge} sequences but fails on the remaining sequences on \texttt{BS3D}. SplaTAM fails on 3 sequences, and MonoGS fails on 1 sequence.} In contrast, our method is robust and surpasses the best 3DGS-based method (RTG-SLAM) by 62.3$\%$ for tracking.

ORB-SLAM3, known for sparse SLAM, achieves robust and accurate pose estimation regardless of dense reconstruction. 
In comparison, our method achieves superior tracking accuracy across most scenes and emerges as the most accurate dense RGB-D SLAM system overall for tracking, surpassing even sparse SLAM systems like ORB-SLAM3 by 34.3$\%$. 
Furthermore, while ORB-SLAM3 focuses on sparse reconstruction, our method excels in dense reconstruction with more accurate pose estimation.

\begin{table*}[t]
  \renewcommand{\arraystretch}{1.1} 
  \centering
  \caption{Comparison of reconstruction accuracy (Acc.), completeness (Comp.), and completeness Ratio(\%) (Comp. Ratio(\%)) with 10cm threshold 
  on \texttt{BS3D}. 
  `--' denotes the failure for the corresponding methods and '$\_$' denotes the second-best method. 
  }
  \setlength{\tabcolsep}{3.3mm}
  \scalebox{0.93}{
  \begin{tabular}{@{}l|l|cccccccc|c@{}}
  \toprule
  {Methods}& Metrics & {cafeteria} & {corridor} & {foobar} & {hub} & {juice} & {lounge} & {study} & {waiting} & {Avg.}  \\
  \hline 
  \multirow{3}{*}{NICE-SLAM} & Acc.$\downarrow$ & 45.65	&20.78&24.81&21.68&10.27&23.35&20.11&18.97 & 23.20 \\
    & Comp.$\downarrow$ & 23.82&\un{6.36}&8.42&6.01&3.95&9.49&6.00&5.16 & 8.65 \\
    & Comp. Ratio(\%)$\uparrow$ & 27.84&\un{87.09}&71.61&87.48&96.60&64.09&82.33&94.16 & 76.40 \\ \hline

  \multirow{3}{*}{Co-SLAM} & Acc.$\downarrow$ & 19.39&9.66	&20.52&5.50&\textbf{4.47}&61.32&5.15&9.63 & 16.96 \\
  & Comp.$\downarrow$ & 34.55&7.13&12.02&5.52&{3.68}&41.85&5.12&5.72& 14.45 \\
  & Comp. Ratio(\%)$\uparrow$ & 45.44&79.17&67.99&91.94&\un{98.09}&20.72&86.57&	86.91 & 72.10\\ \hline

  \multirow{3}{*}{\fix{ESLAM}} & Acc.$\downarrow$ & \un{5.44}&\textbf{6.14}&11.16&\un{4.36}&6.07&6.40&4.05&5.02 & 6.08 \\
  & Comp.$\downarrow$ & 7.43&9.26&8.89&6.69&\un{3.38}&12.74&7.63&10.37& 8.30 \\
  & Comp. Ratio(\%)$\uparrow$ & \un{86.68}&83.86&80.29&91.52&97.19&72.85&86.33&83.01 & 85.22\\ \hline

  \multirow{3}{*}{MIPS-Fusion} & Acc.$\downarrow$ & 102.64	&31.89&	23.81&	13.21&	7.02&	9.42&	5.30&	16.35 & 26.21\\
  & Comp.$\downarrow$ & 108.69&19.82&22.83&19.58&5.35&25.14&13.13&27.52 & 30.26 \\
  & Comp. Ratio(\%)$\uparrow$ & 11.20&58.78&62.26&79.77&94.69&69.63&81.97&	59.62 & 64.74\\ \hline

  \multirow{3}{*}{BAD-SLAM} & Acc.$\downarrow$ & --&21.43&8.18&	-- &6.12&	\textbf{4.85}&\textbf{3.15}&--&--\\
  & Comp.$\downarrow$ & --&50.39&72.45&--&11.08&18.92&8.03&--&--\\
  & Comp. Ratio(\%)$\uparrow$ & --&11.46&30.84&--&71.03&64.33&85.97&--&-- \\ \hline

  \multirow{3}{*}{\fix{SplaTAM}} & Acc.$\downarrow$ & 180.75&138.67&48.69&43.66&10.09&58.38&117.72&23.97&77.74\\
  & Comp.$\downarrow$ & 145.37&172.34&52.57&64.77&9.81&50.13&30.05&29.1&69.27\\
  & Comp. Ratio(\%)$\uparrow$ & 11.65&20.21&29.9	&33.41&	67.92&	20.60&	44.36&	45.41&34.18\\ \hline

  \multirow{3}{*}{\fix{MonoGS}} & Acc.$\downarrow$ & 51.60&60.08&37.56&28.91&71.95&74.61&65.42&48.14&54.78\\
  & Comp.$\downarrow$ & 241.12&360.77&86.99&64.02&92.75&124.39&136.52&67.48&146.76\\
  & Comp. Ratio(\%)$\uparrow$ & 10.86&12.85&15.82&29.74&14.73&9.03&10.81&	27.08&16.37\\ \hline

  \multirow{3}{*}{\ffix{Photo-SLAM}} & Acc.$\downarrow$ & 12.78&13.10&17.72&7.67&7.07	&15.77&16.81&9.76&12.59\\
  & Comp.$\downarrow$ & 15.64&17.80&23.51&14.08&8.75&25.15&15.37&25.24&18.19\\
  & Comp. Ratio(\%)$\uparrow$ & 58.89&49.20&39.41&70.81&75.86&43.57&59.02&53.79&56.32\\ \hline

  \multirow{3}{*}{\fix{RTG-SLAM}} & Acc.$\downarrow$ &10.73&10.38&9.32&7.62&6.64&10.84&7.77&10.12&9.18 \\
  & Comp.$\downarrow$ & 15.10&10.84&11.57&9.31&6.07&13.19&7.94&12.18&10.78 \\
  & Comp. Ratio(\%)$\uparrow$ & 66.29&65.89&64.06&78.95&88.35&64.88&78.79&70.46&72.21\\ \hline

  \multirow{3}{*}{GS-ICP SLAM} & Acc.$\downarrow$ & 18.77&12.03&15.14&9.47	& 6.46&12.93&8.86&9.65&11.66\\
  & Comp.$\downarrow$ & 34.46&12.95&18.70&20.23&8.75&30.78&18.49&21.26&20.70 \\
  & Comp. Ratio(\%)$\uparrow$ & 37.19&63.58&46.75&65.76&82.87&52.77&70.21&	61.87&60.13\\ \hline

  \multirow{3}{*}{RemixFusion} & Acc.$\downarrow$ & \textbf{4.88}&\un{6.90}&\textbf{5.93}&\textbf{3.80}&4.77&\un{5.06}&\un{3.59}&\textbf{3.88}&\textbf{4.85}\\
  & Comp.$\downarrow$ & \textbf{5.39}&\textbf{4.93}&\textbf{4.94}&\textbf{4.12}&\un{3.58}&\textbf{5.71}&\textbf{3.68}&\textbf{3.11}&\textbf{4.43}\\
  & Comp. Ratio(\%)$\uparrow$ & \textbf{92.13}&\textbf{95.10}&\textbf{92.94}&\textbf{94.73}&\textbf{98.26}&\textbf{91.05}&\textbf{95.48}&\textbf{98.72}&\textbf{94.80}\\ \hline

  \multirow{3}{*}{RemixFusion-lite} & Acc.$\downarrow$ & {6.51}&{8.10}&\un{6.72}&{4.49}&	\un{4.72}&{5.19}&3.76&\un{4.05}&\un{5.44}\\
  & Comp.$\downarrow$ & \un{7.21}&6.46&\un{5.31}&\un{4.44}&{3.85}&\un{6.18}&\un{4.09}&\un{3.62}&\un{5.15}\\
  & Comp. Ratio(\%)$\uparrow$ & {85.13}&86.43&\un{91.11}&\un{93.79}&{97.7}&\un{89.51}&\un{93.68}&	\un{97.97}&\un{91.92}\\ 
  \bottomrule 

\end{tabular}
}
\label{tab:BS3D-mesh}
\end{table*}

\begin{table*}[!t]
  \renewcommand{\arraystretch}{1.25}
  \centering
  \caption{\fix{Quantitative comparison on \texttt{Replica}, \texttt{ScanNet}, and \texttt{TUM RGB-D}. `$\_$' denotes the second-best method. The results of all methods are from their original publications, except for RTG-SLAM and GS-ICP SLAM on \texttt{ScanNet}, which were evaluated using their official code due to the absence of results. The FPS (frames per second) is evaluated on average for all the sequences of each dataset. While LoopSplat achieves the best performance on average, its efficiency is undesirable. 3DGS-based methods like GS-ICP SLAM are accurate on the synthetic dataset (\texttt{Replica}),
  \fix{but fall short on real-world datasets.}
  Our method attains comparable tracking accuracy while maintaining real-time running efficiency across all datasets.}} 
  \scalebox{0.88}{
  \setlength{\tabcolsep}{1.05mm}
  {
  \begin{tabular}{@{}l  cccccccccc |cccccccc | ccccc| c}
  \toprule
  \multirow{2}{*}{Methods} & \multicolumn{10}{c|}{\textbf{Replica}} &\multicolumn{8}{c|}{\textbf{ScanNet}}   & \multicolumn{5}{c|}{\textbf{TUM RGB-D}} & \multirow{2}{*}{Avg.}  \\ \cmidrule{2-11}  \cmidrule{12-19} \cmidrule{20-24}    
  & {r0} & {r1} & {r2} & {o0} & {o1} & {o2} & {o3} & {o4} & {Avg.} & {FPS} & {0000} & {0059} & {0106} & {0169} & {0181} & {0207} & {Avg.} & {FPS} & {desk} & {xyz} &  {office} & {Avg.} & {FPS} \\ \hline  
  iMAP$^{*}$ & 70.1&4.5&2.2&2.3&1.7&0.5&58.4&2.6&17.8&0.2&56.0&32.1&17.5&70.5&32.1&11.9&36.7& 0.2&7.2&2.1&9.0&6.1&0.1 &22.4  \\
  NICE-SLAM  & 1.7&2.0&1.6&1.0&0.9&1.4&	4.0&3.1 &2.0 & 1.1& 8.6&12.3&{8.1}&10.3&12.9&\textbf{5.6}&9.6& 0.6&2.7 &1.8&3.0&2.5& 0.2 &4.8\\
  Co-SLAM &0.6&0.9&1.2&0.5&0.5&2.0&1.6&0.7&1.0& 6.6&{7.1}&11.1&9.4&\textbf{5.9}&11.8&7.1&{8.7}& 5.2 &2.7&1.9&{2.4}&2.3&4.8 &4.0\\
  ESLAM & 0.7&0.7&0.5&0.6&0.6&0.6&0.7&0.6&0.6&7.3 &7.3&{8.5}&\un{7.5}&\un{6.5}&\un{9.0}&\un{5.7}&\textbf{7.4}& 2.3 &2.5&{1.1}&2.4&{2.0}&0.2 &\un{3.3} \\ 
  MIPS-Fusion & 1.1&1.2&1.1&0.7&0.8&1.3&2.2&1.1&1.2&2.8 &7.9&10.7&9.7&9.7&14.2&7.8&10.0&3.1 &3.0&1.4&4.6&3.0&3.0 &4.6\\
  \ffix{DROID-SLAM} & 0.4&0.4&0.4&0.3	&0.3&0.5&0.6&0.5&0.4&\un{22.6}&8.4&7.8&9.7&10.8&10.7&6.5&9.0&{13.9}&2.2&1.4&1.8&1.8&\textbf{32.3}&3.7\\ \hline
  Point-SLAM &0.6&0.4&0.4&0.4&0.5&0.5&0.7&0.7&0.5&0.3 &10.2&{7.8}&8.7&22.2&	14.7&9.5&12.2&0.2 &2.6&1.3&3.2&2.4&0.1 &5.0\\
  MonoGS &0.3&0.2&0.3&0.4&0.2&0.3&\textbf{0.1}&0.8&0.3&0.7 & 9.8&32.1&8.9&10.7&21.8&7.9&	15.2 & 1.7& \textbf{1.6}& 1.4& 1.5& 1.5 & 1.4& 5.8\\
  SplaTAM & 0.3&0.4&0.3&0.5&0.3&0.3&0.3&0.6&0.4& 0.3&12.8&10.1&17.7&12.1&{11.1}&7.5&11.9& 0.5 &3.4&1.2&5.2&3.3 & 0.3&4.9\\
  \ffix{LoopSplat} & 0.3&0.2&0.2&0.2&0.2&0.5&0.2&0.3&0.3& 0.5&\textbf{6.2}&\un{7.1}&\textbf{7.4}&10.6&\textbf{8.5}&6.6&\un{7.7}&0.6 &2.1&1.6&3.2&2.3&0.6&\textbf{3.3}\\ 
  \ffix{Photo-SLAM} & 0.5&0.4&0.3&0.5&0.4	&1.3&0.8&0.6&0.6&\textbf{28.6} &8.3&\textbf{6.7}&9.2&8.5&78.8&7.5&19.8 & \un{22.3} &2.6&\textbf{0.3}&\textbf{1.0}&\un{1.3} &21.3 & 7.5\\
  RTG-SLAM & \un{0.2} & \un{0.2}&\un{0.1}&{0.2}&\textbf{0.1}&\un{0.2}&{0.2}&\textbf{0.2}&\un{0.2}&5.9 &125.1&109.1&128.8&7.8&28.1&6.9&67.6&2.9 &\un{1.7}&\un{0.4}&\un{1.1}&\textbf{1.1}&3.8 &24.1 \\
  GS-ICP SLAM & \textbf{0.2}& \textbf{0.2}& \textbf{0.1}& \textbf{0.2}& \un{0.1}& \textbf{0.2}& \un{0.2}& \un{0.2}&\textbf{0.2}&{22.4} &78.3&94.5&41.2&112.6&59.8&20.1&67.8& \textbf{24.3} &2.7&1.8&2.7&2.4&\un{24.2} &{24.4} \\ \hline
  RemixFusion &0.5&0.4&0.3&\un{0.2}&0.5&0.5&0.4&0.4&0.4& {14.3} & \un{6.9}&10.3&9.3&6.7&15.4&7.6&9.4&{12.0} & {2.3} &	1.8 &	{2.4}&2.2 &{10.6} & {3.9}\\ 
  \bottomrule 

\end{tabular}
}}
\label{tab:roomlevel-tracking}
\end{table*}

\begin{figure*}[!t]
  \centering %
    \begin{overpic}[width=1.0\linewidth]{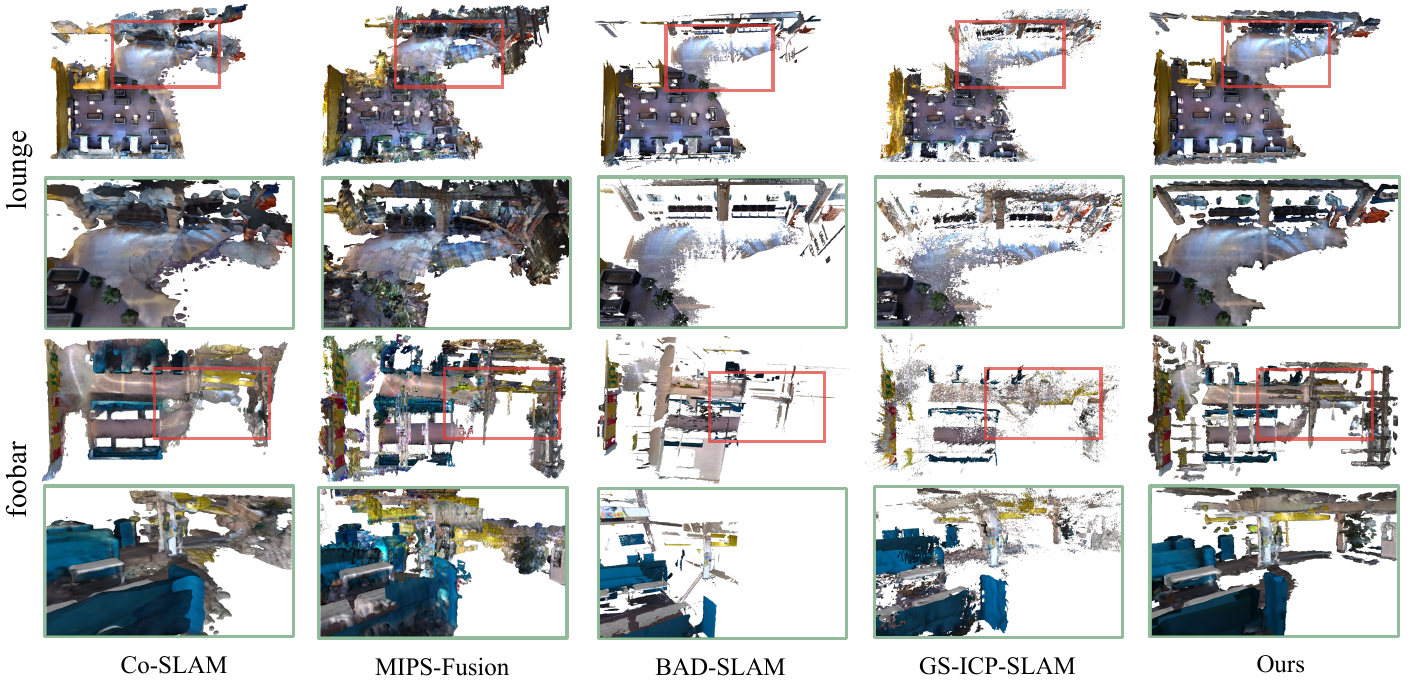}
    \end{overpic}
    \caption{Qualitative comparison of \texttt{lounge} and \texttt{foobar} on \texttt{BS3D} for different methods. The first row of each scene is the overview, and the second row is the zoom-in comparison corresponding to the regions marked with red rectangles. The reconstruction of our method is the most accurate and detailed,
    whereas other alternatives exhibit geometric distortions, severe holes, or over-smoothed results.
    }
    \Description{Figure 8. Fully described in the text.}
    \label{fig:mesh-comp1}
\end{figure*}

Figure~\ref{fig:mesh-comp1} displays the reconstruction with estimated camera poses. 
Significant distortion is noticeable in the reconstructions of other methods, whereas our approach delivers superior accuracy and consistency.
This underscores the precision of our pose estimation proposed in Section~\ref{sec:ba} and Section~\ref{sec:system}. Furthermore, the colorized trajectory in Figure~\ref{fig:mesh-selfcapture} 
demonstrates the robustness and accuracy of our pose estimation even in very large-scale scenes, contrasting with failures or substantial drifts observed in other methods. 
We highly recommend readers refer to our supplementary video and materials for more comprehensive visualization. 

Table~\ref{tab:uhumans2-benchmark} shows the tracking comparison of two large-scale sequences of \texttt{uHumans2}. The \texttt{apartment} contains three floors and the \texttt{office} is more than $1000m^2$, which presents significant challenges. 
Our method is the most accurate in terms of tracking, exceeding the SOTA method (ESLAM) by 12\%, superior to the others on both sequences. DROID-SLAM, utilizing layers of powerful dense bundle adjustment, derives the second-best tracking accuracy on \texttt{apartment} and the third-best accuracy on \texttt{office}, indicating great robustness. Corresponding rendering results are ignored, since DROID-SLAM omits reconstruction. Notably, MonoGS is the best 3DGS-based method on \texttt{uHumans2}, with tracking errors of 6.92cm (\texttt{apartment}) and 26.21cm (\texttt{office}). The increasing error in the larger scene (\texttt{office}) indicates the generalization limitation for 3DGS-based methods. In contrast, our method is robust on both sequences. Moreover, our method is a real-time system, while the best implicit method (ESLAM) and 3DGS-based method (MonoGS) are hard to meet the real-time requirements.

Table~\ref{tab:roomlevel-tracking} shows the quantitative results on room-level datasets, including \texttt{Replica}, \texttt{ScanNet}, and \texttt{TUM RGB-D}. Our method is the second-best on \texttt{scene0000$\_$00} of \texttt{ScanNet}, which is challenging.
Our method is comparable to the SOTA on average, and there is only a 0.6cm decrease in tracking accuracy compared to ESLAM and LoopSplat. Our method performs better than ESLAM on \texttt{Replica} with 37.1$\%$ improvement. While ESLAM achieves the second-best performance on average, its system FPS is about 3, and does not meet the real-time requirements, which is important for the online SLAM system. \ffix{Similar to ESLAM, LoopSplat showcases remarkable accuracy and robustness, which achieves the best tracking accuracy in this room-level benchmark thanks to the powerful loop closure modules. However, the modules are computationally expensive, and the efficiency is significantly limited (<1 FPS).} 
Room-level scenarios require fewer optimizations, where we obtain comparable performance and great running efficiency (3 times faster than ESLAM \ffix{and 21 times faster than LoopSplat}) at the same time with the residual-based mixed representation. Overall, our method is still superior to the others in terms of the online SLAM setting. \ffix{Note that DROID-SLAM achieves higher FPS by prioritizing camera tracking over dense reconstruction.} While the 3DGS-based methods like RTG-SLAM and GS-ICP SLAM are superior to other methods on \texttt{Replica}, they significantly fall short in real-world datasets like \texttt{ScanNet} and \texttt{TUM RGB-D}. 
This demonstrates the inherent instability of their camera tracking when exposed to real-world input noise.
SplaTAM and MonoGS both use the rendering loss as the objective function, and demonstrate robustness in real-world room-level scenarios. Detailed comparisons of room-scale sequences on \texttt{FastCaMo-Synth} (noise-free), are provided in the supplementary materials.

\begin{table}[!t]
  \renewcommand{\arraystretch}{1.25}
  \centering
  \caption{Comparison of training view rendering performance of \texttt{BS3D} using the estimated camera poses. Average results on 8 sequences are reported.
  '$\_$' denotes the second-best method. Our method obtains the best geometric rendering performance and better photometric rendering results than all the implicit methods. Note that GS-ICP SLAM is the best in RGB rendering but the worst in depth rendering, 
  \fix{indicating less attention to 3D geometry. Every 10 frames are evaluated using the estimated camera poses.}
  }
  \scalebox{0.95}{
  \setlength{\tabcolsep}{3.9mm}{
  \begin{tabular}{@{}l ccc c}
  \toprule
  \multirow{2}{*}{Methods} & \multicolumn{4}{c}{Metrics}     \\ \cline{2-5}
  
  & PSNR$\uparrow$ & SSIM$\uparrow$ & LPIPS$\downarrow$&  D-L1$\downarrow$  \\ \hline  
  NICE-SLAM & 20.88 & 0.940&  0.195& 0.173  \\
  Co-SLAM & 24.33 & 0.970&  0.163& 0.052 \\
  \fix{ESLAM} & {22.41}&{0.943}&{\un{0.150}}&{0.145} \\
  MIPS-Fusion & 23.30 & 0.962 & 0.193  & \un{0.044}\\ \hline
  \fix{SplaTAM} & {16.37}&{0.734}&{0.287}&{0.362} \\
  \fix{MonoGS} & {19.93}&{0.690}&{0.468}&{1.156} \\
  \ffix{Photo-SLAM} & 23.45& 0.962& 0.154& 1.085 \\
  \fix{RTG-SLAM} & {23.81}&{0.966}&{0.163}&{0.287} \\
  GS-ICP SLAM & \textbf{26.23} &  \textbf{0.980} & \textbf{0.118} & 0.223\\ \hline
  RemixFusion &\un{24.65} &  \un{0.971} & {0.154} & \textbf{0.031}\\

  \bottomrule 

\end{tabular}
}}
\label{tab:BS3D-rendering}
\end{table}

\subsubsection{3D Reconstruction.}

The traditional explicit methods, including ElasticFusion and BundleFusion, can not succeed in finishing the tracking and reconstruction for almost all scenes. Therefore, we only report the results of the implicit methods, BAD-SLAM, and the explicit 3DGS-based method. Note that the reconstruction mesh for 
3DGS-based methods 
is obtained using TSDF Fusion with rendered RGB-D images following 2DGS~\cite{Huang2DGS2024}. Implicit methods often produce noisy meshes in empty spaces, posing challenges for fair comparisons. Following~\cite{zhu2022nice,wang2023co}, we use the estimated camera poses to cull the mesh with ground-truth depths for evaluation. Ground-truth meshes obtained from LiDAR scans use the same strategies with ground-truth poses. The evaluation is performed three times, and the average results are reported. 
Due to the scaling differences between the results of BAD-SLAM and ground-truth meshes, we align the output point clouds with ground-truth meshes before evaluation. 

In Table~\ref{tab:BS3D-mesh},
\fix{our method outperforms the SOTA implicit method by 20.2$\%$ and the SOTA 3DGS-based method by 47.2$\%$ for the reconstruction accuracy on \texttt{BS3D}, and there is 46.6$\%$ improvement in completeness compared to the SOTA methods.}
Although BAD-SLAM is the most accurate on \texttt{lounge} and \texttt{study}, its completeness suffers due to explicit surfel representations. Our method exhibits marginally lower accuracy on these two sequences (0.21cm and 0.44cm decrease, respectively) but significantly outperforms other methods on the remaining sequences. Moreover, the completeness ratio of ours is over 90$\%$ for all sequences, surpassing the other approaches by over 9.5$\%$. This further validates the effectiveness of our residual-based mixed representations.
ESLAM demonstrates robust 3D reconstruction on \texttt{BS3D}, but is still worse than RemixFusion-lite. SplaTAM, MonoGS and Photo-SLAM exhibit poor reconstruction, primarily due to the discontinuity in depth rendering (Table~\ref{tab:BS3D-rendering}) and the fragile tracking in large-scale scenes (Table~\ref{tab:BS3D-tracking} and Table~\ref{tab:uhumans2-benchmark}).

Figure~\ref{fig:mesh-comp1} presents the reconstruction results, with an overview in the first row and a detailed zoom-in in the second row. Our methods can preserve the finest geometric details in these challenging large-scale scenes. In contrast, alternative methods struggle with unstable pose estimation and surface distortions, highlighting the effectiveness of our residual-based mixed representations. For example, in the second row of Figure~\ref{fig:mesh-comp1}, the comparisons of the white lines on the floor and black sofas illustrate that our reconstruction offers superior details in both texture and geometry.

\begin{figure*}[!t]
  \centering %
    \begin{overpic}[width=1.0\linewidth]{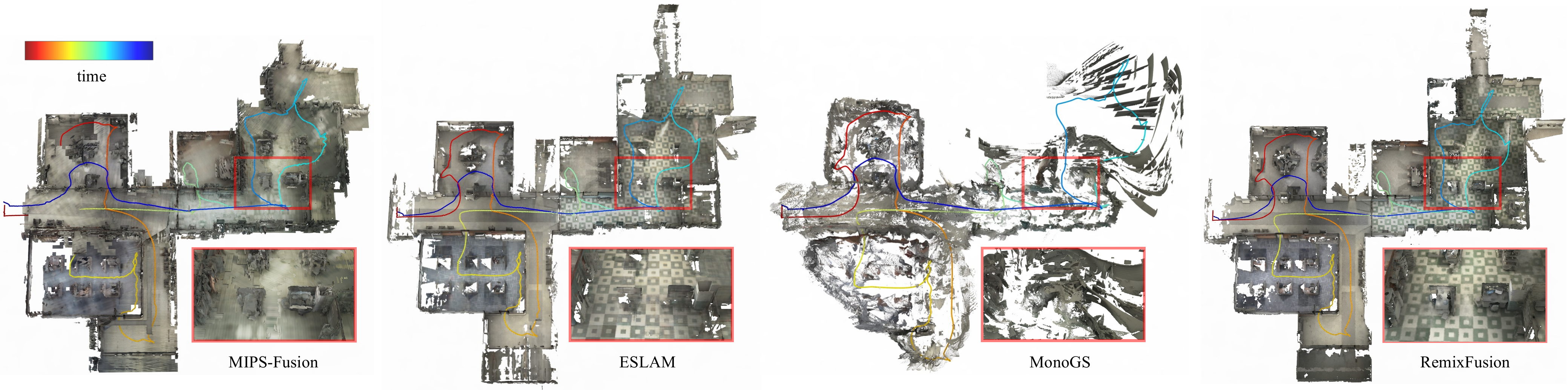}
    \end{overpic} 
    \caption{\fix{Qualitative comparison of \texttt{office} on \texttt{uHumans2}. The area of this challenging scene is $55m\times60m$, covering more than 1000 $m^2$. The reconstruction of MIPS-Fusion is noisy with many artifacts. ESLAM delivers a cleaner result but suffers from blurred reconstruction, particularly in areas with complex textures. While MonoGS successfully completes the tracking, but falls short in depth rendering, resulting in unsatisfactory reconstruction. In contrast, our reconstruction is the cleanest, accurately capturing both geometric and photometric details. Trajectories are colorized by time. Best viewed on screen. }}
    \Description{Figure 8. Fully described in the text.}
    \label{fig:uhumans-mesh}
\end{figure*}

\begin{figure}[!t]
  \centering
\includegraphics[width=8.5cm]{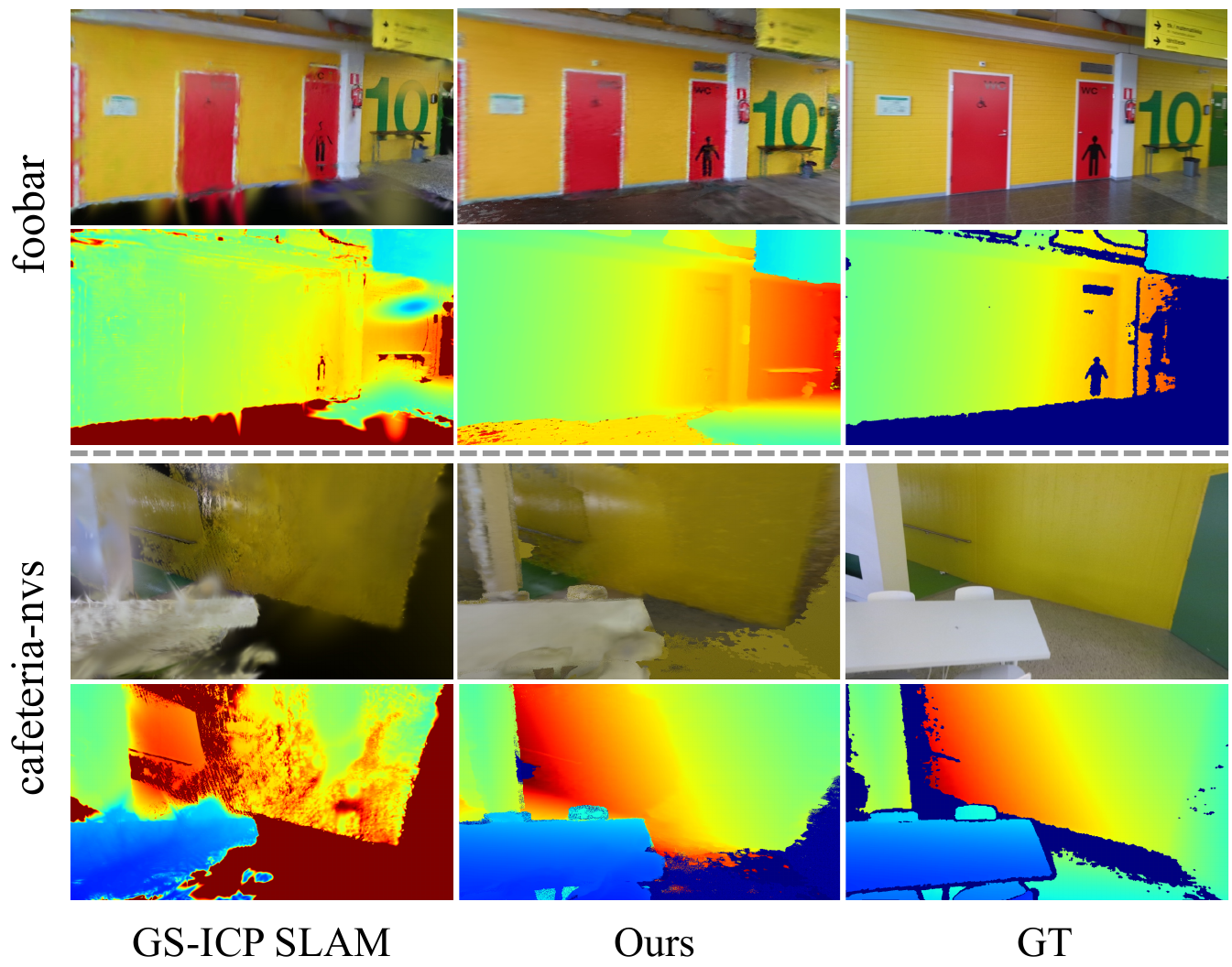}
\caption{Qualitative rendering comparison of training view (top) and novel view synthesis (bottom) using the estimated poses on \texttt{BS3D}. RGB (first row) and depth (second row) rendering are compared for each scene.} 
\Description{RemixFusion can preserve the detailed rendering of not only textures but also geometry.}
\label{fig:vis-2d rendering}       
\end{figure}

\begin{figure*}[p!]
  \centering %
    \begin{overpic}[width=0.94\linewidth]{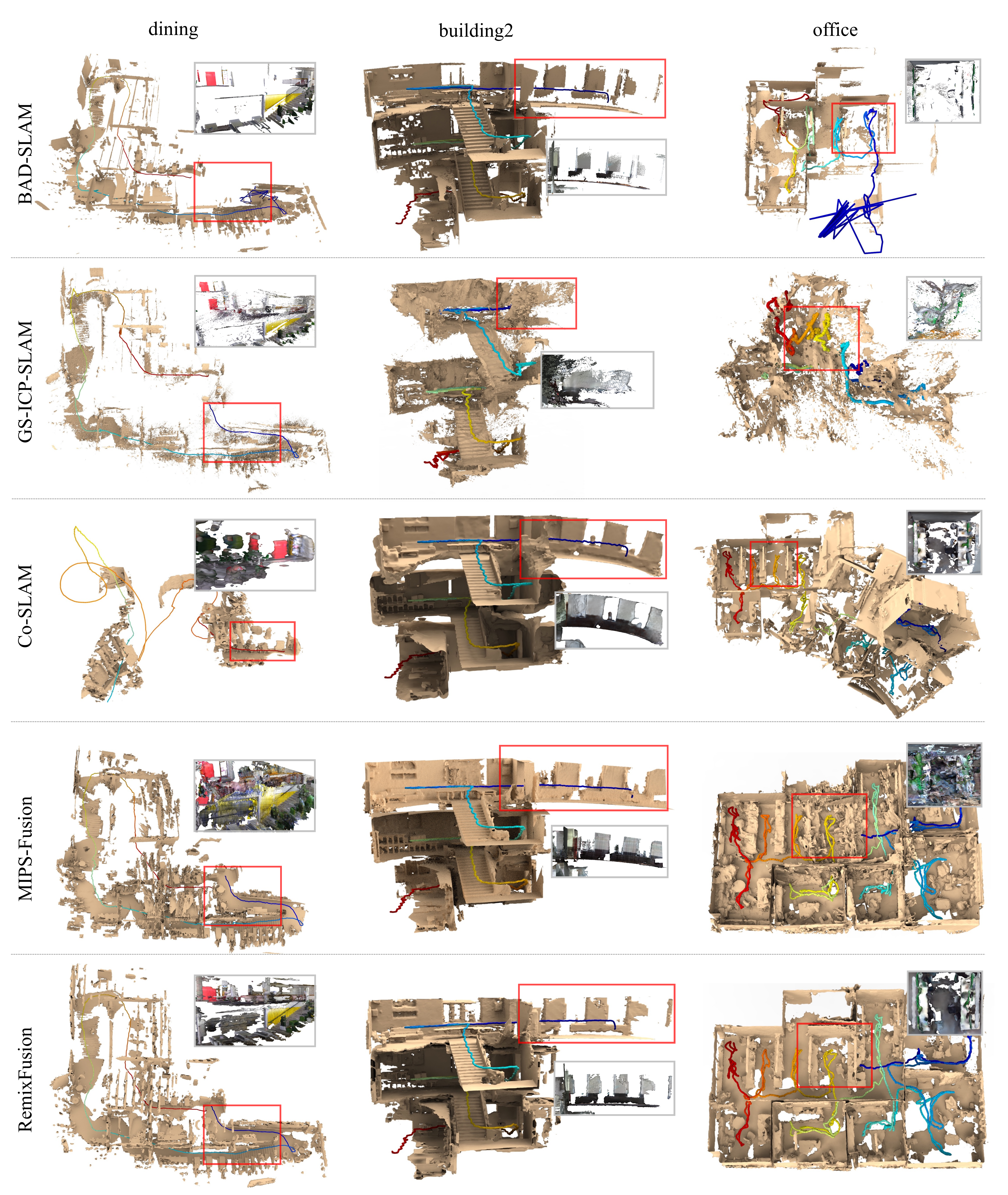}
    \end{overpic}
    \caption{Gallery of 3D reconstruction and camera tracking on \texttt{dining} of \texttt{BS3D}, \texttt{building2} of \texttt{FastCaMo-Large}, and \texttt{office} of \texttt{self-captured sequences}. These three sequences are composed of 5572, 7259, and 8656 images, which correspond to over 1000$m^2$, 200$m^2$, and 180$m^2$, respectively. The colorized trajectory indicates the estimated poses from the beginning (red) to the end (blue). Zoom-in comparisons are marked with red rectangles. Our method achieves the most accurate and robust performance in real-time, while there are failures or severe drifts for other approaches. }
    \Description{RemixFusion is the most accurate and robust approach in large-scale scenes, where there are different and challenging situations for SLAM.}
    \label{fig:mesh-selfcapture}
\end{figure*}

Figure~\ref{fig:mesh-selfcapture} presents additional results about the large-scale scenes, which are challenging for both tracking and reconstruction. The colorized trajectory means the scanning timeline: red denotes the start and blue denotes the end. 
The area of \texttt{dining} on \texttt{BS3D} exceeds $1000m^2$ and there are multiple staircases and floors. Our method successfully finished the pose estimation and detailed reconstruction, whereas other approaches failed either in tracking or returning to the starting position, resulting in reconstruction distortions.
For instance, MIPS-Fusion exhibits upward leaning and significant distortion in reconstruction upon returning to the second floor, despite employing loop closure for sub-maps. 
Note that the input depths for the floor of this scene are missing in the beginning, leading to some empty holes in the reconstruction. 
The second column in Figure~\ref{fig:mesh-selfcapture} shows the results of \texttt{building2} on \texttt{FastCaMo-Large}, where there are three floors. Other methods notably fail on the third floor, where the right corridor should appear flat and straight. The third column in Figure~\ref{fig:mesh-selfcapture} is a challenging self-captured sequence with faster camera motion, posing challenges for both tracking and mapping. 
RemixFusion demonstrates adaptability with robust pose estimation and efficient reconstruction, outperforming all the other methods.
There are severe distortions for MIPS-Fusion at the end of the trajectory (blue), and the reconstruction is of great noise due to the heavy and poorly aligned sub-maps. Our reconstruction is clean and more accurate, with FPS 4 times faster than MIPS-Fusion.
The other methods can succeed in reconstructing the first half of the sequence but fail in the middle, 
demonstrating the importance of robustness in large scenes.

\begin{table}[!t]
  \renewcommand{\arraystretch}{1.25}
  \centering
  \caption{Novel view and training view synthesis results of GS-ICP SLAM and RemixFusion on \texttt{cafeteria-nvs} of \texttt{BS3D}. Differences between novel views and training views (Diff.) are for intuitive understanding. Every 10 frames are evaluated. Both methods use ground-truth poses.}
  \scalebox{0.95}{
  \begin{tabular}{@{}l|c|ccc}
  \toprule
  {Methods}& Metrics & {Training Views} & Novel Views & Diff.  \\
  \hline 

  \multirow{4}{*}{GS-ICP SLAM} & PSNR$\uparrow$ & \textbf{29.68} &16.85 &-12.83 \\
  & SSIM$\uparrow$ & 0.954 & 0.817 & -0.103  \\
  & LPIPS$\downarrow$ & \textbf{0.074} & 0.237 & -0.163  \\ 
  & D-L1$\downarrow$ & 0.246 & 3.002 & -2.756 \\ \hline

  \multirow{4}{*}{RemixFusion} & PSNR$\uparrow$ & 28.53& \textbf{21.14} & -7.39 \\
  & SSIM$\uparrow$ & \textbf{0.991} & \textbf{0.963} & -0.028\\
  & LPIPS$\downarrow$ & 0.093 & \textbf{0.166} & -0.073\\ 
  & D-L1$\downarrow$ & \textbf{0.021} & \textbf{0.381} & -0.36 \\ 
  \bottomrule 
\end{tabular}
}
\label{tab:nvs}
\end{table}

The quality comparison of \texttt{office} on \texttt{uHumans2} is shown in Figure~\ref{fig:uhumans-mesh}. The color of the trajectory indicates the scanning time. The reconstruction of our method is the most compact and detailed, while there is noticeable blurring in some areas or distortion. (see the zoomed-in areas marked by red rectangles). Our method is capable of preserving the texture of the floor as well as the geometric structures of fine-grained objects on the table, which is hard for other methods. The most accurate, for tracking, 3DGS-based baseline (MonoGS) is hard to successfully reconstruct the entire sequence, and there is noticeable distortion and noise in geometry.

\subsubsection{2D Rendering.}
The 2D rendering comparison leverages the estimated poses from different methods to better focus on the scene reconstruction. Table \ref{tab:BS3D-rendering} shows the rendering comparison of training views on \texttt{BS3D}. Every 10 frames of each scene in the dataset are evaluated, and pixels with missing depth are filtered. Although GS-ICP SLAM, as the representative 3DGS-based explicit method, excels in photometric metrics such as PSNR, SSIM, and LPIPS, our method achieves superior photometric rendering compared to all implicit methods and the most accurate geometric rendering on average, as illustrated by D-L1. The D-L1 values of ESLAM are inferior to those of Co-SLAM and MIPS-Fusion, demonstrating the less detailed geometric reconstruction \fix{of the utilized tri-planes}. Our method significantly surpasses GS-ICP SLAM by 19.2cm in D-L1. Among the 3DGS-based methods, RTG-SLAM is the second-best method regarding PSNR and D-L1, which uses significantly fewer 3DGS, albeit at the cost of reduced reconstruction accuracy. \ffix{Photo-SLAM employs a hyper-primitive map that achieves comparable RGB rendering quality to implicit methods, yet exhibits inferior depth rendering performance, indicating potential overfitting of rendering and unsatisfactory geometric reconstruction.}

\begin{table}[!t]
  \centering
\caption{
Comparison of run-time FPS of the system and GPU memory usage 
(GPU mem.)
for \texttt{lounge} on \texttt{BS3D}. The metrics of pose estimation (ATE RMSE in cm) and reconstruction (F1-score) are also included. 
}
\renewcommand{\arraystretch}{1.25}
\scalebox{0.94}{
\setlength{\tabcolsep}{1.5mm}{
\begin{tabular}{l|cccc}
    \toprule
        Methods  & ATE RMSE $\downarrow$ & F1-score$\uparrow$  & FPS $\uparrow$ & GPU mem. $\downarrow$ \\ \hline
        NICE-SLAM & 11.0 & 54.39 & 0.5 & 12.6G    \\ 
        Co-SLAM & 48.8 & 18.02  & 5  & \textbf{5.6G}  \\ 
        MIPS-Fusion & 11.3 & 70.70  & 3  &  7.3G   \\ 
        \fix{ESLAM} & {6.3}  & {78.95}  & {2}  & {15.8G}    \\ \hline
        BAD-SLAM & 8.3 & 75.94 & 28 &  9.9G   \\
        \fix{SplaTAM} & {621.7} & {17.35} & {0.1} & {21.3G}  \\
        \fix{MonoGS} & {317.2} & {7.46} & {0.8} & {15.6G} \\
        \ffix{Photo-SLAM} & {24.6} & {44.42} & {21.5} & {9.1G} \\
        \ffix{LoopSplat} & {149.1} & {19.21} & {0.3} & {22.0G} \\
        \fix{RTG-SLAM} & {12.0} & {66.35} &  {2.1} & {10.1G} \\
        GS-ICP SLAM & 24.0 & 55.64 & \textbf{28} &  16.2G   \\ \hline
        RemixFusion & \textbf{4.2} & \textbf{90.25} & 12 &  9.8G \\ 
        {RemixFusion-lite} & 4.6 & 89.50 & 25 & 8.0G  \\ 
        \bottomrule
\end{tabular}
}}
\label{tab:runtime}
\end{table}

GS-ICP SLAM is good at 2D photometric rendering, but its geometric rendering (D-L1) is notably inferior compared to other alternatives. 
\fix{This can be attributed to the reason that 3DGS-based representations achieve high-fidelity rendering through the fitting of 2D training views, indicating less attention to the 3D geometry in the online setting. }
This can also be proved by the metrics of 3D reconstruction accuracy and completeness presented in Table \ref{tab:BS3D-mesh}.
\fix{In terms of real-time reconstruction, 3DGS-based methods are easy to achieve high-fidelity RGB rendering. However, they struggle to get accurate geometric reconstruction (see the D-L1 metric in Table~\ref{tab:BS3D-rendering}). 
For example, RTG-SLAM performs well on the scene \texttt{study} of \texttt{BS3D} with 6.7cm in ATE RMSE (shown in Table~\ref{tab:BS3D-tracking}). However, its 3D reconstruction quality (shown in Table~\ref{tab:BS3D-mesh}) is not consistent with the performance of tracking. For comparison, Co-SLAM achieves 6.5cm on the scene \texttt{study} of \texttt{BS3D} in ATE RMSE, which is similar to RTG-SLAM; the 3D reconstruction of Co-SLAM is much better than RTG-SLAM regarding the accuracy and completeness. Note that RTG-SLAM focuses on the 3DGS optimization using the tracking module from other methods. Its FPS and reconstruction are still inferior to Co-SLAM. 
Improving geometric reconstruction quality for 3DGS-based methods often requires more optimization iterations, which is generally unacceptable for real-time SLAM.}

Furthermore, the comparison of novel view synthesis in Table \ref{tab:nvs} demonstrates that our residual-based representation significantly outperforms 3DGS in photometric rendering (PSNR) by 25$\%$ and surpasses it in geometric rendering (D-L1) by over 7 times. This indicates that the 3DGS-based methods 
\fix{are less geometry-aware in the online setting.} The difference between training views and novel views indicates the robustness of our mixed representation.
Note that the sequence \texttt{cafeteria-nvs} of \texttt{BS3D} in Table~\ref{tab:nvs} uses the manually partitioned train (50$\%$) and test images (50$\%$), with ground-truth poses 
excluding the effects of tracking. Overall, our residual-based representations surpass both implicit methods and explicit methods, including 3DGS, taking efficiency and performance into consideration in large scenes.

Figure~\ref{fig:vis-2d rendering} illustrates the 2D rendering results of our method and GS-ICP SLAM (the second-best method on \texttt{BS3D}), which indicates that our method is more stable in photometric rendering and much better in geometric rendering than the 3DGS-based method. For example, our rendering of the small table near the green number 10 on \texttt{foobar} is more detailed and complete. We attribute this to the residual-based representation of RemixFusion and the direct supervision of 3D space. The novel view synthesis comparison is presented at the bottom of Figure~\ref{fig:vis-2d rendering}. Our method remains accurate, while there is notable noise for GS-ICP SLAM, which proves that GS-ICP SLAM only overfits the 2D training views without accurately reconstructing the real 3D geometry.

\fix{Table~\ref{tab:uhumans2-benchmark} shows the rendering comparison on uHumans2 dataset. Our method outperforms the SOTA in terms of tracking by 12$\%$ and D-L1 by 12.3$\%$. Our method is the best in terms of PSNR on \texttt{office}, surpassing the SOTA by 0.34dB. While ESLAM is the second-best method for D-L1, its system FPS is over three times slower than our method. For 3DGS-based methods, MonoGS demonstrates the best RGB rendering on \texttt{apartment} (surpassing the second-best one by 4.49dB for PSNR), but its depth rendering is far worse than other methods. This also indicates that 3DGS-based methods are less geometry-aware in the online setting, as mentioned above. Overall, our method is the best real-time method in terms of tracking and rendering on average.}

\begin{figure}[!t]
  \centering %
    \begin{overpic}[width=1.0\linewidth]{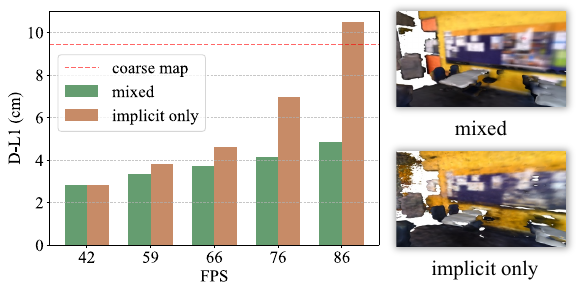}
    \end{overpic}
    \caption{Ablation studies of the reconstruction with (mixed) or without residuals (implicit only) based on the coarse explicit map. The D-L1 (cm) values (left) of the mixed representations are stable and accurate even if the required FPS of mapping increases. The mesh comparison (right) corresponding to FPS=$86$ proves the effectiveness of the mixed representations.}
    \Description{RemixFusion can reconstruct the whole scene within extremely few optimization iterations.}
    \label{fig:ablation-mapping}
\end{figure}

\subsubsection{Runtime and Memory Analysis}
Table ~\ref{tab:runtime} illustrates the system FPS and the performance of both tracking and mapping. 
To ensure a comprehensive comparison, we focus on the scene \texttt{lounge} from \texttt{BS3D}, which is successfully handled by most methods.
Note that the FPS of explicit methods like ElasticFusion and BundleFusion is high, but both of them fail in this scene. Therefore, they are not included in this table. Our method achieves the best trade-off between GPU memory usage and accuracy.  
Our method exhibits relatively consistent GPU memory usage across different scene scales. In contrast, the memory consumption of explicit methods grows substantially as they process more frames in larger scenes.
Our mixed representation demonstrates superior system FPS compared to implicit methods, and the lightweight version of our method can run at 25 FPS with $8$GB GPU memory usage, which is as fast as the explicit alternatives and requires less computational overhead.

Conversely, approaches like Co-SLAM and MIPS-Fusion employ hash schemes or multiple sub-maps to reduce memory consumption but struggle with efficiency and robustness in memorizing large-scale scenes. 
Methods like ESLAM, SplaTAM, and MonoGS are not memory-friendly, requiring half-resolution images as input. Additionally, their system FPS is significantly lower than that of other methods.
Similarly, LoopSplat is not memory-efficient in large scenes and exhibits running inefficiency (<1 FPS), primarily due to the computationally expensive modules of loop closure and post-refinement. Photo-SLAM, leveraging ORB-SLAM3 for tracking and a map with hyper primitives for mapping, is lightweight and efficient. However, Photo-SLAM indicates worse tracking and reconstruction accuracy on large scenes.
The traditional explicit method, like BAD-SLAM, is real-time but lacks robustness. GS-ICP SLAM, as the representative of 3DGS-based methods, is fast but compromises memory efficiency.
Although RTG-SLAM leverages efficient techniques to reduce the redundant 3DGS and requires less GPU memory in the process, it requires large GPU memory for global optimization at the end of tracking.
Our method stands out as the most efficient system, coupled with the most accurate pose estimation as well as reconstruction, considering both FPS and GPU memory usage.

\subsection{Ablation Studies} 

We perform ablation studies mainly on the BS3D~\cite{mustaniemi2023bs3d} dataset and report the average results of all scenes to evaluate the effectiveness of the modules proposed in RemixFusion. We analyze the necessity of the key components proposed in our methods for validation, which are residual-based mapping, residual-based bundle adjustment, and gradient amplification, respectively.

\subsubsection{Residual-based Mapping}

One key innovation is the residual-based mapping, 
crucial for enabling real-time dense reconstruction of large-scale scenes in our system.
To validate this design, we first compare our mixed method against using solely the implicit module without residuals. Parameters are the same except for the residual-based representations. Moreover, we demonstrate how the reconstruction quality varies according to the desired mapping FPS.

\begin{figure}[!t]
  \centering
\includegraphics[width=8.5cm]{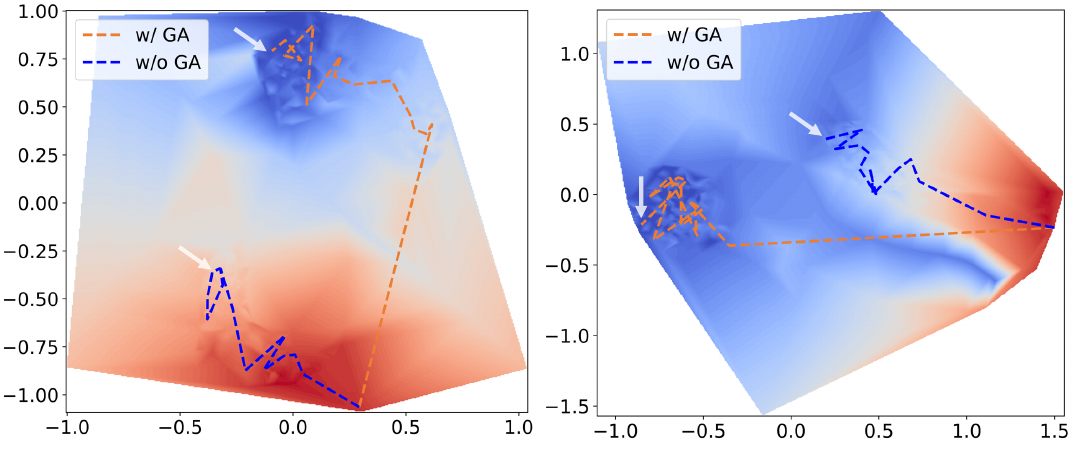}
\caption{Visualization of the optimization of bundle adjustment with (orange) and without (blue) gradient amplification (GA) on \texttt{waiting} (left) and \texttt{corridor} (right) of \texttt{BS3D}. 6D poses of all frames are depicted in 2D using gradient colorized from red to blue, indicating high and low cost function values, respectively. BA with GA converges to a better solution.
} 
\Description{With the proposed gradient amplification, the bundle adjustment is encouraged to explore more and find more global optimal solutions.}
\label{fig:exp-pmvis}       
\end{figure}

\begin{table}[!t]
  \centering
  \caption{
  Ablation studies of 3 designs for the residual-based bundle adjustment. The average ATE RMSE (in cm) of 8 sequences on the \texttt{BS3D} dataset is reported. RBA and TBA denote the residual-based BA and traditional BA, respectively, and GA denotes the gradient amplification.
  }
  \renewcommand{\arraystretch}{1.25}
  \scalebox{1.0}{
  \setlength{\tabcolsep}{4.8mm}{
  \begin{tabular}{  c c c | c}
      \toprule
          RBA & TBA  & GA  &  ATE RMSE \\ \hline
           &  &    & 6.39 \\ 
          \checkmark &  &    & 5.68 \\ 
            & \checkmark &   & 5.74 \\ 
            & \checkmark & \checkmark  & 5.14 \\ 
           \checkmark &  & \checkmark   &  4.61 \\ 
      \bottomrule
  \end{tabular}
  }}
\label{tab:ablation-ba}
\end{table}

We evaluate 8 sequences of \texttt{BS3D} and report the average D-L1 values in Figure~\ref{fig:ablation-mapping}. For a fair comparison focused solely on reconstruction, excluding tracking, we use the ground-truth poses for reconstruction, and no bundle adjustment for poses is utilized. As illustrated by Figure~\ref{fig:ablation-mapping}, the D-L1 values (cm) of our method with the residual design exhibit slower degradation even with higher required FPS. This indicates that our mapping module with residuals achieves faster convergence, and the D-L1 values are below 5cm even if the mapping process is running at 86 FPS. \fix{Moreover, the result of the mixed representations is significantly better than that of the coarse map, indicating the effective residual-based learning based on the coarse map.} Figure~\ref{fig:ablation-mapping} also shows the reconstruction comparison when the mapping FPS is 86, which means only about $50$ iterations for mapping are performed for about 2000 frames. This is quite challenging, yet the results demonstrate that the residual-based mapping module can achieve meaningful reconstruction details with minimal optimization. The experiments prove that the proposed residual-based mapping is efficient in preserving the detailed reconstruction within a limited time.

\begin{figure}[!t]
  \centering
  \includegraphics[width=8.2cm]{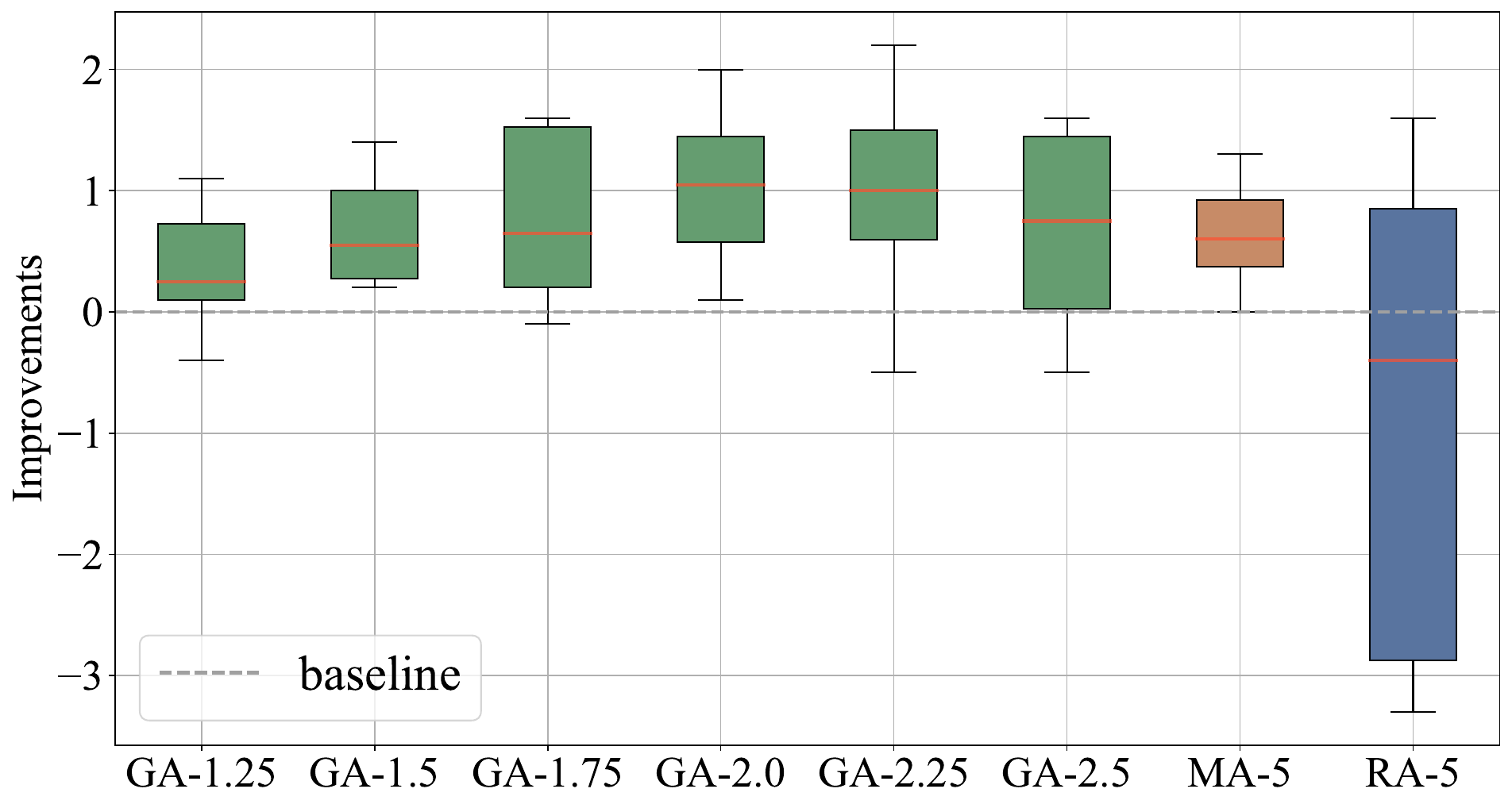}
    \caption{Comparison of different factors $k$ used in the proposed gradient amplification (GA) and the different amplification methods (MA for manually designed amplification and RA for randomized amplification) for residual-based BA. The baseline (gray dashed line) indicates the ATE RMSE (cm) without gradient amplification. Improvements for 8 scenes on \texttt{BS3D} compared to the baseline are shown.}
  \Description{The threshold of 2 for \fix{gradient amplification} improves all the sequences on BS3D, and the average improved performance of it is the highest.}
  \label{fig:ablation-pm}       
\end{figure}

\subsubsection{Residual-based bundle adjustment}
Another key insight of our method is the residual-based bundle adjustment (RBA for short). Here we evaluate the effectiveness of our residual-based bundle adjustment for pose refinement. We compare our method to the traditional bundle adjustment (\fix{TBA} for short) that optimizes the individual pose variables. Additionally, we validate the effectiveness of our proposed gradient amplification (GA for short).

\begin{figure}[!t]
  \centering
\includegraphics[width=8.5cm]{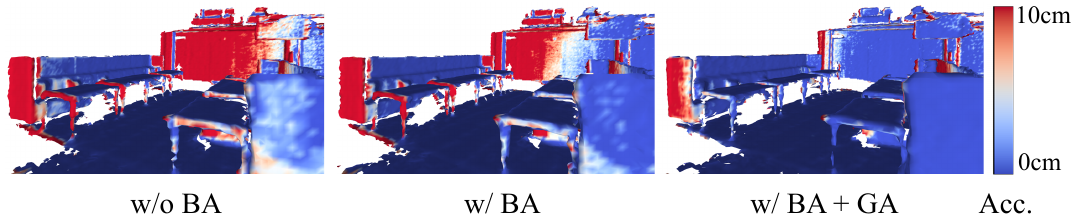}
\caption{Visualization of the residual-based bundle adjustment on \texttt{foobar} of \texttt{BS3D}. The mesh is colorized with the accuracy (Acc.) of the reconstructed mesh compared to the ground-truth mesh, which indicates the distance between the reconstructed mesh and the ground-truth mesh.} 
\Description{The proposed residual-based bundle adjustment coupled with \fix{gradient amplification} improves the accuracy and robustness of camera tracking in large-scale scenes.}
\label{fig:exp-bavis}       
\end{figure}

Compared to the commonly used bundle adjustment, which individually optimizes the pose of each frame, our proposed method employs a single MLP for 6D residual pose prediction, offering enhanced global consistency. Table~\ref{tab:ablation-ba} illustrates that the average ATE RMSE without bundle adjustment (baseline) is 6.39cm. Applying the RBA instead of TBA for pose refinement reduces the error marginally from 5.74cm to 5.68cm. Introducing GA to help bundle adjustment results in improved pose estimation for both TBA and RBA. Moreover, using RBA with GA reduces the error from 5.68cm to 4.61cm, resulting in an improvement of over 1cm. This proves the effectiveness of the proposed gradient amplification, which is simple yet effective for the residual-based BA in large-scale scenes. Figure~\ref{fig:exp-bavis} demonstrates the comparison of different methods on \texttt{foobar} of \texttt{BS3D}, which validates the effectiveness and necessity of the proposed residual-based bundle adjustment and gradient amplification. In summary, the proposed residual-based BA can focus on detailed refinement of the initial camera poses and significantly improve the accuracy of pose estimation in large-scale scenes.

\subsubsection{Adaptive Gradient Amplification}
A key design in the residual-based BA is the proposed adaptive gradient amplification used in BA. As illustrated in Eq. \ref{eq:tsdf-align-k-new}, this technique amplifies the optimization gradients derived from the reconstructed surface, allowing the BA to circumvent local minima and thus achieve more globally consistent solutions in real-world large-scale scenarios.
This design aims to mitigate the risk of falling into local optima, a common challenge in large-scale scenes.
Visualization of the optimization of BA on \texttt{waiting} and \texttt{corridor} of \texttt{BS3D} is shown in Figure~\ref{fig:exp-pmvis}, where the 6D poses of all keyframes are visualized in 2D and colorized by the cost functions. 
Without the proposed gradient amplification, the optimization tends to get trapped in the local minimum (illustrated by the blue dashed line). However, the proposed gradient amplification enables BA to escape from the local minimum and obtain a better solution (the orange dashed line).

Furthermore, we compare our method (GA) with the manually designed 
amplification (MA)
and the randomized 
amplification (RA).
Figure~\ref{fig:ablation-pm} presents the improvement of BA under various truncation thresholds compared to the baseline (without GA). The positive improvement indicates the effectiveness of gradient amplification. Results of eight scenes of \texttt{BS3D} are reported as the range of the bar chart, and the red line indicates the median. GA achieves the best results on average with $k=2$, showing positive improvements across all scenes. MA-5 and RA-5 indicate that the camera poses are guided towards the directions with 5cm, where the cameras are facing or randomized directions. Our proposed GA yields the most accurate pose estimation, while MA also improves the performance, which is less than GA. RA fails to improve the performance of bundle adjustment for all sequences. In summary, the improved results prove the effectiveness of the proposed gradient amplification.

\begin{figure}[h]
  \centering
  \includegraphics[width=8.0cm]{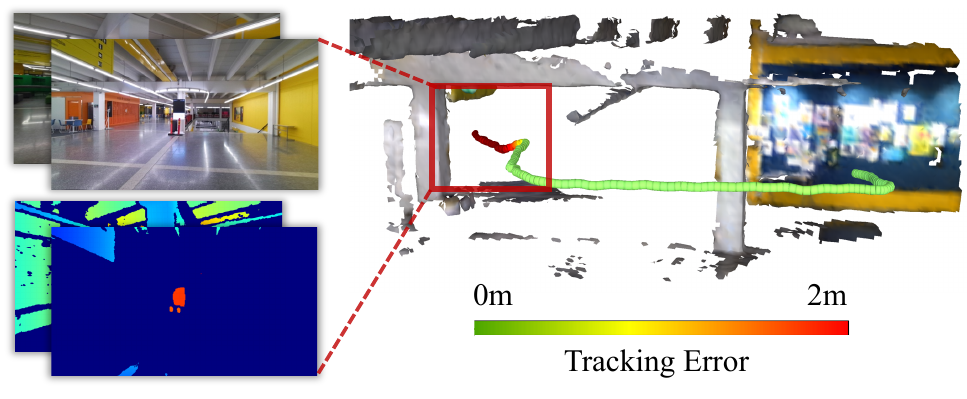}
  \caption{Typical failure case of RemixFusion. The observed depth information is severely missing (left), resulting in obvious drift in camera tracking and distortion in modeling (right). The trajectory is colorized by the errors.}
  \Description{Figure 15. Fully described in the text.}
  \label{fig:failcase}       
\end{figure}

\subsection{Limitation and Failure Cases.}

There is a primary limitation of our method.
Robust pose estimation becomes challenging when the depth images contain extensive missing regions, as our approach critically relies on depth information. 
Consequently, our method can not be applied in scenarios where only RGB images are available. Another limitation is that our method is not robust in dynamic environments.
Failure cases of our method are shown in Figure~\ref{fig:failcase}. The majority of observed depth information in the middle of the sequence (highlighted by red rectangles) is missing, making it hard and unstable for both pose estimation and reconstruction.

%% file: 07-conclusion.tex
\section{CONCLUSION}
It is still a challenging problem to perform online dense reconstruction for a large-scale environment with fine-grained geometry details preserved. It is critical to formulate a memory-friendly scene representation that can support efficient and high-quality tracking and mapping. With our work, we wish to bring to the community's attention that a residual-based mixture is a proper way to take advantage of both explicit and implicit formulations. 
By reducing the learning burden on implicit networks through coarse-grained explicit storage, we have significantly accelerated the efficiency of online reconstruction. This enhancement allows our residual-based framework to preserve more reconstruction details while ensuring real-time performance. Note that the residual idea also inspires a new approach to pose estimation, where we optimize only the pose changes during multi-frame joint optimization, thus reducing the network's learning complexity.
The evaluation comparison between our method and other alternatives demonstrates the superiority of RemixFusion in tracking accuracy and mapping quality for large-scale scene reconstruction. While our method can handle large-scale scene reconstruction with limited memory cost, it is still worth doing as future work to make this mixed representation dynamically scalable, which may be able to support larger online dense reconstruction at a city block level.

%% file: appendix.tex
\clearpage
\setcounter{page}{1}
\section{SUPPLEMENTARY MATERIALS}

We first provide implementation details of our designs mentioned in the methods. We also provide the evaluation strategies and details that are used in the manuscript and supplementary materials. More qualitative and quantitative results of different benchmarks including the large-scale and room-level datasets are reported for pose estimation and scene reconstruction. We deeply analyze these experimental results to validate the core insights of our method.

\subsection{Implementation Details}
\label{sec:implementation}

To be specific, we introduce the preliminaries about randomized optimization and then present details about camera tracking. The details about residual-based mapping and residual-based joint bundle adjustment are then demonstrated. The configuration of the lightweight version of RemixFusion (RemixFusion-lite) is also included here.

\paragraph{Preliminaries of Randomized Optimization}
We first outline the randomized optimization and \textit{particle swam template(PST)} leveraged in \cite{zhang2021rosefusion} for better understanding. The randomized optimization is regarded as an appropriate alternative to solve the optimization problems, especially for highly non-linear optimization. Here, we mainly illustrate the usage in pose estimation. In brief, the goal of pose estimation is to obtain the 6DoF camera poses 
$[\mathbf{R}\mid\mathbf{t}]$ $([\mathbf{R}\mid\mathbf{t}]\in SE(3))$ 
of each of $N$ input RGB-D frames. For each pixel $(i,j)$ in a single  frame, we can back-project the pixels to 3D space with the corresponding depth information and estimated camera poses $[\mathbf{R}\mid\mathbf{t}]$:

\begin{equation}
\label{eq:formulation 2}
\mathrm{X}_{ij}=\mathrm{R}\mathrm{x}_{ij}+t,
\end{equation}
where $\mathrm{x}_{ij}$ indicates the position in the camera coordinate and $\mathrm{X}_{ij}$ is the position in the world coordinate.

Following~\cite{bylow2013real}, we can use the back-projected points with the estimated camera poses $[\mathbf{R}\mid\mathbf{t}]$ to query the TSDF volume in a frame-to-model way. The smaller averaged TSDF values for sampled pixels in a frame are, the closer corresponding points are to the zero-crossing surface, which means the estimated camera pose is more accurate. Therefore, the objective cost function can be formulated as Eq. \ref{eq:3-sup}.

\begin{equation}
\label{eq:3-sup}
(\mathbf{R}^*,\mathbf{t}^*)=\arg\min_{R,t}\sum_{(i,j)\in I_d}\phi(\mathbf{R}\mathbf{x}_{ij}+t)^2,
\end{equation}
where $I_d$ denotes the depth image and $\phi:\mathbb{R}^3\to\mathbb{R}$ means the query of the reconstructed TSDF volume. \fix{$\phi$ is used to calculate the difference between the queried TSDF values from the moving volume and the approximated ground-truth TSDF supervision.}

Gauss-Newton methods~\cite{bylow2013real} and randomized optimization based on \textit{particle filter pose optimization (PFO)} can both be used to optimize the objective cost function (Eq. \ref{eq:3-sup}). The difference is that the latter is more robust and better at handling the highly nonlinear optimization. The randomized optimization provides robust initial pose estimation even in scenarios of large camera rotation and translation, which is necessary for the real-world application in large-scale scenes because it is hard to ensure a low and stable motion when scanning. For randomized optimization, \textit{pre-sampled particle swam template (PST)} in~\cite{zhang2021rosefusion} leverages multiple pre-sampled normalized 6D pose templates, which are rescaled and moved continuously to improve the sampling efficiency. This is employed in \textit{PFO} for faster convergence. The goal is to minimize the cost function in Eq. \ref{eq:3-sup} and the likelihood function can be formulated as Eq. \ref{eq:likelihood}.

\begin{equation}
\label{eq:likelihood}
\rho\left(g\mid\mathbf{R},\mathbf{t},I_d\right)=\exp\left(-\frac1\tau\sum_{i\in|I_d|}\phi\left(\mathbf{R}\mathbf{x}_i+\mathbf{t}\right)^2\right),
\end{equation}
where $g$ denotes the observation function and $\tau$ is a temperature.

\begin{algorithm}[!t]
  \begin{algorithmic}
      \STATE \textbf{Input:} Camera poses $\mathcal{G}$, implicit module $\Theta$, explicit map $\mathcal{V}_{\text{coarse}}$, truncation distance $tr, tr_e$, threshold $\tau_c$ and iterations $N_\text{map},N_\text{ba}$.
      \STATE \textbf{Output:} Optimized camera pose $\mathcal{G}^{\prime}$.
      \STATE \textbf{Initialize:} approximate ground-truth TSDF $T_{gt}$, 

      \FOR{$i = 1$ to $N_\text{map}$}
      \STATE $\beta \leftarrow$ \texttt{Tri-interpolation}($\mathcal{V}_{\text{coarse}}$).  
      \STATE $\text{SDF} \leftarrow T^e$ 
      \STATE $\hat{\beta}\leftarrow\Upsilon(SDF,tr,1)$.
      \STATE \texttt{Minimize} $\hat{\beta}\leftrightarrow T_{gt}$~~\qquad\qquad\quad\quad\quad\quad//mapping process
      \ENDFOR

      \FOR{$i = 1$ to $N_\text{ba}$}
      \STATE $\beta \leftarrow$ \texttt{Tri-interpolation}($\mathcal{V}_{\text{coarse}}$).  
      \STATE $\text{SDF} \leftarrow T^e$ 
      \STATE $ \hat{\beta}\leftarrow\Upsilon(SDF,tr,k\cdot\tau_c)$.

      \STATE \texttt{Minimize} $\hat{\beta}\leftrightarrow T_{gt}$\quad\qquad\qquad\qquad\qquad\quad//BA process
      \ENDFOR
      \STATE Return the updated model $\Theta$ and camera poses $\mathcal{G}^{\prime}$.
  \end{algorithmic}
  \caption{{gradient amplification} for bundle adjustment
  }
  \label{algorithm:ga}
\end{algorithm}

\paragraph{Camera Tracking.} In general, we utilize a scalable moving volume based on randomized optimization, which is robust and lightweight in large-scale scenes. This is inspired by the success of the moving volume techniques used in~\cite{Whelan2012,Roth2012}. By virtue of this front-end tracking module, we can obtain the initial pose for each input RGB-D frame. The initial poses are then refined by the residual-based bundle adjustment. As for the front-end camera tracking, we use $N_{\text{ro}}=20$ iterations for randomized optimization on all datasets for each input RGB-D frame. As for input RGB-D sequence $\{I_c,I_d\}$, we first estimate the camera pose using randomized optimization based on the local moving volume $V_l$ with $N_{\text{ro}}$ iterations. The number of PST is iteratively selected from $\{\varTheta_1,\varTheta_2,\varTheta_3\}=\{10240,3072,1024\}$, which refer to the number of pre-sampled particles. The sampling rates are $\{\varUpsilon_1,\varUpsilon_2,\varUpsilon_3\}=\{32,16,8\}$, which indicates that we sample every $\varUpsilon$ pixel of the current image. In this way, the comprehensive and efficient sampling strategies (e.g., 10240 particles with 32 pixels sampling) achieve a good trade-off of accuracy and efficiency. As for the details about moving and rescaling PST, please refer to ~\cite{zhang2021rosefusion}.

After we obtain the estimated camera pose $\varphi_t$ of the current frame, the camera pose $\varphi_{t+1}$ of the next frame is initialized as $\varphi_t$, which is optimized using the randomized optimization mentioned above. The position of the first moving volume $\mathcal{V}_1$ is initialized as the anchor camera $\varphi_a$, corresponding to the first input frame. When the Euclidean distance between the current camera $\varphi_t$ and the anchor camera $\varphi_a$ exceeds a threshold $\tau_v$ in any direction of 3 axes (e.g., $x,y,z$ of the world coordinate), we would create a new volume and inherit the overlapping regions between $\mathcal{V}_t$ and $\mathcal{V}_{a}$, which refers to the copied volume with the position $\varphi_t$ and the anchor volume.  $\tau_v$ is set to 1$m$ for all datasets. The explicit volume $\mathcal{V}_a$ is $14m\times14m\times6m$ on BS3D~\cite{mustaniemi2023bs3d} and FastCaMo-Large dataset~\cite{tang2023mips}, $8m\times8m\times6m$ on the rest datasets. 
The implementation is based on PyCUDA~\cite{kloeckner_pycuda_2012} for acceleration. 

\fix{For the details of the proposed scalable randomized optimization, there are slight differences for different datasets. The difference mainly involves the initial scale $s_i$ of the PST, the scale coefficient $s_c$ used in the optimization, the voxel size $v_s$ of the moving volume, the truncation distance is the three times as $v_s$, and the range that is $r_d$ times as the truncation distance for sampling the points along the rays. For the BS3D dataset, we use $s_i=0.01$, $s_c=0.09$, $v_s=0.02$, $r_d=0.0$. For the Replica dataset, we use $s_i=0.02$, $s_c=0.09$, $v_s=0.01$, $r_d=0.0$. For the ScanNet dataset, we use $s_i=0.09$, $s_c=0.5$, $v_s=0.04$, $r_d=3.0$ for most scenes, and there are slight differences for different sequences. As for the TUM RGB-D dataset, we use $s_i=0.01$, $s_c=0.09$, $v_s=0.01$, $r_d=0.1$. Details on the uHumans2 dataset are similar to those of the BS3D dataset.}

\begin{figure}[!t]
  \centering
\includegraphics[width=8.5cm]{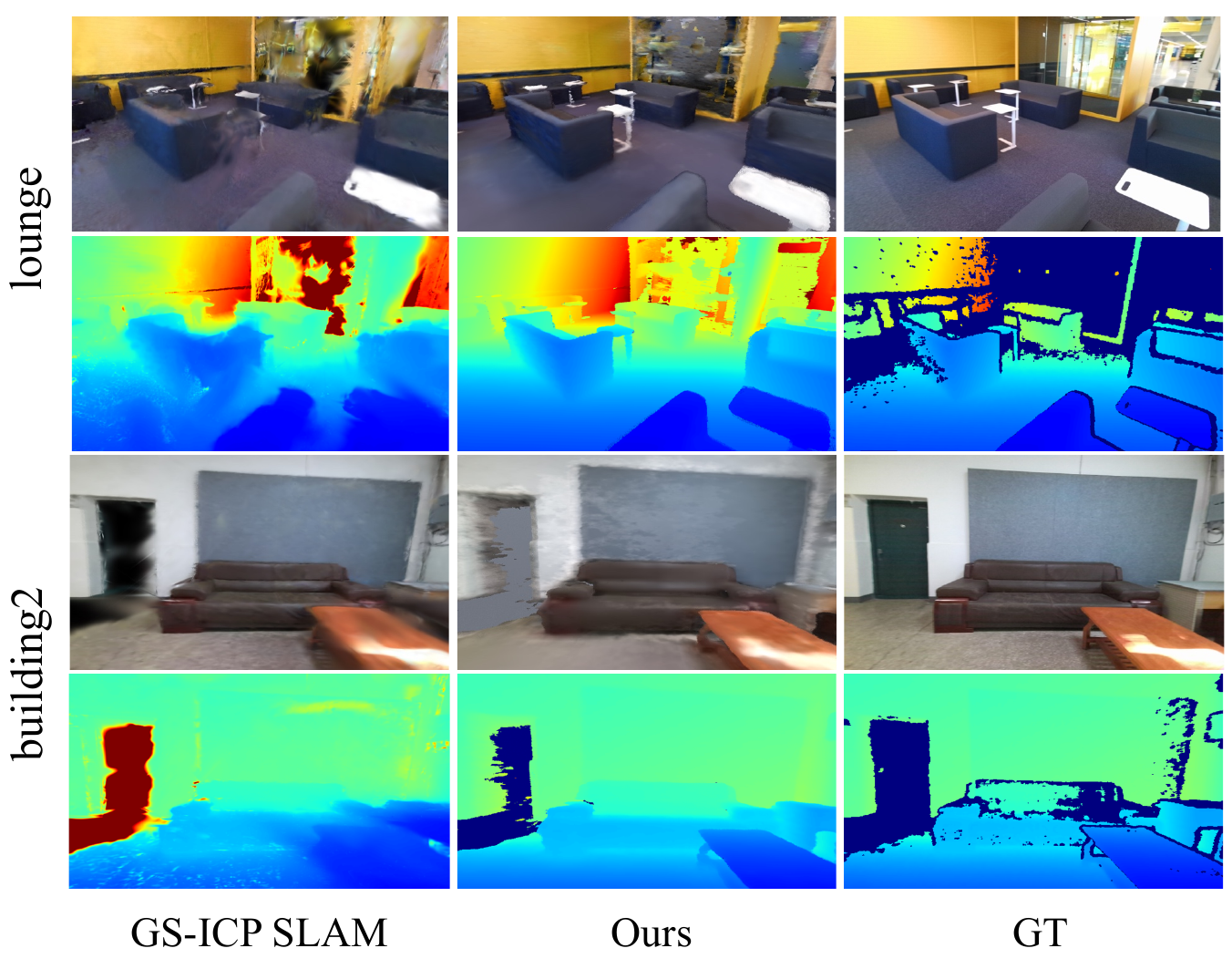}
\caption{Quality rendering comparison of training views using the estimated poses on \texttt{BS3D} (top) and \texttt{FastCaMo-Large} (bottom).} 
\Description{Figure 1. Fully described in the text.}
\label{fig:vis-2d rendering-sup}       
\end{figure}

\paragraph{Mapping and Bundle Adjustment.} Similar to Co-SLAM~\cite{wang2023co}, every $K=5$ frames are selected to be keyframes and only $\eta=0.05 $ percent of pixels are added to the keyframe database $\mathcal{K} $. Based on $\mathcal{K}$, we sample $N_m=2048$ pixels and $N_r=59$ points along the corresponding rays per iteration. Different from the NeRF-based SLAM~\cite{sucar2021imap,zhu2022nice}, we pay more attention to the neighborhood of zero-crossing surfaces. Specifically, $M_u$  uniform samples and $M_i$ importance samples are adopted for both bundle adjustment and mapping. Note that $M_u$ and $M_i$ are set to 11 and 48 for all datasets. We perform joint bundle adjustment for camera poses of keyframes and scene representation in an alternative way with 5 iterations for each. 
We propose to use an \fix{adaptive gradient amplification} to help the residual-based bundle adjustment escape local minima, and the corresponding algorithm is illustrated in Algorithm~\ref{algorithm:ga}. The residual-based mixture of explicit and implicit representations enhance the efficiency and performance of both reconstruction and bundle adjustment.

\paragraph{Experimental Setting.} The details for different benchmarks are reported here. We first introduce the mesh culling strategies used in the 3D reconstruction benchmark. Implicit methods tend to predict lots of noise points in the free space, and it is unfair to compare 3D reconstructions, especially in large-scale real-world scenes. Therefore, we use a similar culling strategy with estimated camera poses following~\cite{zhu2022nice}. The only difference is that we use ground-truth depth images for filtering instead of rendered depth images, which is more appropriate for filtering the noise points. After the mesh culling, the output mesh is transformed with \emph{Iterative Closest Point (ICP)} before the evaluation. The evaluation is carried out 3 times and the average results, including accuracy, completenes,s and completeness ratio with 5$cm$ or 10$cm$ thresholds are reported. We use the estimated camera poses in all the reconstruction and rendering benchmarks except the comparison of mapping using ground-truth poses and the ablation studies of the residual-based mapping. Note that the 3DGS-based method can not output the mesh directly, so we follow 2DGS~\cite{Huang2DGS2024} and use the rendered RGB-D images with TSDF Fusion to produce the output meshes for comparison. Every 10 rendered images are utilized for this, except for the mapping with ground-truth poses. We use every 5 frames for mapping with ground-truth poses for GS-ICP SLAM. Additionally, the output of BAD-SLAM is not on the same scale as other alternatives. Therefore, alignment with the ground-truth mesh is necessary before the evaluation. Note that the ground-truth mesh of \texttt{waiting} of \texttt{BS3D} is not consistent with the observed RGB-D images, so we use TSDF Fusion with ground-truth camera poses to output the ground-truth mesh for this scene instead of using the mesh provided by LiDAR.



\begin{table}[htb]
  \renewcommand{\arraystretch}{1.5}
  \centering
\caption{
Comparing tracking accuracy (ATE in cm) average on \texttt{TUM RGB-D}. The first 6 methods are implicit methods, and the others are explicit methods.
}
\scalebox{1}{
\setlength{\tabcolsep}{2.0mm}{
\begin{tabular}{l|c|c|c|c}
    \toprule
        Methods  & fr1/desk & fr2/xyz & fr3/office & Avg. \\ \hline
        iMAP* & 7.2   & 2.1   & 9 & 6.1   \\ 
        NICE-SLAM & {2.7}  & 1.8  & {3}  & 2.5    \\ 
        Co-SLAM  & {2.7}  & 1.9  & {2.4}  &  {2.3}  \\ 
        MIPS-Fusion  & 3.0  & {1.4} & 4.6  & {3}    \\ 
        \fix{ESLAM} & {2.5}&{1.1}&{2.4}&{2.0}  \\
        \ffix{DROID-SLAM} & 2.2&1.4&1.8&1.8  \\ \hline 

        BAD-SLAM  & 1.7  & 1.1  & 1.7 & 1.5  \\ 
        Kintinuous  & 3.7  & 2.9  & 3 & 3.2  \\ 
        \fix{Point-SLAM} & {2.6}&{1.3}&{3.2}&{2.4}  \\
        \fix{SplaTAM} & {3.4}&{1.2}&{5.2}&{3.3}  \\
        \fix{RTG-SLAM} & {1.7}&{0.4}&{1.1}&{1.1}  \\
        \ffix{LoopSplat} &2.1&1.6&3.2&2.3 \\
        \ffix{Photo-SLAM} &2.6&	\textbf{0.3}&1.0&	1.3 \\
        GS-ICP SLAM  & 2.7  & 1.8  & 2.7 & 2.4  \\ 
        ORB-SLAM2  & \textbf{1.6}  & {0.4}  & \textbf{1.0} & \textbf{1.0}  \\ \hline
        \fix{Ours}  &  {2.3}  & {1.8}  & {2.4} &  {2.2}  \\ 
        \bottomrule
\end{tabular}
}}
\label{tab:app1}
\vspace{5pt}
\end{table}

\fix{
\paragraph{Details of baselines.} We use the official multi-process version of Co-SLAM~\cite{wang2023co} in all experiments. ESLAM~\cite{johari2023eslam} uses tri-planes as key representations. However, it still requires a large amount of GPU memory for the scenes of \text{BS3D} dataset. Thereby, we use images with half resolution as input for ESLAM. The explicit 3DGS-based methods like SplaTAM~\cite{keetha2024splatam} and MonoGS~\cite{matsuki2024gaussian} require a large amount of GPU memory, which is not available on an NVIDIA RTX 3090 Ti GPU with 24GB GPU memory. Therefore, for a fair comparison, we use downsampled RGB-D images as input. For SplaTAM, we use images with half resolution for tracking and one-fourth of resolution for Gaussian mapping. For MonoGS, we use images with half resolution as input. Nonetheless, the GPU memory consumption of these two methods is still larger than other methods. Note that, the performance of MonoGS on ScanNet is provided in LoopSplat~\cite{zhu2025_loopsplat}.  There are similar situations for RTG-SLAM~\cite{peng2024rtgslam}, which uses an efficient 3DGS representation to reduce memory consumption. However, the required GPU memory significantly increases for global optimization at the end of tracking, which exceeds 24GB on most sequences of \texttt{BS3d} dataset. Therefore, we use a modified version of RTG-SLAM to randomly sample 100 keyframes rather than all the keyframes when performing the final global optimization. Similar strategies are used for these methods on the uHumans2 dataset. Regarding the room-level datasets, the performance is referred from their original publication unless the results are not provided. We generally use the same configuration from the most similar dataset, where the configuration is given for those new datasets. For example, RTG-SLAM has not been tested on ScanNet, and we use the configuration of it on TUM RGB-D that is provided to test its performance on ScanNet. BunldeFusion~\cite{dai2017bundlefusion} fails on the uHumans2 dataset, primarily due to the front-end tracking.
}

\ffix{As for DROID-SLAM~\cite{teed2021droid}, we use the default RGB-D configuration. For LoopSplat~\cite{zhu2025_loopsplat}, we find that the configuration on ScanNet or Replica results in unstable camera tracking, and we adapt the configuration on Replica by increasing the number of tracking iterations to 200 and changing the number of points for a new submap (similar to ScanNet). In this way, we obtain basically reasonable results on BS3D, and LoopSplat succeeds on ~\texttt{foobar} and ~\texttt{lounge} of \texttt{BS3D}. As for Photo-SLAM~\cite{huang2024photo}, we perform additional experiments on ScanNet, BS3D and uHumans2 using the configuration of TUM RGB-D.} 

In terms of 2D rendering, we follow ~\cite{keetha2024splatam} to filter the pixels with invalid depth information. Metrics like PSNR, SSIM, LPIPS and Depth-L1 are reported. Estimated camera poses of different approaches are leveraged for rendering. Note that, for RemixFusion, we divide the rendering into two stages including rendering the objects within 5 meters of the camera and rendering the remaining parts if the current scene is too large. As for tracking performance, ATE RMSE is reported to measure the accuracy of estimated camera poses. Only the results of methods that succeed in finishing the whole sequence are reported. 



\paragraph{Network Architecture.} The input to our network is the sampled 3D points $x_i$ from ray-casting. Then the sampled points go through the hash embedding for parametric features. The initial values of the global explicit map are obtained through trilinear interpolation as well. With positional embedding, the features from hash grids and the explicit interpolated values are fed into the decoder $\mathcal{D}$. The decoder $\mathcal{D}$ consists of two MLPs, which are used for color and geometric prediction. There are 2 layers with a hidden dim of 32 for both two MLPs. The feature branch of geometric MLP is utilized to guide the learning of color MLP. As for the residual-based bundle adjustment, an MLP $\mathcal{M}_p$ consisting of 2 layers with the hidden dim of 256 is utilized.


\paragraph{Setting of the \fix{lightweight} version.} The lightweight version of our method is faster and still accurate. Specifically, we perform $N_\text{ro}=11$ iterations for randomized optimization. The residual-based mapping and bundle adjustment are performed with 3 and 5 iterations respectively every 100 frames. As for the sampling strategies, we use $M_u=11$ and $M_i=21$ samples for each selected pixel.

\begin{table*}[!t]\centering
  \renewcommand{\arraystretch}{1.25}
\caption{
\fix{Comparing tracking accuracy (ATE RMSE in cm) on 8 large-scale RGB-D sequences of \texttt{BS3D} using the front-end estimated camera poses from RemixFusion for initialization. The front-end tracking module for other methods is disabled, and only the bundle adjustment is leveraged to refine the estimated camera poses. '--' denotes that the tracking failed for the corresponding methods. }
}
\scalebox{1.0}{
\setlength{\tabcolsep}{4.4mm}{
\begin{tabular}{@{}l|c c c c c c c c | c}
    \toprule 
        Methods  & {cafeteria} & {corridor}  & {foobar} & {hub} & {juice} & {lounge} & {study} & {waiting} & Avg. \\ \hline
        baseline & 7.7&8.8&8.3&5.1&4.1&5.6&4.5&7.0&6.4 \\ \hline
        +Co-SLAM & 8.3&7.6&8.3&5.8&4.4&6.1&4.3&4.2&6.1 \\ 
        +ESLAM  & 7.6& 8.7& 8.2& 5.0& 4.2& 5.6& 4.7 & 6.9& 6.4 \\
        +RTG-SLAM &--&8.4&--&--&5.7&67.3&\textbf{2.7}&--&--\\ \hline 
        RemixFusion &\textbf{6.8} & \textbf{6.3} &\textbf{5.8} & \textbf{4.5} & \textbf{3.1} & \textbf{4.2} & {3.2} & \textbf{3.0} & \textbf{4.6} \\

    \bottomrule 
    
\end{tabular}
}}
\label{tab:BS3D-roinit}
\end{table*}

\begin{table*}[!t]
  \renewcommand{\arraystretch}{1.2}
  \setlength{\tabcolsep}{3.4mm}
  \centering
  \caption{Comparison of training view rendering performance on BS3D using the estimated camera poses. Every 10 frames of each scene in the dataset are evaluated and the pixels where the depth is missing are masked. Our method obtains the best geometric rendering performance and better photometric rendering results than all the implicit methods. '$\_$' denotes the second best method. Note that GS-ICP SLAM is the best in RGB rendering but the worst in depth rendering, indicating the overfitting of 2D training views and ignorance of the 3D geometry.}
  \scalebox{1.0}{
  \begin{tabular}{@{}l|c|ccccccccc@{}}
  \toprule
  {Methods}& Metrics & {cafeteria} & {corridor} & {foobar} & {hub} & {juice} & {lounge} & {study} & {waiting} & {Avg.}  \\
  \hline 

  \multirow{4}{*}{NICE-SLAM} & PSNR$\uparrow$ & 22.02&19.65&22.00&18.96&19.71&20.21&23.18&21.30&20.88 \\
    & SSIM$\uparrow$ & 0.967&0.940&0.958&0.908&0.914&0.926&0.965&0.940&0.940\\
    & LPIPS$\downarrow$ & 0.144&0.252&0.143&0.261&0.198&0.191&0.142&0.227&0.195 \\ 
    & D-L1$\downarrow$ & 0.405&0.301&0.236&0.099&0.056&0.110&0.149&0.029&0.173 \\ \hline

  \multirow{4}{*}{Co-SLAM} & PSNR$\uparrow$ & 26.45&\un{23.24}&26.85&\un{22.39}&22.82 & 24.83&25.87&22.21&24.33 \\
  & SSIM$\uparrow$ & 0.987 & \un{0.973}& 0.983& \un{0.955}& 0.967&0.968& 0.978& 0.950& 0.970 \\
  & LPIPS$\downarrow$ & 0.108&\un{0.195}&0.099&0.212&0.173&0.149&0.116&0.254&0.163 \\ 
  & D-L1$\downarrow$ & {0.100}&\textbf{0.022}&0.063&\textbf{0.031}&\un{0.031}&0.058&\textbf{0.021}&0.086&0.052 \\ \hline

  \multirow{4}{*}{MIPS-Fusion} & PSNR$\uparrow$ & 25.47&22.36&25.78&21.23&21.37&23.82&24.58&21.78&23.30 \\
  & SSIM$\uparrow$ & 0.983&0.964&0.979&0.942&0.956&0.957&0.968&0.948&0.962 \\
  & LPIPS$\downarrow$ & 0.116&0.249&0.117&0.263&0.201&0.178&0.161&0.260&	0.193 \\ 
  & D-L1$\downarrow$ & \un{0.095}&0.041&\un{0.029}&0.064&0.043&\un{0.040}&0.023&\textbf{0.019}&	\un{0.044} \\ \hline
  
  \multirow{4}{*}{\fix{ESLAM}} & PSNR$\uparrow$ & 24.09&20.86&23.31&21.2&	20.95&21.11&	25.17&22.58&22.41 \\
  & SSIM$\uparrow$ & 0.971&0.932&0.953&0.928&0.927&0.921&0.972&0.941&0.943 \\
  & LPIPS$\downarrow$ & 0.105&0.201&0.131&\un{0.174}&{0.156}&0.176&\un{0.092}&{0.166}&	\un{0.150} \\ 
  & D-L1$\downarrow$ & 0.140& 0.233& 0.106& 0.176& 0.097& 0.196& 0.108& 0.103&	0.145 \\ \hline

  \multirow{4}{*}{\fix{RTG-SLAM}} & PSNR$\uparrow$ & 25.01&22.06&25.63&22.29&{23.91}&23.83&25.68&22.03&23.81 \\
  & SSIM$\uparrow$ & 0.977&0.963&0.979&0.948&\un{0.974}&0.959&0.978&0.953&0.966 \\
  & LPIPS$\downarrow$ & 0.114&0.196&0.104&0.225&0.164&0.157&0.125&0.222&	0.163 \\ 
  & D-L1$\downarrow$ & 0.336&0.387&0.360&	0.245&0.206&0.282&0.267&0.213&	0.287 \\ \hline

  \multirow{4}{*}{\fix{SplaTAM}} & PSNR$\uparrow$ & 18.11&15.01&14.30&16.24&16.03&15.35&18.86&17.09&16.37 \\
  & SSIM$\uparrow$ & 0.806&0.653&0.734&0.719&0.736&0.690&0.820&	0.711&0.734 \\
  & LPIPS$\downarrow$ & 0.222&0.373&0.240&0.335&0.285&0.308&0.215&0.320&0.287 \\ 
  & D-L1$\downarrow$ &0.444&0.478&0.327&0.387&0.156&0.420&0.384&0.296&	0.362 \\ \hline

  \multirow{4}{*}{\fix{MonoGS}} & PSNR$\uparrow$ & 20.18&19.72&20.79&20.41&	22.40&17.53&18.77&19.62&19.93 \\
  & SSIM$\uparrow$ & 0.705&0.730&0.703&0.669&0.722&0.644&0.644&0.702&0.690 \\
  & LPIPS$\downarrow$ & 0.528&0.442&0.384&0.532&0.396&0.574&0.426&0.460&0.468 \\ 
  & D-L1$\downarrow$ &1.200&1.021&1.442&0.955&0.658&1.204&1.620&1.151&1.156\\ \hline

  \multirow{4}{*}{\ffix{Photo-SLAM}} & PSNR$\uparrow$ & 26.01&20.39&24.53&21.64&\un{24.29}&23.15&23.89&\un{23.70}&23.45 \\
  & SSIM$\uparrow$ & 0.982&0.947&0.967&0.949&0.973&0.953&0.962&\un{0.960}&0.962 \\
  & LPIPS$\downarrow$ & 0.106&0.222&0.106&0.212&\un{0.144}&0.160&0.143& \textbf{0.137}&0.154 \\ 
  & D-L1$\downarrow$ &0.894&0.817&2.173&0.593&1.442&0.699&0.954&1.111&1.085\\ \hline

  \multirow{4}{*}{GS-ICP SLAM} & PSNR$\uparrow$ & \textbf{28.20}&\textbf{24.14}&\textbf{27.80}&\textbf{24.77}&\textbf{25.23}&\textbf{27.14}&\textbf{27.38}&\textbf{25.21}&\textbf{26.23} \\
  & SSIM$\uparrow$ & \textbf{0.988}& \textbf{0.974}& \textbf{0.987}& \textbf{0.972}& \textbf{0.980} & \textbf{0.981}& \textbf{0.984}& \textbf{0.975}& \textbf{0.980} \\
  & LPIPS$\downarrow$ & \textbf{0.083}& \textbf{0.150}& \textbf{0.078}& \textbf{0.170}& \textbf{0.126} & \textbf{0.109}& \textbf{0.089}&\un{0.142}& \textbf{0.118} \\ 
  & D-L1$\downarrow$ & 0.317&0.216&0.329&0.231&0.160&0.245&0.146&0.141&	0.223 \\ \hline

  \multirow{4}{*}{RemixFusion} & PSNR$\uparrow$ & \un{26.83}&{22.97}&\un{27.39}&22.33&	{23.02}&\un{25.20}&\un{26.46}&{23.03}&\un{24.65} \\
  & SSIM$\uparrow$ & \un{0.987} &0.969&\un{0.984}&0.954&{0.967}&\un{0.970}&\un{0.980}&{0.958}&\un{0.971}\\
  & LPIPS$\downarrow$ & \un{0.102}&0.198&\un{0.097}&{0.209}&{0.172}&\un{0.142}&{0.115}&{0.193}&{0.154} \\ 
  & D-L1$\downarrow$ & \textbf{0.040}&\un{0.026}&\textbf{0.029}&\un{0.039}&\textbf{0.029}&\textbf{0.034}&\un{0.023}&\un{0.024}&\textbf{0.031} \\ 
  \bottomrule 
\end{tabular}
}
\label{tab:BS3D-rendering-sup}
\end{table*}

\begin{figure*}[!t]
  \centering 
    \begin{overpic}[width=0.95\linewidth]{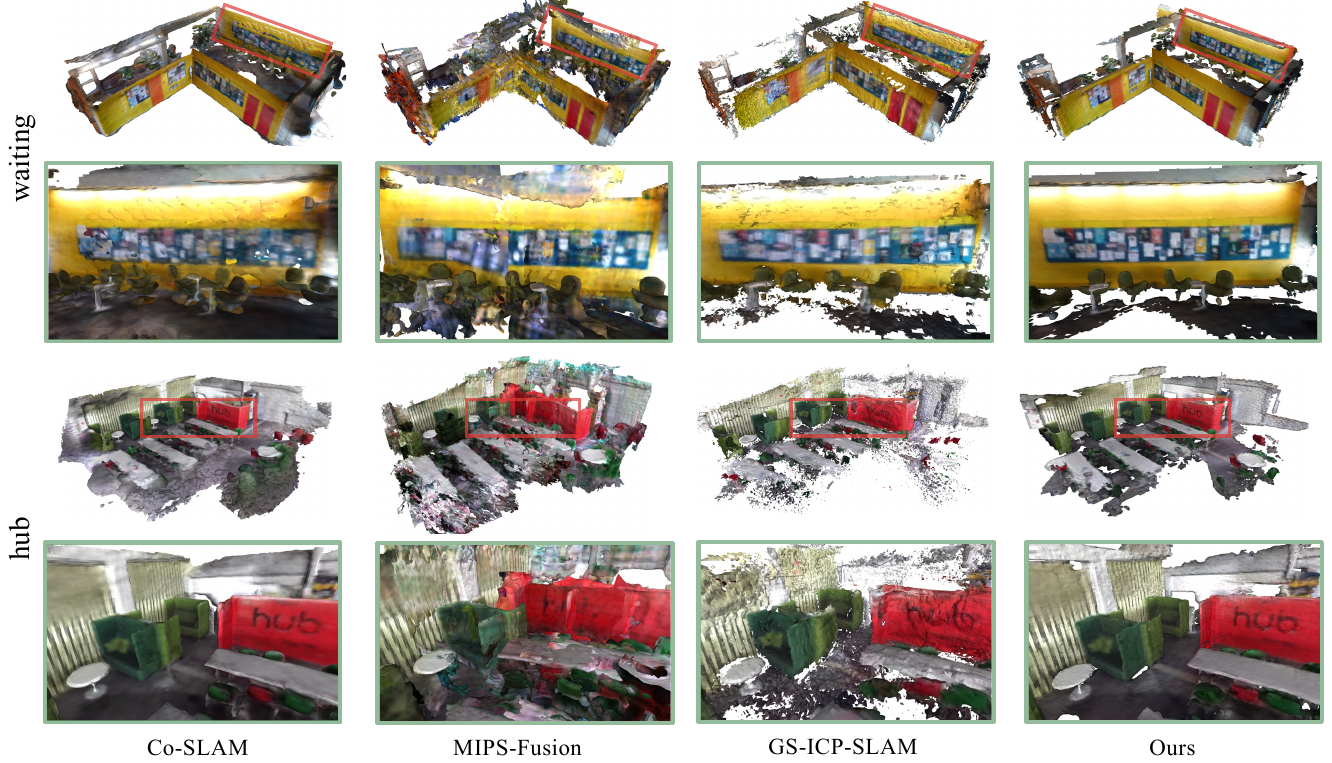}
    \end{overpic}
    \caption{Quality comparison of \texttt{corridor} on BS3D for different methods. Our method preserves detailed and sharper geometry and texture. The first row of each scene is the overview, and the second row is the zoom-in comparison corresponding to the area marked with red rectangles.} 
    \Description{Figure 2. Fully described in the text.}
    \label{fig:mesh-comp3}
\end{figure*}



\begin{table*}
  \renewcommand{\arraystretch}{1.3}
  \centering
\caption{
Comparing tracking accuracy (ATE RMSE in cm) on 10 sequences of \texttt{FastCaMo-Synth}. `--' denotes that the tracking failed for the corresponding methods. Our method is the most accurate and robust.
}
\scalebox{0.9}{
\setlength{\tabcolsep}{1.8mm}{
\begin{tabular}{l|c|c|c|c|c|c|c|c|c|c|c}
    \toprule
        Methods  & Apartment\_1 & Apartment\_2 & Frl\_apartment\_2 & Hotel\_0 & Office\_0 & Office\_1 & Office\_2 & Office\_3 & Room\_0 & Room\_1 & Avg. \\ \hline
        iMAP* & - & - & - & 20.3 & 39.2 & - & - & - & - & - & - \\ 
        NICE-SLAM & - & - & - & 4.2 & 8.4 & 13.7 & - & {14.3} & - & 29.7 & - \\ 
        Co-SLAM & 34.6 & 3.4 & 2.5 & 3.9 & {3} & 24.1 & 14.3 & 15.5 & 9.6 & 16.4 & 12.7 \\ 
        ESLAM & {1.8} & {1.9} & 2.3 & {1.4} & 7.9 & 20.2 & 68.2 & 35.3 & 30.8 & 84.4 & 25.4 \\ 
        MIPS-Fusion & 7.0 & 1.5 & {1.9} & 4.8 & 3.6 & {5.6} & \textbf{7.4} & 17.4 & {4.4} & {5.1} & {5.9} \\ 
        GS-ICP SLAM & 23.1&45.1&30.5&167.3&4.5&32.7	& 173.0&134.4&87.9&50.1&74.9 \\ \hline
        Ours & \textbf{1.9} &\textbf{1.1}&\textbf{1.1}&\textbf{0.6}&\textbf{1.4}&\textbf{3.0}&10.9&\textbf{5.8}&\textbf{1.3}&\textbf{3.2}&\textbf{3.0} \\ 
        \bottomrule
\end{tabular}
}}
\label{tab:fastsyn}
\end{table*}

\begin{table*}[t]
  \renewcommand{\arraystretch}{1.5}
  \centering
  \caption{Comparison of reconstruction accuracy (Acc.), completeness (Comp.) and completeness Ratio(\%) (Comp. Ratio(\%)) with 10$cm$ threshold for different methods using camera poses of RemixFusion on \texttt{BS3D}. Note that, results of RemixFusion reported here are the results of real-time SLAM, while the results of Co-SLAM are obtained by using the camera poses of RemixFusion directly, indicating more optimization iterations. Our method still outperforms Co-SLAM in terms of scene reconstruction. }
  \scalebox{1.0}{
  \setlength{\tabcolsep}{2.8mm}{
  \begin{tabular}{@{}l|l|ccccccccc}
  \toprule
  {Methods}& Metrics & {cafeteria} & {corridor} & {foobar} & {hub} & {juice} & {lounge} & {study} & {waiting} & {Avg.}  \\
  \hline 

  \multirow{3}{*}{Co-SLAM} & Acc.$\downarrow$ & 4.92&7.30&6.99&4.56&\textbf{4.32}&5.43	&3.66&\textbf{3.69} & 5.11 \\
  & Comp.$\downarrow$ &5.46&5.13&5.22&4.53&\textbf{3.50}&5.92&3.80&4.75& 4.79 \\
  & Comp. Ratio(\%)$\uparrow$ & \textbf{92.26}&92.85&91.53&94.64&98.19&90.09&93.75&94.62 & 93.49\\ \hline

  \multirow{3}{*}{RemixFusion} & Acc.$\downarrow$ & \textbf{4.88}&\textbf{6.90}&\textbf{5.93}&\textbf{3.80}&4.77&\textbf{5.06}&\textbf{3.59}&{3.88}&\textbf{4.85}\\
  & Comp.$\downarrow$ & \textbf{5.39}&\textbf{4.93}&\textbf{4.94}&\textbf{4.12}&3.58&\textbf{5.71}&\textbf{3.68}&\textbf{3.11}&\textbf{4.43}\\
  & Comp. Ratio(\%)$\uparrow$ & 92.13&\textbf{95.10}&\textbf{92.94}&\textbf{94.73}&\textbf{98.26}&\textbf{91.05}&\textbf{95.48}&\textbf{98.72}&\textbf{94.80}\\ \hline

\end{tabular}
}}
\vspace{-5pt}
\label{tab:BS3D-remix-posemesh}
\end{table*}

\begin{table*}[htb]\centering
  \caption{
  Ablation studies of 3 designs for the residual-based bundle adjustments. The average ATE RMSE (in cm) of 8 scenes on \texttt{BS3D} dataset is reported. RBA and \fix{TBA} denote the residual-based BA (RBA for short) and traditional BA that optimizes each camera pose independently ({TBA} for short), and OM denotes the {gradient amplification}.
  }
  \renewcommand{\arraystretch}{1.5}
  \scalebox{1.0}{
  \setlength{\tabcolsep}{3.6mm}{
  \begin{tabular}{c c c | ccccccccc}
      \toprule
          RBA & {TBA}  & {GA}  &  {cafeteria} & {corridor} & {foobar} & {hub} & {juice} & {lounge} & {study} & {waiting} & {Avg.} \\ \hline
           &  &    & 7.7&8.8&8.3&5.1&4.1&5.6&4.5&7.0   & 6.39 \\ 
          \checkmark &  & &7.4&6.4&7.7&5.0&4.1&5.5&4.3&5.0& 5.68 \\ 
            & \checkmark &  & 7.7& 7.1& 7.9& 4.8& 4.2& 5.3& 4.1& 4.8 & 5.74 \\ 
            & \checkmark & \checkmark & 8.1&6.7&5.8&5.8&\textbf{2.7}&5.0&3.4&3.6 & 5.14 \\ 
           \checkmark &  & \checkmark & \textbf{6.8}&\textbf{6.3}&\textbf{5.8}&\textbf{4.5}&3.1&\textbf{4.2}&\textbf{3.2}&\textbf{3.0} &  \textbf{4.61} \\ 
      \bottomrule
  \end{tabular}
  }}
\label{tab:ablation-ba-sup}
\end{table*}

\subsection{Quantitative Comparison}

\fix{
In this section, we analyse the qualitative comparison of tracking and mapping for different methods. As for room-level scenes, we retain comparable tracking performance on the Replica and TUM RGB-D datasets since our designs are mainly targeted at large-scale scenes. We report the detailed comparison of tracking performance on TUM RGB-D in Table~\ref{tab:app1}. Table~\ref{tab:fastsyn} demonstrates the ATE RMSE on room-level sequences of FastCaMo-Synth(noise-free). Our method achieves comparable performance to the alternatives on TUM RGB-D. Note that although GS-ICP SLAM is the best approach on Replica, which is the synthetic dataset, its performance on the real-world dataset significantly declines compared to other methods. This indicates that 3DGS-based methods are less generalizable in real-world scenes.

Note that sequences on the FastCaMo-Synth dataset are challenging due to the fast camera motion, while our method can estimate the camera pose robustly as well as preserve the reconstruction details. Table ~\ref{tab:fastsyn} shows the results of the same methods on 10 noise-free scenes of FastCaMo-Synth. iMAP* and NICE-SLAM failed in tracking on these scenes. Compared with the second-best method (MIPS-Fusion), our method achieves an improvement of more than 48 percent on average.

Table~\ref{tab:BS3D-roinit} shows the comparison results of different methods using the front-end estimated camera poses of RemixFusion as initialization. Only the global bundle adjustment of these methods, including Co-SLAM~\cite{wang2023co}, ESLAM~\cite{johari2023eslam}, and RTG-SLAM~\cite{peng2024rtgslam}, is turned on for evaluation of the bundle adjustment module. The global bundle adjustment of Co-SLAM and ESLAM can slightly improve the tracking accuracy of the baseline (results of the front-end poses of RemixFusion). RTG-SLAM leverages the backend optimization module from ORB-SLAM2~\cite{mur2017orb}, which is the best on \texttt{study}, but fails on 4 scenes. Our residual-based bundle adjustment is the best on average, proving the effectiveness of our method.
}

Table~\ref{tab:BS3D-remix-posemesh} illustrates the comparison of scene reconstruction using the camera poses of RemixFusion. Our method still surpasses Co-SLAM in terms of scene reconstruction. Note that Co-SLAM directly loads the optimized camera poses, and there is no pose refinement, which means there are more iterations for mapping for Co-SLAM in this table.

For the comparison of 2D rendering of training views, we present the details comparison of 8 scenes of \texttt{BS3D} in Table~\ref{tab:BS3D-rendering-sup}. Although GS-ICP SLAM is the best in photometric rendering, it is the worst for geometric rendering, which significantly falls behind other methods. This can be attributed to the overfitting of 2D images, and it could not reconstruct the correct 3D geometry of the scenes. In contrast, our method is comparable in terms of photometric rendering and surpasses other alternatives in geometric rendering.

Table~\ref{tab:ablation-ba-sup} shows the detailed ablation studies of the residual-based adjustment and {gradient amplification} on 8 scenes of \texttt{BS3D}. All types of bundle adjustment techniques improve the pose estimation. However, the improvement is marginal without the proposed {gradient amplification} in large-scale scenes, which proves the effectiveness of our methods.

\begin{figure*}[!t]
  \centering 
    \begin{overpic}[width=0.7\linewidth]{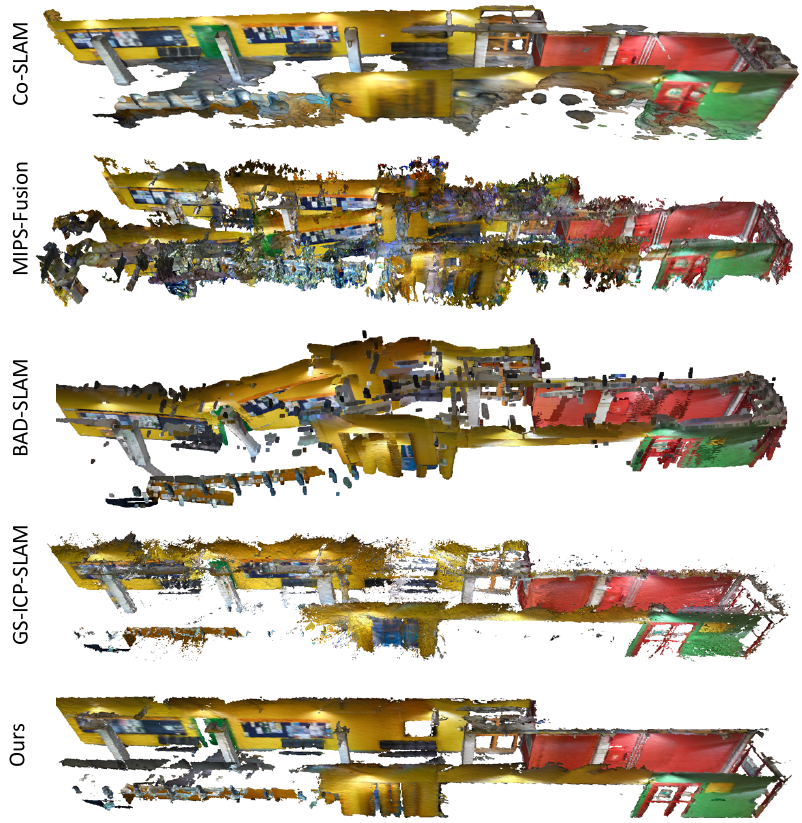}
    \end{overpic}
    \caption{The quality comparison of \texttt{corridor} on \texttt{BS3D} for different methods. Our method preserves detailed and complete geometry at the same time, while other methods struggle with real-time camera tracking and reconstruction.}
    \Description{Figure 3. Fully described in the text.}
    \label{fig:mesh-corridor}
\end{figure*}

\subsection{Qualitative Comparison}
We compare our method to several state-of-the-art methods for large-scale scenes on BS3D for scene reconstruction. Our methods can preserve the finest geometric details within limited time in these challenging large-scale scenes.
The compared alternatives suffer from fragile pose estimation and surface distortion, which emphasizes the effectiveness of our residual-based mixed representation. For example, the pictures on the wall in the second row of Figure~\ref{fig:mesh-comp3} reveal that our method can better preserve the photometric and geometric details in real time, while there are distortions or blurry surfaces for other methods like Co-SLAM~\cite{wang2023co} and GS-ICP SLAM~\cite{ha2024rgbd}. 
Figure~\ref{fig:mesh-corridor} shows the reconstruction comparison of a corridor whose length is more than 36$m$. Our method successfully reconstructs the whole scene, while others, excluding Co-SLAM, struggle in texture or geometry reconstruction.

Figure~\ref{fig:vis-2d rendering-sup} illustrates the 2D rendering results of our method and GS-ICP SLAM, which indicates that our method is more stable in photometric rendering and much better in geometric rendering than the 3DGS-based method. For example, our renderings of the black sofas and textures of the yellow walls in the first and second rows are more detailed and complete. We attribute this to the residual-based mixed representation of RemixFusion and the direct supervision of 3D space.

%% file: main.bbl

\begin{thebibliography}{61}


\ifx \showCODEN    \undefined \def \showCODEN     #1{\unskip}     \fi
\ifx \showDOI      \undefined \def \showDOI       #1{#1}\fi
\ifx \showISBNx    \undefined \def \showISBNx     #1{\unskip}     \fi
\ifx \showISBNxiii \undefined \def \showISBNxiii  #1{\unskip}     \fi
\ifx \showISSN     \undefined \def \showISSN      #1{\unskip}     \fi
\ifx \showLCCN     \undefined \def \showLCCN      #1{\unskip}     \fi
\ifx \shownote     \undefined \def \shownote      #1{#1}          \fi
\ifx \showarticletitle \undefined \def \showarticletitle #1{#1}   \fi
\ifx \showURL      \undefined \def \showURL       {\relax}        \fi
\providecommand\bibfield[2]{#2}
\providecommand\bibinfo[2]{#2}
\providecommand\natexlab[1]{#1}
\providecommand\showeprint[2][]{arXiv:#2}

\bibitem[Azinovi{\'c} et~al\mbox{.}(2022)]%
        {azinovic2022neural}
\bibfield{author}{\bibinfo{person}{Dejan Azinovi{\'c}}, \bibinfo{person}{Ricardo Martin-Brualla}, \bibinfo{person}{Dan~B Goldman}, \bibinfo{person}{Matthias Nie{\ss}ner}, {and} \bibinfo{person}{Justus Thies}.} \bibinfo{year}{2022}\natexlab{}.
\newblock \showarticletitle{Neural RGB-D surface reconstruction}. In \bibinfo{booktitle}{\emph{Proceedings of the IEEE/CVF Conference on Computer Vision and Pattern Recognition}}. \bibinfo{pages}{6290--6301}.
\newblock


\bibitem[Bian et~al\mbox{.}(2024)]%
        {bian2024porf}
\bibfield{author}{\bibinfo{person}{Jia-Wang Bian}, \bibinfo{person}{Wenjing Bian}, \bibinfo{person}{Victor~Adrian Prisacariu}, {and} \bibinfo{person}{Philip Torr}.} \bibinfo{year}{2024}\natexlab{}.
\newblock \showarticletitle{PoRF: Pose Residual Field for Accurate Neural Surface Reconstruction}. In \bibinfo{booktitle}{\emph{ICLR}}.
\newblock


\bibitem[Bylow et~al\mbox{.}(2013)]%
        {bylow2013real}
\bibfield{author}{\bibinfo{person}{Erik Bylow}, \bibinfo{person}{J{\"u}rgen Sturm}, \bibinfo{person}{Christian Kerl}, \bibinfo{person}{Fredrik Kahl}, {and} \bibinfo{person}{Daniel Cremers}.} \bibinfo{year}{2013}\natexlab{}.
\newblock \showarticletitle{Real-time camera tracking and 3D reconstruction using signed distance functions}. In \bibinfo{booktitle}{\emph{Robotics: Science and Systems}}, Vol.~\bibinfo{volume}{2}. \bibinfo{pages}{2}.
\newblock


\bibitem[Campos et~al\mbox{.}(2021)]%
        {campos2021orb}
\bibfield{author}{\bibinfo{person}{Carlos Campos}, \bibinfo{person}{Richard Elvira}, \bibinfo{person}{Juan J~G{\'o}mez Rodr{\'\i}guez}, \bibinfo{person}{Jos{\'e}~MM Montiel}, {and} \bibinfo{person}{Juan~D Tard{\'o}s}.} \bibinfo{year}{2021}\natexlab{}.
\newblock \showarticletitle{Orb-slam3: An accurate open-source library for visual, visual--inertial, and multimap slam}.
\newblock \bibinfo{journal}{\emph{IEEE Transactions on Robotics}} \bibinfo{volume}{37}, \bibinfo{number}{6} (\bibinfo{year}{2021}), \bibinfo{pages}{1874--1890}.
\newblock


\bibitem[Chan et~al\mbox{.}(2022)]%
        {chan2022efficient}
\bibfield{author}{\bibinfo{person}{Eric~R Chan}, \bibinfo{person}{Connor~Z Lin}, \bibinfo{person}{Matthew~A Chan}, \bibinfo{person}{Koki Nagano}, \bibinfo{person}{Boxiao Pan}, \bibinfo{person}{Shalini De~Mello}, \bibinfo{person}{Orazio Gallo}, \bibinfo{person}{Leonidas~J Guibas}, \bibinfo{person}{Jonathan Tremblay}, \bibinfo{person}{Sameh Khamis}, {et~al\mbox{.}}} \bibinfo{year}{2022}\natexlab{}.
\newblock \showarticletitle{Efficient geometry-aware 3D generative adversarial networks}. In \bibinfo{booktitle}{\emph{Proceedings of the IEEE/CVF Conference on Computer Vision and Pattern Recognition}}. \bibinfo{pages}{16123--16133}.
\newblock


\bibitem[Chen et~al\mbox{.}(2022)]%
        {chen2022tensorf}
\bibfield{author}{\bibinfo{person}{Anpei Chen}, \bibinfo{person}{Zexiang Xu}, \bibinfo{person}{Andreas Geiger}, \bibinfo{person}{Jingyi Yu}, {and} \bibinfo{person}{Hao Su}.} \bibinfo{year}{2022}\natexlab{}.
\newblock \showarticletitle{Tensorf: Tensorial radiance fields}. In \bibinfo{booktitle}{\emph{European conference on computer vision}}. Springer, \bibinfo{pages}{333--350}.
\newblock


\bibitem[Chung et~al\mbox{.}(2022)]%
        {chung2022orbeez}
\bibfield{author}{\bibinfo{person}{Chi-Ming Chung}, \bibinfo{person}{Yang-Che Tseng}, \bibinfo{person}{Ya-Ching Hsu}, \bibinfo{person}{Xiang-Qian Shi}, \bibinfo{person}{Yun-Hung Hua}, \bibinfo{person}{Jia-Fong Yeh}, \bibinfo{person}{Wen-Chin Chen}, \bibinfo{person}{Yi-Ting Chen}, {and} \bibinfo{person}{Winston~H Hsu}.} \bibinfo{year}{2022}\natexlab{}.
\newblock \showarticletitle{Orbeez-SLAM: A Real-time Monocular Visual SLAM with ORB Features and NeRF-realized Mapping}.
\newblock \bibinfo{journal}{\emph{arXiv preprint arXiv:2209.13274}} (\bibinfo{year}{2022}).
\newblock


\bibitem[Curless and Levoy(1996)]%
        {curless1996volumetric}
\bibfield{author}{\bibinfo{person}{Brian Curless} {and} \bibinfo{person}{Marc Levoy}.} \bibinfo{year}{1996}\natexlab{}.
\newblock \showarticletitle{A volumetric method for building complex models from range images}. In \bibinfo{booktitle}{\emph{Proceedings of the 23rd annual conference on Computer graphics and interactive techniques}}. \bibinfo{pages}{303--312}.
\newblock


\bibitem[Dai et~al\mbox{.}(2017a)]%
        {dai2017scannet}
\bibfield{author}{\bibinfo{person}{Angela Dai}, \bibinfo{person}{Angel~X Chang}, \bibinfo{person}{Manolis Savva}, \bibinfo{person}{Maciej Halber}, \bibinfo{person}{Thomas Funkhouser}, {and} \bibinfo{person}{Matthias Nie{\ss}ner}.} \bibinfo{year}{2017}\natexlab{a}.
\newblock \showarticletitle{Scannet: Richly-annotated 3d reconstructions of indoor scenes}. In \bibinfo{booktitle}{\emph{Proc. CVPR}}. \bibinfo{pages}{5828--5839}.
\newblock


\bibitem[Dai et~al\mbox{.}(2017b)]%
        {dai2017bundlefusion}
\bibfield{author}{\bibinfo{person}{Angela Dai}, \bibinfo{person}{Matthias Nie{\ss}ner}, \bibinfo{person}{Michael Zollh{\"o}fer}, \bibinfo{person}{Shahram Izadi}, {and} \bibinfo{person}{Christian Theobalt}.} \bibinfo{year}{2017}\natexlab{b}.
\newblock \showarticletitle{Bundlefusion: Real-time globally consistent 3d reconstruction using on-the-fly surface reintegration}.
\newblock \bibinfo{journal}{\emph{ACM Transactions on Graphics (ToG)}} \bibinfo{volume}{36}, \bibinfo{number}{4} (\bibinfo{year}{2017}), \bibinfo{pages}{1}.
\newblock


\bibitem[Dai et~al\mbox{.}(2017c)]%
        {Dai2017}
\bibfield{author}{\bibinfo{person}{Angela Dai}, \bibinfo{person}{Matthias Nie{\ss}ner}, \bibinfo{person}{Michael Zollh{\"o}fer}, \bibinfo{person}{Shahram Izadi}, {and} \bibinfo{person}{Christian Theobalt}.} \bibinfo{year}{2017}\natexlab{c}.
\newblock \showarticletitle{BundleFusion: Real-time Globally Consistent 3D Reconstruction using On-the-fly Surface Reintegration}.
\newblock \bibinfo{journal}{\emph{ACM Transactions on Graphics (TOG)}} \bibinfo{volume}{36}, \bibinfo{number}{3} (\bibinfo{year}{2017}), \bibinfo{pages}{24}.
\newblock


\bibitem[Ha et~al\mbox{.}(2024)]%
        {ha2024rgbd}
\bibfield{author}{\bibinfo{person}{Seongbo Ha}, \bibinfo{person}{Jiung Yeon}, {and} \bibinfo{person}{Hyeonwoo Yu}.} \bibinfo{year}{2024}\natexlab{}.
\newblock \showarticletitle{Rgbd gs-icp slam}. In \bibinfo{booktitle}{\emph{European Conference on Computer Vision}}. Springer, \bibinfo{pages}{180--197}.
\newblock


\bibitem[He et~al\mbox{.}(2016)]%
        {he2016deep}
\bibfield{author}{\bibinfo{person}{Kaiming He}, \bibinfo{person}{Xiangyu Zhang}, \bibinfo{person}{Shaoqing Ren}, {and} \bibinfo{person}{Jian Sun}.} \bibinfo{year}{2016}\natexlab{}.
\newblock \showarticletitle{Deep residual learning for image recognition}. In \bibinfo{booktitle}{\emph{Proceedings of the IEEE conference on computer vision and pattern recognition}}. \bibinfo{pages}{770--778}.
\newblock


\bibitem[Hu et~al\mbox{.}(2023)]%
        {hu2023cp}
\bibfield{author}{\bibinfo{person}{Jiarui Hu}, \bibinfo{person}{Mao Mao}, \bibinfo{person}{Hujun Bao}, \bibinfo{person}{Guofeng Zhang}, {and} \bibinfo{person}{Zhaopeng Cui}.} \bibinfo{year}{2023}\natexlab{}.
\newblock \showarticletitle{CP-SLAM: Collaborative Neural Point-based SLAM System}.
\newblock \bibinfo{journal}{\emph{arXiv preprint arXiv:2311.08013}} (\bibinfo{year}{2023}).
\newblock


\bibitem[Huang et~al\mbox{.}(2024b)]%
        {Huang2DGS2024}
\bibfield{author}{\bibinfo{person}{Binbin Huang}, \bibinfo{person}{Zehao Yu}, \bibinfo{person}{Anpei Chen}, \bibinfo{person}{Andreas Geiger}, {and} \bibinfo{person}{Shenghua Gao}.} \bibinfo{year}{2024}\natexlab{b}.
\newblock \showarticletitle{2D Gaussian Splatting for Geometrically Accurate Radiance Fields}. In \bibinfo{booktitle}{\emph{SIGGRAPH 2024 Conference Papers}}. \bibinfo{publisher}{Association for Computing Machinery}.
\newblock
\urldef\tempurl%
\url{https://doi.org/10.1145/3641519.3657428}
\showDOI{\tempurl}


\bibitem[Huang et~al\mbox{.}(2024a)]%
        {huang2024photo}
\bibfield{author}{\bibinfo{person}{Huajian Huang}, \bibinfo{person}{Longwei Li}, \bibinfo{person}{Hui Cheng}, {and} \bibinfo{person}{Sai-Kit Yeung}.} \bibinfo{year}{2024}\natexlab{a}.
\newblock \showarticletitle{Photo-SLAM: Real-time Simultaneous Localization and Photorealistic Mapping for Monocular Stereo and RGB-D Cameras}. In \bibinfo{booktitle}{\emph{Proceedings of the IEEE/CVF Conference on Computer Vision and Pattern Recognition}}. \bibinfo{pages}{21584--21593}.
\newblock


\bibitem[Izadi et~al\mbox{.}(2011)]%
        {Izadi2011}
\bibfield{author}{\bibinfo{person}{Shahram Izadi}, \bibinfo{person}{David Kim}, \bibinfo{person}{Otmar Hilliges}, \bibinfo{person}{David Molyneaux}, \bibinfo{person}{Richard Newcombe}, \bibinfo{person}{Pushmeet Kohli}, \bibinfo{person}{Jamie Shotton}, \bibinfo{person}{Steve Hodges}, \bibinfo{person}{Dustin Freeman}, \bibinfo{person}{Andrew Davison}, {and} \bibinfo{person}{Andrew Fitzgibbon}.} \bibinfo{year}{2011}\natexlab{}.
\newblock \showarticletitle{Kinect{F}usion: Real-time {3D} Reconstruction and Interaction Using a Moving Depth Camera}. In \bibinfo{booktitle}{\emph{UIST}}. \bibinfo{pages}{559--568}.
\newblock


\bibitem[Johari et~al\mbox{.}(2023)]%
        {johari2023eslam}
\bibfield{author}{\bibinfo{person}{Mohammad~Mahdi Johari}, \bibinfo{person}{Camilla Carta}, {and} \bibinfo{person}{Fran{\c{c}}ois Fleuret}.} \bibinfo{year}{2023}\natexlab{}.
\newblock \showarticletitle{ESLAM: Efficient Dense SLAM System Based on Hybrid Representation of Signed Distance Fields}. In \bibinfo{booktitle}{\emph{Proc. CVPR}}.
\newblock


\bibitem[Keetha et~al\mbox{.}(2024)]%
        {keetha2024splatam}
\bibfield{author}{\bibinfo{person}{Nikhil Keetha}, \bibinfo{person}{Jay Karhade}, \bibinfo{person}{Krishna~Murthy Jatavallabhula}, \bibinfo{person}{Gengshan Yang}, \bibinfo{person}{Sebastian Scherer}, \bibinfo{person}{Deva Ramanan}, {and} \bibinfo{person}{Jonathon Luiten}.} \bibinfo{year}{2024}\natexlab{}.
\newblock \showarticletitle{Splatam: Splat track \& map 3d gaussians for dense rgb-d slam}. In \bibinfo{booktitle}{\emph{Proceedings of the IEEE/CVF Conference on Computer Vision and Pattern Recognition}}. \bibinfo{pages}{21357--21366}.
\newblock


\bibitem[Keller et~al\mbox{.}(2013)]%
        {keller2013real}
\bibfield{author}{\bibinfo{person}{Maik Keller}, \bibinfo{person}{Damien Lefloch}, \bibinfo{person}{Martin Lambers}, \bibinfo{person}{Shahram Izadi}, \bibinfo{person}{Tim Weyrich}, {and} \bibinfo{person}{Andreas Kolb}.} \bibinfo{year}{2013}\natexlab{}.
\newblock \showarticletitle{Real-time 3d reconstruction in dynamic scenes using point-based fusion}. In \bibinfo{booktitle}{\emph{2013 International Conference on 3D Vision-3DV 2013}}. IEEE, \bibinfo{pages}{1--8}.
\newblock


\bibitem[Kerbl et~al\mbox{.}(2023)]%
        {kerbl20233d}
\bibfield{author}{\bibinfo{person}{Bernhard Kerbl}, \bibinfo{person}{Georgios Kopanas}, \bibinfo{person}{Thomas Leimk{\"u}hler}, {and} \bibinfo{person}{George Drettakis}.} \bibinfo{year}{2023}\natexlab{}.
\newblock \showarticletitle{3d gaussian splatting for real-time radiance field rendering}.
\newblock \bibinfo{journal}{\emph{ACM Transactions on Graphics}} \bibinfo{volume}{42}, \bibinfo{number}{4} (\bibinfo{year}{2023}), \bibinfo{pages}{1--14}.
\newblock


\bibitem[Kim et~al\mbox{.}(2016)]%
        {kim2016multimodal}
\bibfield{author}{\bibinfo{person}{Jin-Hwa Kim}, \bibinfo{person}{Sang-Woo Lee}, \bibinfo{person}{Donghyun Kwak}, \bibinfo{person}{Min-Oh Heo}, \bibinfo{person}{Jeonghee Kim}, \bibinfo{person}{Jung-Woo Ha}, {and} \bibinfo{person}{Byoung-Tak Zhang}.} \bibinfo{year}{2016}\natexlab{}.
\newblock \showarticletitle{Multimodal residual learning for visual qa}.
\newblock \bibinfo{journal}{\emph{Advances in neural information processing systems}}  \bibinfo{volume}{29} (\bibinfo{year}{2016}).
\newblock


\bibitem[{Kl{\"o}ckner} et~al\mbox{.}(2012)]%
        {kloeckner_pycuda_2012}
\bibfield{author}{\bibinfo{person}{Andreas {Kl{\"o}ckner}}, \bibinfo{person}{Nicolas {Pinto}}, \bibinfo{person}{Yunsup {Lee}}, \bibinfo{person}{B. {Catanzaro}}, \bibinfo{person}{Paul {Ivanov}}, {and} \bibinfo{person}{Ahmed {Fasih}}.} \bibinfo{year}{2012}\natexlab{}.
\newblock \showarticletitle{{PyCUDA and PyOpenCL: A Scripting-Based Approach to GPU Run-Time Code Generation}}.
\newblock \bibinfo{journal}{\emph{Parallel Comput.}} \bibinfo{volume}{38}, \bibinfo{number}{3} (\bibinfo{year}{2012}), \bibinfo{pages}{157--174}.
\newblock
\showISSN{0167-8191}
\urldef\tempurl%
\url{https://doi.org/10.1016/j.parco.2011.09.001}
\showDOI{\tempurl}


\bibitem[Koestler et~al\mbox{.}(2022)]%
        {koestler2022tandem}
\bibfield{author}{\bibinfo{person}{Lukas Koestler}, \bibinfo{person}{Nan Yang}, \bibinfo{person}{Niclas Zeller}, {and} \bibinfo{person}{Daniel Cremers}.} \bibinfo{year}{2022}\natexlab{}.
\newblock \showarticletitle{Tandem: Tracking and dense mapping in real-time using deep multi-view stereo}. In \bibinfo{booktitle}{\emph{Conference on Robot Learning}}. PMLR, \bibinfo{pages}{34--45}.
\newblock


\bibitem[Li et~al\mbox{.}(2023)]%
        {li2023neuralangelo}
\bibfield{author}{\bibinfo{person}{Zhaoshuo Li}, \bibinfo{person}{Thomas M{\"u}ller}, \bibinfo{person}{Alex Evans}, \bibinfo{person}{Russell~H Taylor}, \bibinfo{person}{Mathias Unberath}, \bibinfo{person}{Ming-Yu Liu}, {and} \bibinfo{person}{Chen-Hsuan Lin}.} \bibinfo{year}{2023}\natexlab{}.
\newblock \showarticletitle{Neuralangelo: High-fidelity neural surface reconstruction}. In \bibinfo{booktitle}{\emph{Proceedings of the IEEE/CVF Conference on Computer Vision and Pattern Recognition}}. \bibinfo{pages}{8456--8465}.
\newblock


\bibitem[Liu et~al\mbox{.}(2020)]%
        {liu2020neural}
\bibfield{author}{\bibinfo{person}{Lingjie Liu}, \bibinfo{person}{Jiatao Gu}, \bibinfo{person}{Kyaw Zaw~Lin}, \bibinfo{person}{Tat-Seng Chua}, {and} \bibinfo{person}{Christian Theobalt}.} \bibinfo{year}{2020}\natexlab{}.
\newblock \showarticletitle{Neural sparse voxel fields}.
\newblock \bibinfo{journal}{\emph{Advances in Neural Information Processing Systems}}  \bibinfo{volume}{33} (\bibinfo{year}{2020}), \bibinfo{pages}{15651--15663}.
\newblock


\bibitem[Mao et~al\mbox{.}(2023)]%
        {mao2023ngel}
\bibfield{author}{\bibinfo{person}{Yunxuan Mao}, \bibinfo{person}{Xuan Yu}, \bibinfo{person}{Kai Wang}, \bibinfo{person}{Yue Wang}, \bibinfo{person}{Rong Xiong}, {and} \bibinfo{person}{Yiyi Liao}.} \bibinfo{year}{2023}\natexlab{}.
\newblock \showarticletitle{NGEL-SLAM: Neural Implicit Representation-based Global Consistent Low-Latency SLAM System}.
\newblock \bibinfo{journal}{\emph{arXiv preprint arXiv:2311.09525}} (\bibinfo{year}{2023}).
\newblock


\bibitem[Matsuki et~al\mbox{.}(2024)]%
        {matsuki2024gaussian}
\bibfield{author}{\bibinfo{person}{Hidenobu Matsuki}, \bibinfo{person}{Riku Murai}, \bibinfo{person}{Paul~HJ Kelly}, {and} \bibinfo{person}{Andrew~J Davison}.} \bibinfo{year}{2024}\natexlab{}.
\newblock \showarticletitle{Gaussian splatting slam}. In \bibinfo{booktitle}{\emph{Proceedings of the IEEE/CVF Conference on Computer Vision and Pattern Recognition}}. \bibinfo{pages}{18039--18048}.
\newblock


\bibitem[Mildenhall et~al\mbox{.}(2021)]%
        {mildenhall2021nerf}
\bibfield{author}{\bibinfo{person}{Ben Mildenhall}, \bibinfo{person}{Pratul~P Srinivasan}, \bibinfo{person}{Matthew Tancik}, \bibinfo{person}{Jonathan~T Barron}, \bibinfo{person}{Ravi Ramamoorthi}, {and} \bibinfo{person}{Ren Ng}.} \bibinfo{year}{2021}\natexlab{}.
\newblock \showarticletitle{Nerf: Representing scenes as neural radiance fields for view synthesis}.
\newblock \bibinfo{journal}{\emph{Commun. ACM}} \bibinfo{volume}{65}, \bibinfo{number}{1} (\bibinfo{year}{2021}), \bibinfo{pages}{99--106}.
\newblock


\bibitem[M{\"u}ller et~al\mbox{.}(2022)]%
        {muller2022instant}
\bibfield{author}{\bibinfo{person}{Thomas M{\"u}ller}, \bibinfo{person}{Alex Evans}, \bibinfo{person}{Christoph Schied}, {and} \bibinfo{person}{Alexander Keller}.} \bibinfo{year}{2022}\natexlab{}.
\newblock \showarticletitle{Instant neural graphics primitives with a multiresolution hash encoding}.
\newblock \bibinfo{journal}{\emph{ACM Transactions on Graphics (ToG)}} \bibinfo{volume}{41}, \bibinfo{number}{4} (\bibinfo{year}{2022}), \bibinfo{pages}{1--15}.
\newblock


\bibitem[M{\"u}ller et~al\mbox{.}(2019)]%
        {muller2019neural}
\bibfield{author}{\bibinfo{person}{Thomas M{\"u}ller}, \bibinfo{person}{Brian McWilliams}, \bibinfo{person}{Fabrice Rousselle}, \bibinfo{person}{Markus Gross}, {and} \bibinfo{person}{Jan Nov{\'a}k}.} \bibinfo{year}{2019}\natexlab{}.
\newblock \showarticletitle{Neural importance sampling}.
\newblock \bibinfo{journal}{\emph{ACM Transactions on Graphics (ToG)}} \bibinfo{volume}{38}, \bibinfo{number}{5} (\bibinfo{year}{2019}), \bibinfo{pages}{1--19}.
\newblock


\bibitem[Mur-Artal and Tard{\'o}s(2017)]%
        {mur2017orb}
\bibfield{author}{\bibinfo{person}{Raul Mur-Artal} {and} \bibinfo{person}{Juan~D Tard{\'o}s}.} \bibinfo{year}{2017}\natexlab{}.
\newblock \showarticletitle{Orb-slam2: An open-source slam system for monocular, stereo, and rgb-d cameras}.
\newblock \bibinfo{journal}{\emph{IEEE Transactions on Robotics}} \bibinfo{volume}{33}, \bibinfo{number}{5} (\bibinfo{year}{2017}), \bibinfo{pages}{1255--1262}.
\newblock


\bibitem[Mustaniemi et~al\mbox{.}(2023)]%
        {mustaniemi2023bs3d}
\bibfield{author}{\bibinfo{person}{Janne Mustaniemi}, \bibinfo{person}{Juho Kannala}, \bibinfo{person}{Esa Rahtu}, \bibinfo{person}{Li Liu}, {and} \bibinfo{person}{Janne Heikkil{\"a}}.} \bibinfo{year}{2023}\natexlab{}.
\newblock \showarticletitle{BS3D: Building-Scale 3D Reconstruction from RGB-D Images}. In \bibinfo{booktitle}{\emph{Scandinavian Conference on Image Analysis}}. Springer, \bibinfo{pages}{551--565}.
\newblock


\bibitem[Nie{\ss}ner et~al\mbox{.}(2013a)]%
        {niessner2013real}
\bibfield{author}{\bibinfo{person}{Matthias Nie{\ss}ner}, \bibinfo{person}{Michael Zollh{\"o}fer}, \bibinfo{person}{Shahram Izadi}, {and} \bibinfo{person}{Marc Stamminger}.} \bibinfo{year}{2013}\natexlab{a}.
\newblock \showarticletitle{Real-time 3D reconstruction at scale using voxel hashing}.
\newblock \bibinfo{journal}{\emph{ACM Transactions on Graphics (ToG)}} \bibinfo{volume}{32}, \bibinfo{number}{6} (\bibinfo{year}{2013}), \bibinfo{pages}{1--11}.
\newblock


\bibitem[Nie{\ss}ner et~al\mbox{.}(2013b)]%
        {Niessner2013}
\bibfield{author}{\bibinfo{person}{M. Nie{\ss}ner}, \bibinfo{person}{M. Zollh\"ofer}, \bibinfo{person}{S. Izadi}, {and} \bibinfo{person}{M. Stamminger}.} \bibinfo{year}{2013}\natexlab{b}.
\newblock \showarticletitle{Real-time {3D} Reconstruction at Scale using Voxel Hashing}.
\newblock \bibinfo{journal}{\emph{ACM Trans. on Graph. (SIGGRAPH Asia)}} \bibinfo{volume}{32}, \bibinfo{number}{6} (\bibinfo{year}{2013}), \bibinfo{pages}{169}.
\newblock


\bibitem[Peng et~al\mbox{.}(2024)]%
        {peng2024rtgslam}
\bibfield{author}{\bibinfo{person}{Zhexi Peng}, \bibinfo{person}{Tianjia Shao}, \bibinfo{person}{Liu Yong}, \bibinfo{person}{Jingke Zhou}, \bibinfo{person}{Yin Yang}, \bibinfo{person}{Jingdong Wang}, {and} \bibinfo{person}{Kun Zhou}.} \bibinfo{year}{2024}\natexlab{}.
\newblock \showarticletitle{RTG-SLAM: Real-time 3D Reconstruction at Scale using Gaussian Splatting}.
\newblock  (\bibinfo{year}{2024}).
\newblock


\bibitem[Qin et~al\mbox{.}(2022)]%
        {qin2022geometric}
\bibfield{author}{\bibinfo{person}{Zheng Qin}, \bibinfo{person}{Hao Yu}, \bibinfo{person}{Changjian Wang}, \bibinfo{person}{Yulan Guo}, \bibinfo{person}{Yuxing Peng}, {and} \bibinfo{person}{Kai Xu}.} \bibinfo{year}{2022}\natexlab{}.
\newblock \showarticletitle{Geometric transformer for fast and robust point cloud registration}. In \bibinfo{booktitle}{\emph{Proceedings of the IEEE/CVF conference on computer vision and pattern recognition}}. \bibinfo{pages}{11143--11152}.
\newblock


\bibitem[Rosinol et~al\mbox{.}(2020)]%
        {Rosinol20icra-Kimera}
\bibfield{author}{\bibinfo{person}{Antoni Rosinol}, \bibinfo{person}{Marcus Abate}, \bibinfo{person}{Yun Chang}, {and} \bibinfo{person}{Luca Carlone}.} \bibinfo{year}{2020}\natexlab{}.
\newblock \showarticletitle{Kimera: an Open-Source Library for Real-Time Metric-Semantic Localization and Mapping}. In \bibinfo{booktitle}{\emph{IEEE Intl. Conf. on Robotics and Automation (ICRA)}}.
\newblock
\urldef\tempurl%
\url{https://github.com/MIT-SPARK/Kimera}
\showURL{%
\tempurl}


\bibitem[Roth and Vona(2012)]%
        {Roth2012}
\bibfield{author}{\bibinfo{person}{Henry Roth} {and} \bibinfo{person}{Marsette Vona}.} \bibinfo{year}{2012}\natexlab{}.
\newblock \showarticletitle{Moving Volume {KinectFusion}}. In \bibinfo{booktitle}{\emph{Proc. BMVC}}. \bibinfo{pages}{112:1--112:11}.
\newblock


\bibitem[Sandstr{\"o}m et~al\mbox{.}(2023)]%
        {Sandstrm2023PointSLAMDN}
\bibfield{author}{\bibinfo{person}{Erik Sandstr{\"o}m}, \bibinfo{person}{Yue Li}, \bibinfo{person}{Luc~Van Gool}, {and} \bibinfo{person}{Martin~R. Oswald}.} \bibinfo{year}{2023}\natexlab{}.
\newblock \showarticletitle{Point-SLAM: Dense Neural Point Cloud-based SLAM}.
\newblock \bibinfo{journal}{\emph{ArXiv}}  \bibinfo{volume}{abs/2304.04278} (\bibinfo{year}{2023}).
\newblock
\urldef\tempurl%
\url{https://api.semanticscholar.org/CorpusID:258049300}
\showURL{%
\tempurl}


\bibitem[Schops et~al\mbox{.}(2019)]%
        {schops2019bad}
\bibfield{author}{\bibinfo{person}{Thomas Schops}, \bibinfo{person}{Torsten Sattler}, {and} \bibinfo{person}{Marc Pollefeys}.} \bibinfo{year}{2019}\natexlab{}.
\newblock \showarticletitle{{BAD SLAM}: Bundle adjusted direct rgb-d slam}. In \bibinfo{booktitle}{\emph{Proceedings of the IEEE/CVF Conference on Computer Vision and Pattern Recognition}}. \bibinfo{pages}{134--144}.
\newblock


\bibitem[Straub et~al\mbox{.}(2019)]%
        {straub2019replica}
\bibfield{author}{\bibinfo{person}{Julian Straub}, \bibinfo{person}{Thomas Whelan}, \bibinfo{person}{Lingni Ma}, \bibinfo{person}{Yufan Chen}, \bibinfo{person}{Erik Wijmans}, \bibinfo{person}{Simon Green}, \bibinfo{person}{Jakob~J Engel}, \bibinfo{person}{Raul Mur-Artal}, \bibinfo{person}{Carl Ren}, \bibinfo{person}{Shobhit Verma}, {et~al\mbox{.}}} \bibinfo{year}{2019}\natexlab{}.
\newblock \showarticletitle{The Replica dataset: A digital replica of indoor spaces}.
\newblock \bibinfo{journal}{\emph{arXiv preprint arXiv:1906.05797}} (\bibinfo{year}{2019}).
\newblock


\bibitem[Sturm et~al\mbox{.}(2012)]%
        {sturm2012benchmark}
\bibfield{author}{\bibinfo{person}{J{\"u}rgen Sturm}, \bibinfo{person}{Nikolas Engelhard}, \bibinfo{person}{Felix Endres}, \bibinfo{person}{Wolfram Burgard}, {and} \bibinfo{person}{Daniel Cremers}.} \bibinfo{year}{2012}\natexlab{}.
\newblock \showarticletitle{A benchmark for the evaluation of RGB-D SLAM systems}. In \bibinfo{booktitle}{\emph{2012 IEEE/RSJ International Conference on Intelligent Robots and Systems}}. IEEE, \bibinfo{pages}{573--580}.
\newblock


\bibitem[Sucar et~al\mbox{.}(2021)]%
        {sucar2021imap}
\bibfield{author}{\bibinfo{person}{Edgar Sucar}, \bibinfo{person}{Shikun Liu}, \bibinfo{person}{Joseph Ortiz}, {and} \bibinfo{person}{Andrew~J Davison}.} \bibinfo{year}{2021}\natexlab{}.
\newblock \showarticletitle{iMAP: Implicit mapping and positioning in real-time}. In \bibinfo{booktitle}{\emph{Proceedings of the IEEE/CVF International Conference on Computer Vision}}. \bibinfo{pages}{6229--6238}.
\newblock


\bibitem[Tancik et~al\mbox{.}(2022)]%
        {tancik2022block}
\bibfield{author}{\bibinfo{person}{Matthew Tancik}, \bibinfo{person}{Vincent Casser}, \bibinfo{person}{Xinchen Yan}, \bibinfo{person}{Sabeek Pradhan}, \bibinfo{person}{Ben Mildenhall}, \bibinfo{person}{Pratul~P Srinivasan}, \bibinfo{person}{Jonathan~T Barron}, {and} \bibinfo{person}{Henrik Kretzschmar}.} \bibinfo{year}{2022}\natexlab{}.
\newblock \showarticletitle{Block-nerf: Scalable large scene neural view synthesis}. In \bibinfo{booktitle}{\emph{Proc. CVPR}}. \bibinfo{pages}{8248--8258}.
\newblock


\bibitem[Tang et~al\mbox{.}(2023)]%
        {tang2023mips}
\bibfield{author}{\bibinfo{person}{Yijie Tang}, \bibinfo{person}{Jiazhao Zhang}, \bibinfo{person}{Zhinan Yu}, \bibinfo{person}{He Wang}, {and} \bibinfo{person}{Kai Xu}.} \bibinfo{year}{2023}\natexlab{}.
\newblock \showarticletitle{Mips-fusion: Multi-implicit-submaps for scalable and robust online neural rgb-d reconstruction}.
\newblock \bibinfo{journal}{\emph{ACM Transactions on Graphics (TOG)}} \bibinfo{volume}{42}, \bibinfo{number}{6} (\bibinfo{year}{2023}), \bibinfo{pages}{1--16}.
\newblock


\bibitem[Teed and Deng(2021)]%
        {teed2021droid}
\bibfield{author}{\bibinfo{person}{Zachary Teed} {and} \bibinfo{person}{Jia Deng}.} \bibinfo{year}{2021}\natexlab{}.
\newblock \showarticletitle{Droid-slam: Deep visual slam for monocular, stereo, and rgb-d cameras}.
\newblock \bibinfo{journal}{\emph{Advances in Neural Information Processing Systems}}  \bibinfo{volume}{34} (\bibinfo{year}{2021}), \bibinfo{pages}{16558--16569}.
\newblock


\bibitem[Wang et~al\mbox{.}(2023)]%
        {wang2023co}
\bibfield{author}{\bibinfo{person}{Hengyi Wang}, \bibinfo{person}{Jingwen Wang}, {and} \bibinfo{person}{Lourdes Agapito}.} \bibinfo{year}{2023}\natexlab{}.
\newblock \showarticletitle{Co-SLAM: Joint Coordinate and Sparse Parametric Encodings for Neural Real-Time SLAM}. In \bibinfo{booktitle}{\emph{Proceedings of the IEEE/CVF Conference on Computer Vision and Pattern Recognition}}. \bibinfo{pages}{13293--13302}.
\newblock


\bibitem[Wang et~al\mbox{.}(2021)]%
        {wang2021neus}
\bibfield{author}{\bibinfo{person}{Peng Wang}, \bibinfo{person}{Lingjie Liu}, \bibinfo{person}{Yuan Liu}, \bibinfo{person}{Christian Theobalt}, \bibinfo{person}{Taku Komura}, {and} \bibinfo{person}{Wenping Wang}.} \bibinfo{year}{2021}\natexlab{}.
\newblock \showarticletitle{NeuS: Learning Neural Implicit Surfaces by Volume Rendering for Multi-view Reconstruction}.
\newblock \bibinfo{journal}{\emph{arXiv preprint arXiv:2106.10689}} (\bibinfo{year}{2021}).
\newblock


\bibitem[Whelan et~al\mbox{.}(2012)]%
        {Whelan2012}
\bibfield{author}{\bibinfo{person}{Thomas Whelan}, \bibinfo{person}{Michael Kaess}, \bibinfo{person}{Maurice Fallon}, \bibinfo{person}{Hordur Johannsson}, \bibinfo{person}{John Leonard}, {and} \bibinfo{person}{John McDonald}.} \bibinfo{year}{2012}\natexlab{}.
\newblock \showarticletitle{Kintinuous: Spatially Extended {KinectFusion}}. In \bibinfo{booktitle}{\emph{RSS Workshop on RGB-D: Advanced Reasoning with Depth Cameras}}.
\newblock


\bibitem[Whelan et~al\mbox{.}(2015)]%
        {Whelan2015}
\bibfield{author}{\bibinfo{person}{Thomas Whelan}, \bibinfo{person}{Stefan Leutenegger}, \bibinfo{person}{Renato~F Salas-Moreno}, \bibinfo{person}{Ben Glocker}, {and} \bibinfo{person}{Andrew~J Davison}.} \bibinfo{year}{2015}\natexlab{}.
\newblock \showarticletitle{ElasticFusion: Dense SLAM without a pose graph}. In \bibinfo{booktitle}{\emph{Proc. Robotics: Science and Systems}}.
\newblock


\bibitem[Whelan et~al\mbox{.}(2016)]%
        {Whelan2016ElasticFusionRD}
\bibfield{author}{\bibinfo{person}{Thomas Whelan}, \bibinfo{person}{Renato~F. Salas-Moreno}, \bibinfo{person}{Ben Glocker}, \bibinfo{person}{Andrew~J. Davison}, {and} \bibinfo{person}{Stefan Leutenegger}.} \bibinfo{year}{2016}\natexlab{}.
\newblock \showarticletitle{ElasticFusion: Real-time dense SLAM and light source estimation}.
\newblock \bibinfo{journal}{\emph{The International Journal of Robotics Research}}  \bibinfo{volume}{35} (\bibinfo{year}{2016}), \bibinfo{pages}{1697 -- 1716}.
\newblock
\urldef\tempurl%
\url{https://api.semanticscholar.org/CorpusID:21124365}
\showURL{%
\tempurl}


\bibitem[Xiangli et~al\mbox{.}(2022)]%
        {xiangli2022bungeenerf}
\bibfield{author}{\bibinfo{person}{Yuanbo Xiangli}, \bibinfo{person}{Linning Xu}, \bibinfo{person}{Xingang Pan}, \bibinfo{person}{Nanxuan Zhao}, \bibinfo{person}{Anyi Rao}, \bibinfo{person}{Christian Theobalt}, \bibinfo{person}{Bo Dai}, {and} \bibinfo{person}{Dahua Lin}.} \bibinfo{year}{2022}\natexlab{}.
\newblock \showarticletitle{Bungeenerf: Progressive neural radiance field for extreme multi-scale scene rendering}. In \bibinfo{booktitle}{\emph{European conference on computer vision}}. Springer, \bibinfo{pages}{106--122}.
\newblock


\bibitem[Xu et~al\mbox{.}(2022b)]%
        {Xu2022PointNeRFPN}
\bibfield{author}{\bibinfo{person}{Qiangeng Xu}, \bibinfo{person}{Zexiang Xu}, \bibinfo{person}{Julien Philip}, \bibinfo{person}{Sai Bi}, \bibinfo{person}{Zhixin Shu}, \bibinfo{person}{Kalyan Sunkavalli}, {and} \bibinfo{person}{Ulrich Neumann}.} \bibinfo{year}{2022}\natexlab{b}.
\newblock \showarticletitle{Point-NeRF: Point-based Neural Radiance Fields}.
\newblock \bibinfo{journal}{\emph{2022 IEEE/CVF Conference on Computer Vision and Pattern Recognition (CVPR)}} (\bibinfo{year}{2022}), \bibinfo{pages}{5428--5438}.
\newblock
\urldef\tempurl%
\url{https://api.semanticscholar.org/CorpusID:246210101}
\showURL{%
\tempurl}


\bibitem[Xu et~al\mbox{.}(2022a)]%
        {Xu2022HRBFFusionA3}
\bibfield{author}{\bibinfo{person}{Yabin Xu}, \bibinfo{person}{Liangliang Nan}, \bibinfo{person}{Laishui Zhou}, \bibinfo{person}{Jun Wang}, {and} \bibinfo{person}{Charlie~C.L. Wang}.} \bibinfo{year}{2022}\natexlab{a}.
\newblock \showarticletitle{HRBF-Fusion: Accurate 3D Reconstruction from RGB-D Data Using On-the-fly Implicits}.
\newblock \bibinfo{journal}{\emph{ACM Transactions on Graphics (TOG)}}  \bibinfo{volume}{41} (\bibinfo{year}{2022}), \bibinfo{pages}{1 -- 19}.
\newblock
\urldef\tempurl%
\url{https://api.semanticscholar.org/CorpusID:246608194}
\showURL{%
\tempurl}


\bibitem[Yang et~al\mbox{.}(2022)]%
        {yang2022vox}
\bibfield{author}{\bibinfo{person}{Xingrui Yang}, \bibinfo{person}{Hai Li}, \bibinfo{person}{Hongjia Zhai}, \bibinfo{person}{Yuhang Ming}, \bibinfo{person}{Yuqian Liu}, {and} \bibinfo{person}{Guofeng Zhang}.} \bibinfo{year}{2022}\natexlab{}.
\newblock \showarticletitle{Vox-Fusion: Dense Tracking and Mapping with Voxel-based Neural Implicit Representation}.
\newblock \bibinfo{journal}{\emph{arXiv preprint arXiv:2210.15858}} (\bibinfo{year}{2022}).
\newblock


\bibitem[Zhang et~al\mbox{.}(2022)]%
        {zhang2022asro}
\bibfield{author}{\bibinfo{person}{Jiazhao Zhang}, \bibinfo{person}{Yijie Tang}, \bibinfo{person}{He Wang}, {and} \bibinfo{person}{Kai Xu}.} \bibinfo{year}{2022}\natexlab{}.
\newblock \showarticletitle{ASRO-DIO: Active subspace random optimization based depth inertial odometry}.
\newblock \bibinfo{journal}{\emph{IEEE Transactions on Robotics}} \bibinfo{volume}{39}, \bibinfo{number}{2} (\bibinfo{year}{2022}), \bibinfo{pages}{1496--1508}.
\newblock


\bibitem[Zhang et~al\mbox{.}(2021)]%
        {zhang2021rosefusion}
\bibfield{author}{\bibinfo{person}{Jiazhao Zhang}, \bibinfo{person}{Chenyang Zhu}, \bibinfo{person}{Lintao Zheng}, {and} \bibinfo{person}{Kai Xu}.} \bibinfo{year}{2021}\natexlab{}.
\newblock \showarticletitle{ROSEFusion: random optimization for online dense reconstruction under fast camera motion}.
\newblock \bibinfo{journal}{\emph{ACM Trans. on Graph. (SIGGRAPH)}} \bibinfo{volume}{40}, \bibinfo{number}{4} (\bibinfo{year}{2021}), \bibinfo{pages}{1--17}.
\newblock


\bibitem[Zhang et~al\mbox{.}(2023)]%
        {zhang2023go}
\bibfield{author}{\bibinfo{person}{Youmin Zhang}, \bibinfo{person}{Fabio Tosi}, \bibinfo{person}{Stefano Mattoccia}, {and} \bibinfo{person}{Matteo Poggi}.} \bibinfo{year}{2023}\natexlab{}.
\newblock \showarticletitle{Go-slam: Global optimization for consistent 3d instant reconstruction}. In \bibinfo{booktitle}{\emph{Proceedings of the IEEE/CVF International Conference on Computer Vision}}. \bibinfo{pages}{3727--3737}.
\newblock


\bibitem[Zhu et~al\mbox{.}(2025)]%
        {zhu2025_loopsplat}
\bibfield{author}{\bibinfo{person}{Liyuan Zhu}, \bibinfo{person}{Yue Li}, \bibinfo{person}{Erik Sandström}, \bibinfo{person}{Shengyu Huang}, \bibinfo{person}{Konrad Schindler}, {and} \bibinfo{person}{Iro Armeni}.} \bibinfo{year}{2025}\natexlab{}.
\newblock \showarticletitle{LoopSplat: Loop Closure by Registering 3D Gaussian Splats}. In \bibinfo{booktitle}{\emph{International Conference on 3D Vision (3DV)}}.
\newblock


\bibitem[Zhu et~al\mbox{.}(2022)]%
        {zhu2022nice}
\bibfield{author}{\bibinfo{person}{Zihan Zhu}, \bibinfo{person}{Songyou Peng}, \bibinfo{person}{Viktor Larsson}, \bibinfo{person}{Weiwei Xu}, \bibinfo{person}{Hujun Bao}, \bibinfo{person}{Zhaopeng Cui}, \bibinfo{person}{Martin~R Oswald}, {and} \bibinfo{person}{Marc Pollefeys}.} \bibinfo{year}{2022}\natexlab{}.
\newblock \showarticletitle{Nice-slam: Neural implicit scalable encoding for slam}. In \bibinfo{booktitle}{\emph{Proceedings of the IEEE/CVF Conference on Computer Vision and Pattern Recognition}}. \bibinfo{pages}{12786--12796}.
\newblock


\end{thebibliography}
